\pdfoutput=1 

\documentclass[10pt,twocolumn,letterpaper]{article}

\usepackage{cvpr}              
\definecolor{cvprblue}{rgb}{0.21,0.49,0.74}
\usepackage[pagebackref,breaklinks,colorlinks,allcolors=cvprblue]{hyperref}
\usepackage[utf8]{inputenc}
\usepackage{amsmath,amsfonts,amssymb}
\usepackage{array,booktabs,colortbl,multirow,tabularx}
\usepackage{algorithm,algpseudocode}
\usepackage{enumitem}
\usepackage{longtable}
\usepackage{array}
\usepackage{listings}
\usepackage[most]{tcolorbox}
\usepackage{newfloat}
\usepackage{makecell}
\usepackage{graphicx}
\usepackage{soul}
\usepackage{flafter}
\usepackage{float}
\definecolor{rankone}{gray}{0.85}
\definecolor{ranktwo}{gray}{0.93}
\usepackage{xcolor} 
\definecolor{SoftBlue}{RGB}{230, 240, 255}
\usepackage{booktabs}
\definecolor{NiceGreen}{RGB}{2,158,115} 
\definecolor{NiceRed}{RGB}{213,94,0}
\newcommand{\up}[1]{\textcolor{NiceGreen}{\scriptsize$\uparrow$#1}}
\newcommand{\down}[1]{\textcolor{NiceRed}{\scriptsize$\downarrow$#1}}

\newcommand{\Down}[1]{\textcolor{NiceRed}{$\downarrow$#1}}

\definecolor{headerblue}{RGB}{240, 248, 255}
\sethlcolor{headerblue}
\newcommand{\myfinding}[1]{%
  \vspace{0.3em}\noindent\hl{\textbf{#1.}}}

\usepackage{pifont} 
\newtcolorbox{finding}[1]{%
  breakable, enhanced,
  colback=black!5, colframe=black!45,
  boxrule=0.6pt, arc=4pt,
  left=8pt, right=8pt, top=6pt, bottom=6pt,
  before skip=8pt, after skip=8pt,
  before upper={\bfseries\itshape Finding~#1:\quad\normalfont}%
}
\usepackage{amsthm}
\theoremstyle{definition}
\newtheorem{theorem}{Theorem}
\newtheorem{definition}[theorem]{Definition}
\newcolumntype{L}[1]{>{\raggedright\arraybackslash}p{#1}}
\newcolumntype{M}[1]{>{\raggedright\arraybackslash}m{#1}}
\newcolumntype{B}[1]{>{\bfseries\raggedright\arraybackslash}m{#1}}

  \DeclareRobustCommand{\apicode}[1]{\begingroup\Urlmuskip=0mu plus 1mu\urlstyle{tt}\nolinkurl{#1}\endgroup}
\newcommand{\std}[1]{$_{\pm#1}$}                     
\newcommand{\best}[1]{\textbf{#1}}                    
\newcommand{\second}[1]{\cellcolor{SoftBlue!40}#1}    

\definecolor{skname}{HTML}{1F3A5F}   
\definecolor{subtle}{gray}{0.45}     
\definecolor{rowalt}{HTML}{F2F4F7}   

\newcommand{\Tpad}{\rule{0pt}{2.3ex}}
\newcommand{\Bpad}{\rule[-1.0ex]{0pt}{0pt}}
 
\newlist{steps}{enumerate}{1}
\setlist[steps]{label={\color{subtle}\scriptsize\arabic*}, leftmargin=0.95em,
  labelsep=0.4em, labelwidth=0.55em, align=right, nosep, itemsep=0.6pt}
\newenvironment{Steps}
  {\begin{minipage}[t]{\linewidth}\raggedright\begin{steps}}
  {\Bpad\end{steps}\end{minipage}}

\def\paperID{12324} 
\def\confName{CVPR}
\def\confYear{2027}

\title{SpatialCraft: Evolving Procedural Intelligence from Experience \\and Skills for Spatial Reasoning Agents}

\author{
Xiwei Liu$^{1}$,
Yulong Li$^{1}$,
Xuhui Li$^{1}$,
Zhixiang Lu$^{2}$,
Haolin Yang$^{1}$ \\
Jianxu Chen$^{1}$,
Yifan Lu$^{1}$,
Di Fan$^{3}$,
Yutong Xie$^{1,}$\thanks{Corresponding author.}\\[4pt]
$^{1}$Mohamed bin Zayed University of Artificial Intelligence \quad
$^{2}$University of Liverpool \\
$^{3}$University of Southern California
}

\begin{document}
\maketitle

\begin{abstract}
Spatial reasoning agents increasingly rely on external tools, such as 3D reconstruction, segmentation, and geometric computation, to answer questions that a multimodal large language model (MLLM) cannot resolve from pixels alone. Tool access, however, shifts the bottleneck to orchestration: the agent must decide which tools to call, in what order, and how to interpret their outputs, yet a stateless agent repeats the same mistakes across tasks. We ask whether a frozen tool-using agent can improve its orchestration from its own interaction history, without parameter updates. We present \textsc{SpatialCraft}, a self-evolving framework that maintains two forms of external knowledge: \emph{Experiences}, condition-action tactics distilled by contrasting rollouts of the same task, and \emph{Skills}, multi-step procedures with explicit activation and termination conditions. Three designs couple the two streams. Experiences can be anchored to a Skill, appearing only while it is active rather than once per task. A Contrastive Utility Score estimates each Experience's effect by randomly withholding it across rollouts of the same task, separating its contribution from task difficulty, and Experiences with consistently positive utility are consolidated into the Skills they support. Across five spatial benchmarks and four backbones, \textsc{SpatialCraft} achieves the best accuracy in all 20 settings, showing that a frozen agent can learn to orchestrate spatial tools from its own experience. \href{https://github.com/Xiwei-web/SpatialCraft}{Code}
\end{abstract}

\section{Introduction}
\label{sec:intro}
Spatial reasoning, the ability to infer 3D distances, orientations, and object relations from visual observations, is a fundamental prerequisite for embodied agents and robotic manipulation~\cite{yang2025thinkingspacemultimodallarge, marsili2025visualagenticai, ma20253dsrbenchcomprehensive3dspatial, jia2026omnispatialcomprehensivespatialreasoning, daxberger2025mmspatialexploring3dspatial}. Yet multimodal large language models (MLLMs) frequently fail at complex spatial queries when relying solely on internal parametric knowledge~\cite{chen2024spatialvlmendowingvisionlanguagemodels, cheng2024spatialrgptgroundedspatialreasoning, cai2025spatialbotprecisespatialunderstanding, ma20253dsrbenchcomprehensive3dspatial, jia2026omnispatialcomprehensivespatialreasoning, daxberger2025mmspatialexploring3dspatial, li2025viewspatialbenchevaluatingmultiperspectivespatial, song2026robospatialteachingspatialunderstanding, wang2025sitespatialintelligencethorough}. Consequently, a growing line of work equips MLLMs with external spatial tools, such as 3D reconstruction, metric depth estimation, and pose estimation, enabling them to explicitly compute metric geometric quantities rather than relying on ungrounded visual perception~\cite{dai2026sagentspatialtooluseelicits, chen2026spacetoolstoolaugmentedspatialreasoning, cho2026spatialclawrethinkingactioninterface, marsili2025visualagenticai, luo2026pyspatialvisualprograms, zhou2026spatialclireasonspatialtools, liu2026groundingisntknowingvlms}.


However, tool integration shifts the bottleneck from perception to orchestration. A tool-augmented agent must decide which tools to call, in what order, and how to interpret their outputs. Because these decisions are interdependent, a single tactical error can derail the entire reasoning chain, yet most spatial agents operate statelessly, discarding trial-and-error insights after each episode. Supervised fine-tuning or reinforcement learning can internalize spatial and tool-use competence~\cite{chen2024spatialvlmendowingvisionlanguagemodels, cai2025spatialbotprecisespatialunderstanding, ray2025satdynamicspatialaptitude, li2026spatialevoselfevolvingspatialintelligence, liu2026selfevolvingspatialreasoningvision, zhao2026ouroborosspatialclosingdatamodelloop, chen2026spacetoolstoolaugmentedspatialreasoning, zhou2026spatialclireasonspatialtools}, but must be repeated whenever the backbone or toolset changes. An alternative is parameter-update-free self-evolution, in which a frozen agent accumulates knowledge in an external memory~\cite{shinn2023reflexion, zhao2024expel, wang2024agent, zhang2026memrlselfevolvingagentsruntime,zhang2026agentic}. SMA~\cite{zhang2026spatial} shows that such memory benefits tool-free spatial reasoning. For tool-augmented agents, however, useful knowledge spans both multi-step tool procedures and step-level tactics, raising the challenge of how to structure, validate, and integrate it without accumulating spurious or redundant heuristics.

To address this challenge, we draw on the hierarchical nature of tool use, which prior work captures with two complementary forms of reusable knowledge~\cite{jiang2026xskillcontinuallearningexperience}: procedural \emph{Skills}, which orchestrate multi-step tool pipelines (e.g., reconstructing a scene before querying distances)~\cite{wang2023voyageropenendedembodiedagent, wang2024agent, zheng2025skillweaver, mi2026skill}, and tactical \emph{Experiences}, which guide localized decisions (e.g., re-prompting a segmentation model upon detecting a failure)~\cite{zhao2024expel, tang2025agent, ouyang2026reasoningbank}. In spatial reasoning, the two are intertwined, as a tactic's usefulness depends on the procedural stage: re-prompting segmentation helps during object grounding but is irrelevant during metric comparison. This suggests structuring knowledge hierarchically, with tactical Experiences anchored to the procedural Skills they support. Tactics that prove consistently effective can then be absorbed into the procedure itself, so that the Skill evolves and the Experience bank stays compact, much as human expertise compiles practiced rules into procedures~\cite{anderson1982acquisition}.


Beyond structural organization, a self-evolving agent requires empirical validation of its accumulated knowledge. MLLMs often distill lessons that are linguistically plausible but practically misleading~\cite{huang2024large}, and retaining such untested heuristics lets the memory accumulate errors alongside competence. Correlating a lesson with successful episodes, as in reward-based memory valuation~\cite{zhang2026memrlselfevolvingagentsruntime, zhang2026spatial}, is insufficient because the correlation is confounded by task difficulty: heuristics retrieved for inherently easier tasks may appear effective regardless of their actual impact. Isolating the utility of a lesson instead requires a controlled comparison. By randomly showing an Experience in only a subset of the parallel rollouts of the same task~\cite{shao2024deepseekmath}, the agent can contrast their outcomes to isolate the effect of the tactic from task difficulty, at no extra interaction cost, since these rollouts are already collected to distill Experiences.

Building on these insights, we present \textsc{SpatialCraft}, a parameter-update-free framework that co-evolves tactical Experiences and procedural Skills for tool-augmented spatial agents. During exploration, \textsc{SpatialCraft} leverages multi-trajectory rollouts to distill Experiences, anchoring each strictly to the active Skill. The framework evaluates each Experience using a Contrastive Utility Score, which estimates its contribution by comparing within-task rollouts with and without the tactic. Experiences that systematically improve outcomes are consolidated into their anchor Skills, while ineffective ones are retired. Skills themselves are refined through trajectory feedback, with proposed revisions rigorously verified against prior decision states before adoption. During deployment, the accumulated dual-level knowledge base is frozen, enabling the agent to seamlessly retrieve task-level insights and trigger context-aware tactics precisely when the corresponding Skill is invoked, so tactics surface where they apply.

Experiments across five representative spatial benchmarks and four
base VLMs show that \textsc{SpatialCraft} consistently outperforms
tool-only and memory-based baselines, demonstrating that evolving
procedural knowledge provides an effective parameter-update-free path
for spatial self-evolution. Our core contributions are three-fold:


\noindent\ding{182} We propose \textsc{SpatialCraft}, a parameter-update-free framework that organizes tactical Experiences around the task-level Skills they support and co-evolves both, enabling frozen spatial reasoning agents to transform verified interaction into reusable procedural intelligence.


\noindent\ding{183} We introduce the Contrastive Utility Score, which randomizes each Experience's exposure across rollouts of the same task to estimate its effect separately from task difficulty, together with a judge-based verification that accepts Skill revisions by replaying recorded decision states and requires no token likelihoods, suiting proprietary MLLMs.

\noindent\ding{184} We evaluate \textsc{SpatialCraft} across diverse spatial
    benchmarks and VLM backbones, demonstrating consistent performance
    gains together with effective in-domain, cross-task, and
    cross-agent knowledge reuse.

\section{Methodology}
\label{sec:method}
\begin{figure*}[!t]
    \centering
    \includegraphics[width=\linewidth]{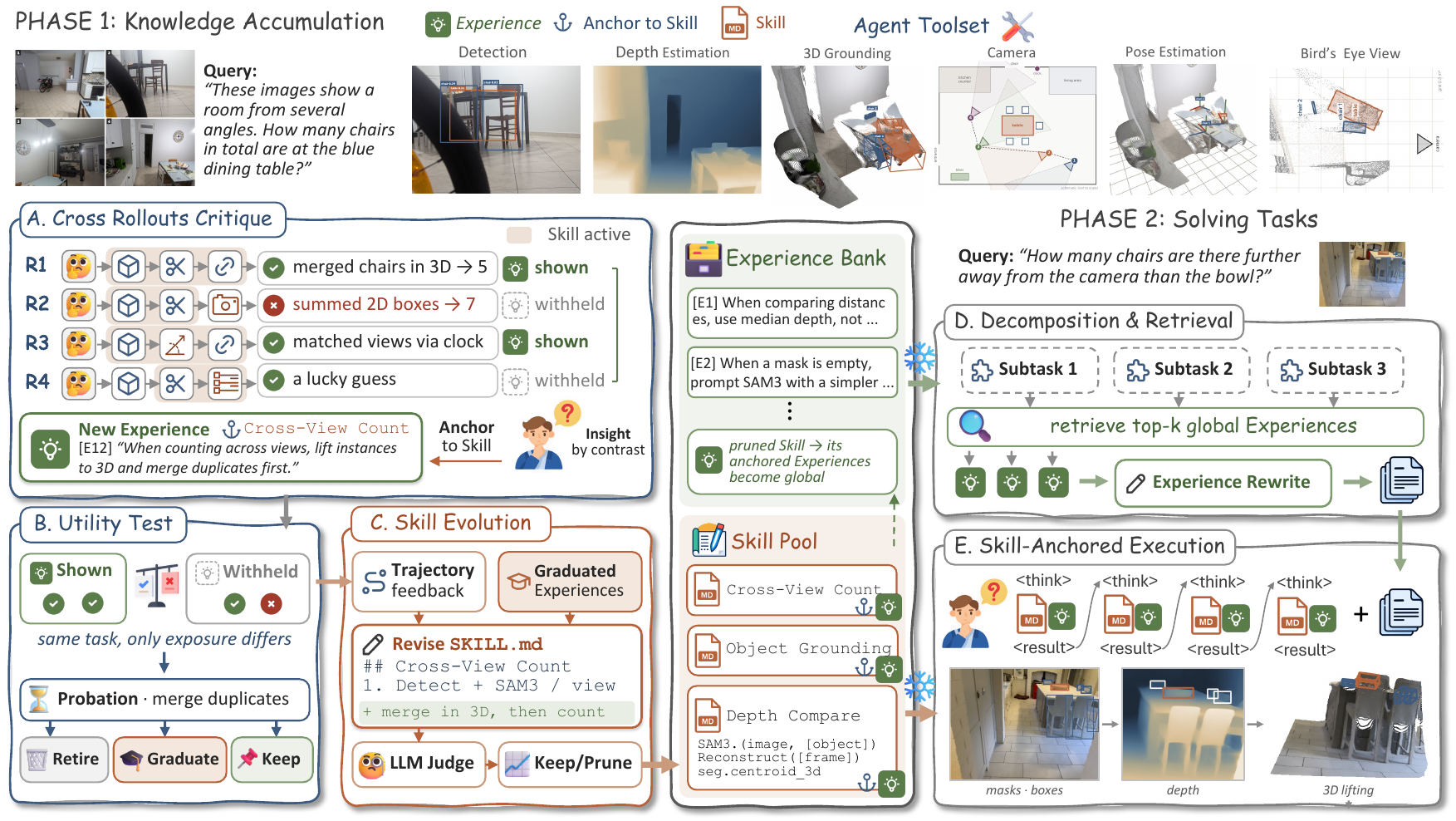}
    \vspace{-2em}
    \caption{\textbf{Overview of SpatialCraft.}
\textbf{Phase 1:} \textbf{(A)} contrasting rollouts of the same task distills Experiences, each anchored to the active Skill; \textbf{(B)} randomly withholding an Experience across rollouts isolates its effect from task difficulty; \textbf{(C)} proven Experiences are folded into their Skills.
\textbf{Phase 2:} the agent \textbf{(D)} retrieves global Experiences per subtask and \textbf{(E)} activates Skills that bring their anchored Experiences.}
    \label{fig:overview}
    \vspace{-1em}
\end{figure*}
As shown in Figure~\ref{fig:overview}, \textsc{SpatialCraft} augments a frozen multimodal agent through two external knowledge stores: an Experience Bank $\mathcal{E}$ of action-level tactics and a Skill Pool $\mathcal{K}$ of temporally extended procedures.

\subsection{Problem Setup}
\label{sec:setup}
 
We consider a task instance $\mathcal{T}=(q,\mathcal{I})$, where $q$ is a spatial query and $\mathcal{I}=\{I_1,\ldots,I_m\}$ are the image inputs. A frozen MLLM $\mathcal{M}_{\text{exec}}$ solves it as a sequential decision process with a toolset $\mathcal{F}=\{f_1,\ldots,f_n\}$ covering reconstruction, segmentation, pose estimation, and geometric computation. At step $t$, conditioned on the state $s_t$ (query, images, and interaction history), it selects $a_t\in\mathcal{F}\cup\{\textsc{answer}\}$ and receives an observation $o_t$, yielding a trajectory $\tau=[(s_0,a_0,o_0),\ldots,(s_T,a_T,o_T)]$ that concludes with an answer $\hat{y}$ and a binary reward $R(\tau)=\mathbb{1}[\hat{y}=y^{*}]$. A second instance $\mathcal{M}_{\text{kb}}$ of the same backbone performs knowledge operations over an external knowledge base $\mathcal{KB}=(\mathcal{K},\mathcal{E})$.
\begin{definition}[\textbf{Skill}]
A \textit{Skill} $k\in\mathcal{K}$ is a reusable procedural unit written in structured natural language that provides temporal guidance. It takes the form $k=(\omega_k,\pi_k,\beta_k)$, where $\omega_k$ is the activation condition specifying when $k$ should be invoked, $\pi_k$ is the execution procedure that guides the action sequence while $k$ is active, and $\beta_k$ is the termination condition that returns control to the agent.
\end{definition}

\begin{definition}[\textbf{Experience}]
An \textit{Experience} $e\in\mathcal{E}$ is a non-parametric tactical prompt that records action-level knowledge. It takes the form $e=(c,a,\alpha_e,\mathbf{v}_e)$, where $c$ is the triggering condition, $a$ is the recommended action, $\alpha_e\in\mathcal{K}\cup\{\varnothing\}$ is the anchor Skill under which $e$ applies, and $\mathbf{v}_e\in\mathbb{R}^d$ is the semantic embedding used for retrieval. $e$ is \emph{anchored} if $\alpha_e\in\mathcal{K}$ and \emph{global} if $\alpha_e=\varnothing$.
\end{definition}
With $\mathcal{M}_{\text{exec}}$ frozen, $\mathcal{KB}$ is constructed on the environment split $\mathcal{D}_{\text{env}}$ and then used read-only on the deployment split $\mathcal{D}_{\text{dep}}$, where no knowledge is written and evolved:
\begin{equation}
\max_{\mathcal{KB}}\ \mathbb{E}_{\mathcal{T}\sim\mathcal{D}_{\text{dep}}}\,\mathbb{P}\big[\hat{y}=y^{*}\mid\mathcal{T},\mathcal{KB};\mathcal{M}_{\text{exec}}\big].
\setlength\abovedisplayskip{4pt}%
\setlength\belowdisplayskip{4pt}%
\end{equation}
 
\subsection{Experience Accumulation}
\label{sec:exp}
 
\paragraph{Visually Grounded Rollout Summary.}
For each task $\mathcal{T}_i=(q_i,\mathcal{I}_i)\in\mathcal{D}_{\text{env}}$, $\mathcal{M}_{\text{exec}}$ runs $N$ rollouts with the current knowledge base, yielding $\mathcal{R}_i=\{\tau_i^{(1)},\ldots,\tau_i^{(N)}\}$. Following~\cite{jiang2026xskillcontinuallearningexperience}, $\mathcal{M}_{\text{kb}}$ summarizes them by jointly considering the trajectories, task images, query, and ground truth, $\mathcal{S}_i=\mathcal{M}_{\text{kb}}(\mathcal{R}_i,\mathcal{I}_i,q_i,y_i^{*})$, which captures key decision points, tool-use patterns, visual evidence, and failure causes. The summary additionally records, for each step, the active Skill and the Experiences shown to the agent. See Appendix~\ref{app:prompt_design} for prompt details; Appendix~\ref{app:case study} for examples.
 
\paragraph{Cross-Rollout Critique.}
Given the Experiences $\mathcal{E}_i$ used during the rollouts, $\mathcal{M}_{\text{kb}}$ contrasts successful and failed rollouts to identify the causal factors behind their outcomes, and produces a set of structured updates:
\begin{equation}
\Delta\mathcal{E}_i=\mathcal{M}_{\text{kb}}\big(\mathcal{S}_i,\,y_i^{*},\,\mathcal{E}_i\big)=\{op_1,\ldots,op_{M_i}\},
\setlength\abovedisplayskip{4pt}%
\setlength\belowdisplayskip{4pt}%
\end{equation}
where each operation is either $(\textit{add},e)$, which introduces a new Experience, or $(\textit{modify},e_{\text{id}},e')$, which refines an existing Experience $e_{\text{id}}$ into $e'$.
Each new Experience is anchored to the Skill that the summary records as active at the step whose decision it addresses, or left global ($\alpha_e=\varnothing$) if no Skill was active at that step or it spans several Skills; a modified Experience inherits the anchor of the one it refines.
\vspace{-1em}
\paragraph{Experience Consolidation.}
Before an added Experience $e$ is committed, the system checks whether any existing entry has cosine similarity above $\theta_{\text{sim}}$ with $e$; if so, $\mathcal{M}_{\text{kb}}$ merges them into a single Experience that preserves their generalizable insights. A merge of Experiences with the same anchor keeps that anchor; otherwise the merged Experience becomes global, since its lesson has applied beyond a single procedure. When the bank exceeds its capacity $N^{\mathcal{E}}_{\max}$, $\mathcal{M}_{\text{kb}}$ first merges remaining redundancies, and the excess is removed in ascending order of utility (Sec.~\ref{sec:cus}).
 
\subsection{Contrastive Utility Estimation}
\label{sec:cus}
Not every distilled Experience helps: a lesson may be overly specific, or plausible in language yet misleading in practice. We therefore estimate its utility by contrasting same-task rollouts that differ in whether it is shown.
\vspace{-1em} 
\paragraph{Randomized Exposure.}
An Experience is \emph{eligible} in a rollout if it is retrieved for the task or its anchor Skill is activated during the rollout. Before the rollouts of $\mathcal{T}_i$, each Experience $e$ is assigned to a random subset $M_i(e)\subset\{1,\ldots,N\}$ with $|M_i(e)|=\lfloor N/2\rfloor$, independently across Experiences, and is shown in rollout $j$ only if it is eligible and $j\in M_i(e)$. Let $X_i^{+}(e)$ and $X_i^{-}(e)$ denote the rollouts in which $e$ is eligible and shown or withheld, respectively. Because exposure is randomized, the two groups differ systematically only in whether $e$ is shown, so their outcome difference reflects the effect of $e$ rather than task difficulty or the agent's own choice to follow it.
\vspace{-1em} 
\paragraph{Contrastive Utility Score.}
For each task in which both groups are non-empty, we compute the within-task contrast
\begin{equation}
\scriptsize
\delta_i(e)=\frac{1}{|X_i^{+}(e)|}\sum_{j\in X_i^{+}(e)}R\big(\tau_i^{(j)}\big)-\frac{1}{|X_i^{-}(e)|}\sum_{j\in X_i^{-}(e)}R\big(\tau_i^{(j)}\big),
\setlength\abovedisplayskip{4pt}%
\setlength\belowdisplayskip{4pt}%
\end{equation}
and accumulate $S_e\leftarrow S_e+\delta_i(e)$ and $n_e\leftarrow n_e+1$. The Contrastive Utility Score (CUS) is $U(e)=\frac{S_e}{\lambda+n_e}$, where $\lambda$ acts as $\lambda$ virtual observations of zero effect, shrinking low-evidence estimates toward zero. Unlike an average reward, $U$ is signed: a positive value indicates that showing $e$ improves outcomes, and a negative value that it harms them.
\vspace{-2em} 


\paragraph{Utility-Driven Lifecycle.}
Each Experience $e$ is assigned a state by $U(e)$ and its number of contrasts $n_e$: (1) \textbf{Probation} if $n_e < n_{\min}$, ensuring sufficient initial evaluation; (2) \textbf{Retired} if $n_e \ge n_{\min}$ and $U(e) \le 0$, triggering its eviction from the memory bank; (3) \textbf{Graduated} if $n_e \ge n_{\min}$, the anchor skill $\alpha_e \neq \varnothing$, and $U(e) \ge \tau_g$, promoting it for consolidation; and (4) \textbf{Retained} otherwise, maintaining it for ongoing use.
A new Experience starts at $S_e=n_e=0$, and merged Experiences sum their statistics. A modified Experience keeps its statistics if $U(e)>0$ (refining a helpful lesson) but resets them to zero if $U(e) \le 0$, so that the corrected lesson is not biased by its predecessor's failures.
 
\subsection{Skill Evolution}
\label{sec:skill}
 

Following the seed-based strategy of~\cite{mi2026skill}, the Skill Pool $\mathcal{K}$ is initialized with six general-purpose seeds (Appendix~\ref{app:seeds}) that encode generic reasoning procedures rather than task-specific spatial knowledge. The pool then evolves through interaction, refining existing Skills, synthesizing new ones, and pruning those that underperform.

\subsubsection{Semantic Gradients and Consolidation}
\label{sec:consolidate}
\paragraph{Semantic Gradients.}
We use semantic gradients, natural-language update directions derived from trajectories, as learning signals for Skills. For each of the $B$ trajectories $\tau_i\in\mathcal{T}^{(B)}_k$ in which $k$ was active, $\mathcal{M}_{\text{kb}}$ analyzes the segment controlled by $k$ and attributes its outcome to $(\omega_k,\pi_k,\beta_k)$:
\begin{equation}
g_i=\nabla_{\text{sem}}(\tau_i,k)=\big(g_i^{(\omega)},\,g_i^{(\pi)},\,g_i^{(\beta)}\big).
\setlength\abovedisplayskip{4pt}%
\setlength\belowdisplayskip{4pt}%
\end{equation}
Individual gradients may conflict, so $\mathcal{M}_{\text{kb}}$ aggregates them into $\bar{g}_k=\mathrm{Agg}(\{g_i\}_{i=1}^{B})$, retaining recurring failure patterns and consistent refinement suggestions.


\vspace{-1em}
\paragraph{Experience-to-Skill Consolidation.}
A graduated Experience has proven useful within its anchor Skill, yet as a separate prompt it remains outside the procedure it supports. We therefore fold it into the Skill during revision. Let $\mathcal{G}_k$ denote the graduated Experiences anchored to $k$; $\mathcal{M}_{\text{kb}}$ generates $N_c$ candidate revisions from both $\bar{g}_k$ and $\mathcal{G}_k$:
\begin{equation}
k'=k\oplus\big(\bar{g}_k,\,\mathcal{G}_k\big),
\setlength\abovedisplayskip{4pt}%
\setlength\belowdisplayskip{4pt}%
\end{equation}
where $\oplus$ is an LLM-driven update with two modes. In \emph{refinement} mode, it revises $k$ along $\bar g_k$ and writes each Experience in $\mathcal{G}_k$ into the step of $\pi_k$, or the condition $\beta_k$, at which its trigger arises; these Experiences leave the bank once the revision is accepted. In \emph{discovery} mode, used when the gradients reveal a recurring situation that $k$ does not cover, it synthesizes a structurally new Skill from the same evidence.

\subsubsection{LLM-as-Judge Skill Verification}
\label{sec:verify}
Revisions derived from hindsight may introduce steps that are plausible in language but unsupported in practice. We therefore compare the actions that the parent and candidate Skills induce at the same historical decision states.
\vspace{-1em}
\paragraph{Key-State Replay.}
From each $\tau_i\in\mathcal{T}^{(B)}_k$, we select one state $s_i$ at which $k$ guided the action: the earliest such step with direct evidence of an action error, or else the earliest such step. These states and their recorded actions $a_i$ are fixed in advance and shared across all candidates. For each candidate $k'$, we replace only the active Skill in the recorded context and let $\mathcal{M}_{\text{exec}}$ generate one action $a'_i\sim\mathcal{M}_{\text{exec}}(\cdot\mid s_i,k')$ under the same budget, without executing it or observing $a_i$ or the trajectory outcome.
\vspace{-1em}
\paragraph{Action Comparison.}
An MLLM judge $\mathcal{M}_{\text{judge}}$ compares $a'_i$ and $a_i$ against the task requirements, shared context, and available evidence, and returns a candidate-directed score $j_i(k')\in\{+1,0,-1\}$, where $0$ denotes a tie or insufficient evidence; identical actions receive $0$ without a judge call. With $J(k')=\frac{1}{B}\sum_{i=1}^{B}j_i(k')$, the best candidate is accepted only if its score is positive:
\begin{equation}
k_{\text{new}}=\arg\max_{k'\in\mathcal{C}(k)}J(k')\quad\text{s.t.}\quad J(k_{\text{new}})>0,
\setlength\abovedisplayskip{4pt}%
\setlength\belowdisplayskip{4pt}%
\end{equation}
where $\mathcal{C}(k)$ is the set of $N_c$ candidates; otherwise $k$ is retained. A positive $J$ indicates more local wins than losses at the replayed states; whether the revision improves full trajectories is monitored by the online score (Sec.~\ref{sec:maintain}).

\subsubsection{Score-Based Skill Pool Maintenance}
\label{sec:maintain}
 Skills are retained or pruned according to an online score of their empirical contribution, computed with the same within-task contrast used for Experiences. For task $\mathcal{T}_i$, let $A_i(k)$ and $\bar{A}_i(k)$ be the rollouts in which $k$ is and is not activated. When both are non-empty,
\begin{equation}
\scriptsize
\delta_i(k)=\frac{1}{|A_i(k)|}\sum_{j\in A_i(k)}R\big(\tau_i^{(j)}\big)-\frac{1}{|\bar{A}_i(k)|}\sum_{j\in\bar{A}_i(k)}R\big(\tau_i^{(j)}\big),
\setlength\abovedisplayskip{4pt}%
\setlength\belowdisplayskip{4pt}%
\end{equation}
which accumulates into $S_k$ and $n_k$, giving $\mathrm{Score}(k)=S_k/(\lambda+n_k)$. Skill activation depends on the trajectory rather than being randomized, and revisions are gated by verification. An accepted revision inherits the statistics of its parent under the same rule as Experience modification.

To respect the pool capacity $N^{\mathcal{K}}_{\max}$, we remove (i) Skills with $n_k\ge n_{\min}$ and $\mathrm{Score}(k)\le 0$, and (ii) Skills that are semantically redundant. If $\mathcal{K}$ still exceeds capacity, Skills are removed in ascending order of score. Experiences anchored to a removed Skill become global, so tactical knowledge outlives the procedure in which it was learned.

\begin{table*}[t]
\centering
\caption{Main results (accuracy, \%) on five spatial benchmarks (Robo.\,=\,RoboSpatial, Omni.\,=\,Omni3D, View.\,=\,ViewSpatial); Avg.\ is the macro average, with the gain over the no-memory CoT baseline. Mean$_{\pm\text{std}}$ over three runs. \textbf{Bold}: best; \colorbox{SoftBlue!40}{shaded}: second best.}
\label{tab:main}
\footnotesize
\setlength{\tabcolsep}{2.6pt}
\renewcommand{\arraystretch}{1.0}
\vspace{-0.8em}
\begin{tabular}{@{}l *{5}{c} c @{\hspace{5pt}}c@{\hspace{5pt}} *{5}{c} c@{}}
\toprule
 & \multicolumn{6}{c}{\textbf{GPT-5.4}} & & \multicolumn{6}{c}{\textbf{GPT-5.4-mini}} \\
\cmidrule(lr){2-7}\cmidrule(lr){9-14}
Method & Robo. & ERQA & Omni. & SAT & View. & Avg. & & Robo. & ERQA & Omni. & SAT & View. & Avg. \\
\midrule
\rowcolor{black!5}CoT (no mem.) & 59.4\std{1.1} & 47.5\std{0.9} & 25.7\std{0.2} & 66.1\std{1.6} & 45.6\std{0.1} & 48.9 & & 54.5\std{1.8} & 34.5\std{0.1} & 23.6\std{1.1} & 56.1\std{1.4} & 33.5\std{0.6} & 40.4 \\
ReAct (tools) & 61.1\std{2.1} & 58.0\std{0.5} & 30.4\std{1.4} & 68.7\std{1.2} & 49.2\std{0.6} & 53.5 \up{4.6} & & 55.2\std{1.2} & 35.2\std{1.9} & 29.5\std{1.0} & 60.2\std{0.8} & 38.9\std{0.3} & 43.8 \up{3.4} \\
RAG & 58.9\std{1.5} & 64.0\std{1.8} & 31.2\std{1.1} & 91.3\std{0.9} & 52.2\std{0.4} & 59.5 \up{10.6} & & 61.5\std{2.0} & 55.2\std{0.8} & 33.9\std{1.5} & 85.1\std{0.5} & 46.2\std{0.4} & 56.4 \up{16.0} \\
MemP & 65.1\std{0.6} & 65.5\std{1.3} & 34.0\std{1.6} & 91.7\std{1.3} & 60.6\std{0.3} & 63.4 \up{14.5} & & 63.8\std{2.2} & 58.2\std{2.1} & 30.9\std{0.6} & 85.1\std{1.4} & 45.2\std{0.5} & 56.6 \up{16.2} \\
MemRL-R & 67.4\std{2.3} & 67.0\std{2.0} & 35.6\std{0.4} & 92.3\std{1.0} & 61.8\std{0.5} & 64.8 \up{15.9} & & 63.2\std{0.9} & 59.0\std{1.3} & 29.9\std{1.7} & 85.3\std{1.5} & 47.3\std{0.7} & 56.9 \up{16.5} \\
MemRL-GT & 69.1\std{1.7} & 69.5\std{1.5} & 37.2\std{1.2} & 92.7\std{0.3} & 65.9\std{0.7} & 66.9 \up{18.0} & & 65.5\std{1.4} & 51.0\std{1.8} & 32.8\std{1.6} & 86.3\std{1.2} & 47.7\std{0.5} & 56.7 \up{16.3} \\
SMA & \second{72.0\std{1.1}} & \second{71.0\std{1.0}} & \second{42.4\std{0.8}} & \second{93.7\std{0.7}} & \second{66.3\std{0.7}} & \second{69.1 \up{20.2}} & & \second{67.4\std{2.1}} & \second{60.5\std{0.5}} & \second{38.4\std{1.1}} & \second{88.7\std{0.9}} & \second{55.9\std{0.3}} & \second{62.2 \up{21.8}} \\
\rowcolor{SoftBlue}\textbf{SpatialCraft} & \best{75.2\std{1.8}} & \best{74.7\std{1.6}} & \best{50.7\std{1.3}} & \best{94.2\std{1.1}} & \best{69.8\std{0.6}} & \best{72.9 \up{24.0}} & & \best{70.9\std{0.6}} & \best{62.0\std{2.0}} & \best{46.0\std{1.4}} & \best{90.3\std{0.7}} & \best{62.6\std{0.4}} & \best{66.4 \up{26.0}} \\
\midrule[\heavyrulewidth]
 & \multicolumn{6}{c}{\textbf{Qwen3.6-27B}} & & \multicolumn{6}{c}{\textbf{Qwen3.5-9B}} \\
\cmidrule(lr){2-7}\cmidrule(lr){9-14}
Method & Robo. & ERQA & Omni. & SAT & View. & Avg. & & Robo. & ERQA & Omni. & SAT & View. & Avg. \\
\midrule
\rowcolor{black!5}CoT (no mem.) & 54.3\std{0.7} & 52.1\std{1.3} & 39.5\std{2.0} & 80.9\std{1.5} & 49.2\std{0.2} & 55.2 & & 58.0\std{0.1} & 45.7\std{1.2} & 35.5\std{1.6} & 76.8\std{0.4} & 45.8\std{0.1} & 52.4 \\
ReAct (tools) & 62.9\std{1.5} & 54.0\std{1.5} & 42.8\std{1.2} & 84.3\std{1.0} & 53.9\std{0.6} & 59.6 \up{4.4} & & 57.0\std{2.2} & 43.0\std{1.8} & 32.1\std{1.7} & 77.0\std{1.5} & 43.1\std{0.6} & 50.4 \down{2.0} \\
RAG & 59.2\std{2.3} & 54.5\std{1.0} & 44.0\std{0.4} & 85.0\std{1.3} & 49.8\std{0.5} & 58.5 \up{3.3} & & 55.6\std{1.4} & 49.5\std{0.5} & 31.6\std{1.1} & 77.7\std{1.3} & 42.3\std{0.7} & 51.3 \down{1.1} \\
MemP & 65.3\std{2.0} & 51.7\std{1.6} & 44.0\std{0.8} & 86.0\std{0.3} & 55.3\std{0.4} & 60.5 \up{5.3} & & 53.5\std{1.2} & \second{53.0\std{1.3}} & 34.4\std{1.4} & 78.0\std{1.2} & 48.7\std{0.4} & 53.5 \up{1.1} \\
MemRL-R & 62.3\std{1.7} & 51.3\std{1.9} & 44.7\std{1.3} & 83.8\std{0.5} & 57.6\std{0.3} & 59.9 \up{4.7} & & 54.3\std{1.5} & 43.5\std{1.0} & 36.4\std{1.6} & 76.7\std{1.0} & 48.0\std{0.5} & 51.8 \down{0.6} \\
MemRL-GT & 67.2\std{0.9} & 55.7\std{0.8} & 43.7\std{0.6} & \second{86.9\std{0.8}} & \second{64.3\std{0.7}} & 63.6 \up{8.4} & & 52.8\std{1.2} & 49.0\std{2.0} & 34.4\std{0.8} & 80.3\std{0.9} & 48.7\std{0.4} & 53.0 \up{0.6} \\
SMA & \second{68.6\std{1.1}} & \second{58.3\std{2.1}} & \second{47.7\std{1.0}} & \second{86.9\std{1.1}} & 63.2\std{0.6} & \second{64.9 \up{9.7}} & & \second{58.5\std{0.9}} & 51.8\std{0.8} & \second{40.8\std{0.4}} & \second{81.3\std{0.7}} & \second{50.2\std{0.3}} & \second{56.5 \up{4.1}} \\
\rowcolor{SoftBlue}\textbf{SpatialCraft} & \best{70.5\std{1.8}} & \best{61.5\std{1.5}} & \best{49.3\std{1.5}} & \best{89.1\std{1.4}} & \best{65.8\std{0.3}} & \best{67.2 \up{12.0}} & & \best{59.4\std{0.6}} & \best{53.8\std{1.6}} & \best{43.6\std{1.2}} & \best{83.3\std{0.3}} & \best{54.7\std{0.5}} & \best{59.0 \up{6.6}} \\
\bottomrule
\end{tabular}
\vspace{-1em}
\end{table*}

\subsection{Solving Tasks with Experiences and Skills}
\label{sec:infer}

Global Experiences are retrieved once per task and adapted to it, Skills are selected as the reasoning state evolves, and each activated Skill brings its anchored Experiences with it.
\vspace{-1em}
\paragraph{Task Decomposition Retrieval.}
$\mathcal{M}_{\text{kb}}$ decomposes the task into $n_g$ abstract subtasks $\mathcal{G}=\{g_1,\ldots,g_{n_g}\}$ that capture distinct reasoning needs, such as depth ambiguity or viewpoint-dependent relations, and generates a short retrieval query for each. For each subtask, we retrieve the top-$n$ global Experiences by semantic similarity:
\begin{equation}
\footnotesize
\mathcal{E}^{\text{ret}}=\bigcup_{g\in\mathcal{G}}\operatorname{TopN}\big(\{e\in\mathcal{E}_{\varnothing}:\cos(\mathbf{v}_g,\mathbf{v}_e)>\theta_{\min}\}\big),
\setlength\abovedisplayskip{4pt}%
\setlength\belowdisplayskip{4pt}%
\end{equation}
where $\mathcal{E}_{\varnothing}=\{e\in\mathcal{E}:\alpha_e=\varnothing\}$ excludes anchored Experiences, which enter through their Skills. $\mathcal{M}_{\text{kb}}$ then rewrites each retrieved Experience for the current query and images, discarding those that do not apply.
\vspace{-1em}
\paragraph{Skill Selection.}
Skills are activated at state-dependent boundaries: at the start of a task and whenever the active Skill terminates. At a boundary, the Skill whose activation condition best matches the current state is selected,
\begin{equation}
k_t=\arg\max_{k\in\mathcal{K}}\ \cos\big(\mathbf{v}(s_t),\mathbf{v}_{\omega_k}\big),
\setlength\abovedisplayskip{4pt}%
\setlength\belowdisplayskip{4pt}%
\end{equation}
and the agent proceeds without a Skill if the maximum similarity is below $\theta_{\text{act}}$. A Skill remains active until $\mathcal{M}_{\text{kb}}$ judges its termination $\beta_{k_t}$ satisfied or a horizon of $H_{\max}$ steps is reached, so a single task may invoke a sequence of Skills. The anchored Experiences are injected into the context $\mathcal{E}^{\text{anc}}_{k_t}=\operatorname{TopN}\big(\{e\in\mathcal{E}:\alpha_e=k_t\}\big),$
ranked by similarity to the task query, and withdrawn when $k_t$ terminates.
\vspace{-2em}
\paragraph{Joint Guidance.}
At each step, $\mathcal{M}_{\text{exec}}$ selects the next action conditioned on the state, the active Skill, and both sets of Experiences: $a_t\sim\mathcal{M}_{\text{exec}}\big(\cdot\mid s_t,\ k_t,\ \mathcal{E}^{\text{ret}},\ \mathcal{E}^{\text{anc}}_{k_t}\big).$

\section{Experiments}

\subsection{Experimental Setup}
\begin{figure*}[t]
    \centering
    \includegraphics[width=\linewidth]{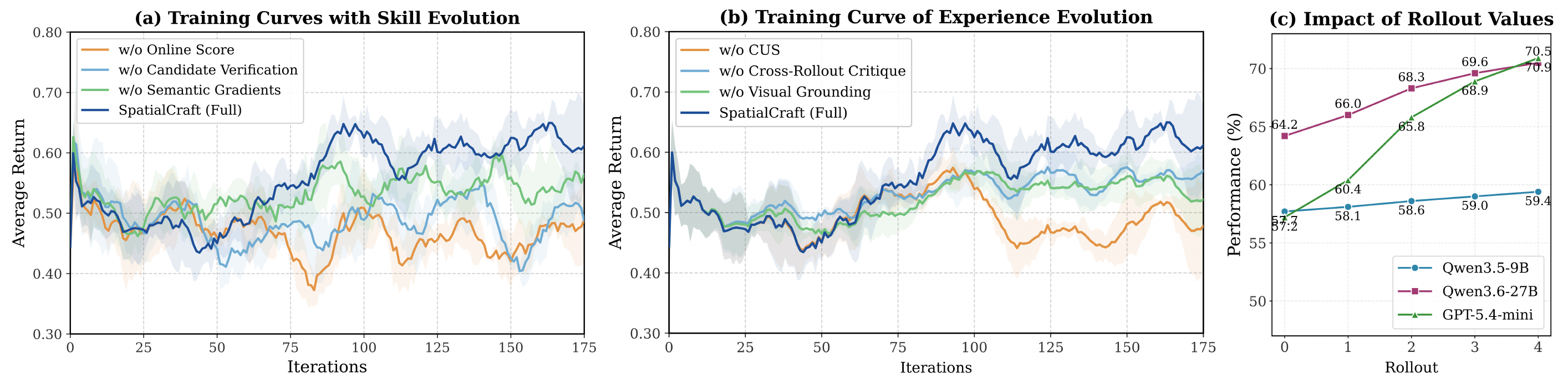}
    \vspace{-2em}
    \caption{(a) Training curves under ablations of skill evolution. (b) Training curves under ablation of experience evolution; Solid lines and shaded areas denote the smoothed mean and standard deviation of average returns, respectively. (c) Performance comparison across different rollout values on Robospatial. $N=0$ denotes the initial agent without knowledge accumulation.}
    \label{fig:training}
    \vspace{-1em}
\end{figure*}

\paragraph{Benchmarks.}
We evaluate on five spatial benchmarks: RoboSpatial~\cite{song2026robospatialteachingspatialunderstanding} and ERQA~\cite{geminirobotics2025bringingphysicalworld} for embodied scene and robot reasoning, Omni3D-Bench~\cite{marsili2025visualagenticai} for metric 3D queries, SAT~\cite{ray2025satdynamicspatialaptitude} for egocentric and two-frame dynamics, and ViewSpatial~\cite{li2025viewspatialbenchevaluatingmultiperspectivespatial} for multi-perspective orientation. All results are reported on $\mathcal{D}_{\text{dep}}$. Benchmark details and split statistics are given in Appendix~\ref{app:benchmark}.

\vspace{-1em}
\paragraph{Implementation.}
We use four backbones: GPT-5.4, GPT-5.4-mini, Qwen3.6-27B, and Qwen3.5-9B, with open-weight models served by vLLM~\cite{kwon2023efficientmemorymanagementlarge} in non-thinking mode. In each setting, $\mathcal{M}_{\text{exec}}$, $\mathcal{M}_{\text{kb}}$, and the Judge share the same backbone. Accumulation runs $N=4$ rollouts per task with temperature $0.7$ and top-$p$ $0.9$; deployment uses temperature $0$. Each call is capped at 16,384 output tokens and each trajectory at 50 steps. Retrieval and Skill activation use \texttt{text-embedding-3-large}. We set $B=6$, $N_c=3$, $N_p=2$, $\lambda=2$, $n_{\min}=3$, and $\tau_g=0.2$; the remaining detailed hyperparameters are listed in Appendix~\ref{app:hyper}. 
\vspace{-1em}
\paragraph{Baselines.}
We compare \textsc{SpatialCraft} with seven baselines: CoT, Tools-only ReAct~\cite{yao2022react}, RAG~\cite{lewis2021retrievalaugmentedgenerationknowledgeintensivenlp}, MemP~\cite{fang2026mempexploringagentprocedural}, MemRL-R~\cite{zhang2026memrlselfevolvingagentsruntime}, MemRL-GT and SMA~\cite{zhang2026spatial}. Since the original MemRL uses reward-only reflection, we implement MemRL-GT, which provides ground-truth answers during reflection for fair comparison. Baselines do not receive the seed Skills. Further details are specified in Appendix~\ref{app:setup}.
\vspace{-1em}
\paragraph{Tool Sets.}
We equip the spatial reasoning agent with a unified library of perception and geometry tools:
(1) \textbf{Reconstruct} and \textbf{Scale} for recovering depth and metric spatial structure;
(2) \textbf{SAM3} and \textbf{Detect} for object segmentation and localization;
(3) \textbf{Geometry} and \textbf{Pose} for estimating geometric relations and object orientation;
(4) \textbf{Mask} and \textbf{Draw} for region manipulation and visual annotation.
Full tool definitions are provided in Appendix~\ref{app:tool_definitions}.

\begin{table}[t]
\centering
\caption{Ablations on RoboSpatial and Omni3D with GPT-5.4-mini. Naive score: mean reward without the within-task contrast.}
\vspace{-0.8em}
\label{tab:ablation}
\setlength{\tabcolsep}{3pt}
\resizebox{\linewidth}{!}{%
\begin{tabular}{lcccc}
\toprule
\textbf{Setting} & \textbf{RoboSpatial} & $\Delta$ & \textbf{Omni3D} & $\Delta$ \\
\midrule
\rowcolor{gray!20}\textbf{SpatialCraft (full)} & 70.9{\scriptsize$\pm$0.6} & -- & 46.0{\scriptsize$\pm$1.4} & -- \\
\midrule
\quad w/o Experiences & 61.3{\scriptsize$\pm$0.9} & \Down{9.6} & 40.5{\scriptsize$\pm$0.6} & \Down{5.5} \\
\quad w/o Skills & 58.9{\scriptsize$\pm$0.6} & \Down{12.0} & 38.8{\scriptsize$\pm$0.8} & \Down{7.2} \\
\midrule
\quad w/o Anchoring (All global) & 65.7{\scriptsize$\pm$1.0} & \Down{5.2} & 42.4{\scriptsize$\pm$0.8} & \Down{3.6} \\
\quad w/o Consolidation & 67.8{\scriptsize$\pm$0.7} & \Down{3.1} & 43.6{\scriptsize$\pm$1.1} & \Down{2.4} \\
\quad CUS $\rightarrow$ LLM-based pruning & 63.4{\scriptsize$\pm$1.1} & \Down{7.5} & 42.0{\scriptsize$\pm$1.1} & \Down{4.0} \\
\quad CUS $\rightarrow$ Naive score & 66.2{\scriptsize$\pm$1.2} & \Down{4.7} & 42.9{\scriptsize$\pm$1.0} & \Down{3.1} \\
\midrule
\quad w/o Skill Evolution (Seeds only) & 61.9{\scriptsize$\pm$0.9} & \Down{9.0} & 40.8{\scriptsize$\pm$0.8} & \Down{5.2} \\
\quad w/o Decomposition Retrieval & 68.6{\scriptsize$\pm$0.6} & \Down{2.3} & 44.9{\scriptsize$\pm$0.6} & \Down{1.1} \\
\quad w/o Experience Rewrite & 67.4{\scriptsize$\pm$1.1} & \Down{3.5} & 44.1{\scriptsize$\pm$1.0} & \Down{1.9} \\
\bottomrule
\end{tabular}
}
\vspace{-1em}
\end{table}

\subsection{Main Results}
\myfinding{SpatialCraft consistently helps agents translate tool access into effective spatial reasoning} As shown in Table~\ref{tab:main}, SpatialCraft improves macro-averaged accuracy over Tool-Only by 7.6–22.6 percentage points across four backbones, and exceeds the strongest baseline SMA by 2.3–4.2 points. This benefit is especially evident with Qwen3.5-9B: tool access alone lowers average accuracy from 52.4\% to 50.4\%, whereas SpatialCraft achieves 59.0\%. These results suggest that reusable experience and procedural skills provide useful guidance for exploiting spatial tools, including when the backbone struggles to benefit from tool access alone.

\myfinding{SpatialCraft is particularly effective when spatial judgments depend on 3D geometry and reference frames} On Omni3D, which requires reasoning about three-dimensional quantities and relations, SpatialCraft reaches 50.7\% and 46.0\% accuracy with GPT-5.4 and GPT-5.4-mini, exceeding SMA by 8.3 and 7.6 percentage points. On ViewSpatial, which evaluates localization from different perspectives, GPT-5.4-mini achieves 62.6\%, compared with SMA's 55.9\%. Together, these gains are consistent with the role of learned guidance in linking spatial evidence to the geometric and viewpoint constraints posed by the question. Appendix~\ref{app:case study} displays case examples that trace step-by-step tool use on queries alongside baseline trajectories.



\subsection{What Makes SpatialCraft Work?}
We ablate the component contributions of SpatialCraft as shown in Table~\ref{tab:ablation}, disabling knowledge streams at deployment and individual mechanisms during accumulation.

\myfinding{A tactic is only as useful as its timing}
Skills contribute more than Experiences (12.0 vs.\ 9.6 points on RoboSpatial; 7.2 vs.\ 5.5 on Omni3D-Bench), yet Experiences matter mostly through their placement: treating them as global erases 5.2/3.6 points, over half of their total contribution. The same tactic shown at task start is often absent from the step that needs it, which is precisely what anchoring fixes.

\myfinding{What the agent keeps matters more than how it retrieves}
Curation mechanisms dominate: replacing CUS with LLM-based pruning costs 7.5/4.0 points, freezing Skills at their seeds costs 9.0/5.2, and disabling consolidation costs 3.1/2.4. Even an outcome-based score falls short: rating Experiences by naive success rate, with exposure still randomized, costs 4.7/3.1 points; without the within-task contrast, easy tasks are mistaken for useful tactics (see the correlation analysis in App.~\ref{app:cus_diag}). Decomposition retrieval and rewriting together account for at most 3.5 points.











\begin{table*}[ht]
\renewcommand{\arraystretch}{1.2}
\setlength{\tabcolsep}{3pt}
\centering

\caption{
\textbf{Main Results on Memory Reuse and Efficiency.} 
Results are reported as $\text{Mean}_{\pm \text{Std Dev}}$, with GPT-5.4-mini as the source backbone.
Unit types in ``Avg Tokens Per Unit'' denote T: Trajectory, P: Procedural memory card, S: Skill, E: Condition-action experience. 
}
\label{tab:reuse_efficiency_final}
\vspace{-0.8em}
\resizebox{\linewidth}{!}{%
\begin{tabular}{l c cc cc cccc c}
\toprule
\textbf{Method} & \multicolumn{5}{c}{\textbf{Skill and Experience Reuse Rate $\%$} ($\uparrow$)} & \multicolumn{4}{c}{\textbf{Efficiency Metrics}} \\
\cmidrule(lr){2-6} \cmidrule(lr){7-10}

& \textbf{In-domain} 
& \multicolumn{2}{c}{\textbf{Cross-task Reuse}} 
& \multicolumn{2}{c}{\textbf{Cross-agent Reuse}} 
& \multicolumn{2}{c}{\textbf{Storage Cost} ($\downarrow$)} 
& \multicolumn{2}{c}{\textbf{Execution Cost}}  
& \textbf{Units}\\

\cmidrule(lr){2-2} 
\cmidrule(lr){3-4} 
\cmidrule(lr){5-6} 
\cmidrule(lr){7-8} 
\cmidrule(lr){9-10}  
\cmidrule(lr){11-11}

& \makecell{Robospatial} 
& \makecell{Omni-3D} 
& \makecell{SAT} 
& \makecell{\texttt{Qwen3.5-9B}} 
& \makecell{\texttt{Qwen3.6-27B}} 
& \makecell{Total Stored\\Tokens} 
& \makecell{Avg Tokens\\per Unit} 
& \makecell{$\Delta$ Prompt\\Tokens/Step} 
& \makecell{Retrieval\\Ratio ($\%$)} \\
\midrule

\textbf{RAG} 
& 38.7$_{\pm 4.8}$ 
& 34.3$_{\pm 2.1}$ 
& 36.8$_{\pm 4.4}$ 
& 13.9$_{\pm 0.7}$ 
& 18.3$_{\pm 0.6}$ 
& 159,600 
& 912$_{\pm 248}$ (T) 
& 2744.8$_{\pm 7.2}$ 
& 100$_{\pm 0.0}$ 
& 175\\

\textbf{MemP}      
& 55.2$_{\pm 3.9}$ 
& 41.7$_{\pm 5.2}$ 
& 39.5$_{\pm 2.7}$ 
& 28.1$_{\pm 1.4}$ 
& 34.9$_{\pm 1.4}$ 
& 22,464 
& 208$_{\pm 53}$ (P) 
& 601.8$_{\pm 8.8}$ 
& 18.6$_{\pm 0.7}$ 
& 108\\

\textbf{SMA}        
& 61.1$_{\pm 2.9}$ 
& 53.1$_{\pm 7.6}$ 
& 47.4$_{\pm 6.0}$ 
& 37.3$_{\pm 4.0}$ 
& 46.1$_{\pm 3.1}$ 
& 39,200 
& 224$_{\pm 56}$ (P) 
& 632.6$_{\pm 4.9}$ 
& 19.9$_{\pm 0.8}$ 
& 175\\

\midrule
\rowcolor[gray]{0.95}
\textbf{SpatialCraft} 
& 
& 
& 
& 
& 
& 
& 
& 
& 
& 
\\

\quad \textbf{-(Skill)} 
& 94.4$_{\pm 4.8}$ 
& 77.8$_{\pm 4.8}$ 
& 80.6$_{\pm 4.8}$ 
& 66.7$_{\pm 0.0}$ 
& 75.0$_{\pm 0.0}$ 
& 2,580 
& 215$_{\pm 36}$ (S) 
& 209.3$_{\pm 2.2}$ 
& 64.7$_{\pm 1.7}$ 
& 12\\

\quad \textbf{-(Experience)} 
& 85.2$_{\pm 3.4}$ 
& 72.6$_{\pm 6.4}$ 
& 68.9$_{\pm 3.8}$ 
& 73.3$_{\pm 2.2}$ 
& 71.9$_{\pm 3.4}$ 
& 3,510 
& 78$_{\pm 19}$ (E) 
& 346.6$_{\pm 7.2}$ 
& 21.7$_{\pm 0.8}$ 
& 45\\

\bottomrule
\end{tabular}
}
\vspace{-1em}
\end{table*}

\begin{figure}[!t]
    \centering
    \includegraphics[width=\linewidth]{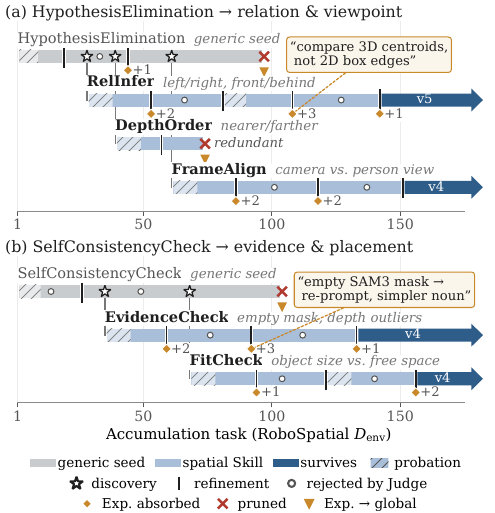}
    \vspace{-2em}
    \caption{Evolutionary lineage of Skills on RoboSpatial.}
    \label{fig:skillevolve}
    \vspace{-1em}
\end{figure}

\subsection{How Does Knowledge Evolve?}
\label{sec:evolve}

We track how knowledge evolves during accumulation on RoboSpatial with GPT-5.4-mini, ablating one mechanism at a time with others fixed. For Skill evolution (Fig.~\ref{fig:training}a), \textbf{w/o Semantic Gradients} replaces diagnostic gradients with neutral trajectory summaries, \textbf{w/o Candidate Verification} adopts a randomly chosen candidate without Judge comparison, and \textbf{w/o Online Score} replaces score-based pruning with first-in-first-out (FIFO) eviction; the x-axis counts Skill Pool update rounds. For Experience evolution (Fig.~\ref{fig:training}b), \textbf{w/o Visual Grounding} removes images from rollout summaries, \textbf{w/o Cross-Rollout Critique} extracts Experiences from each rollout independently, and \textbf{w/o CUS} replaces utility-driven retirement and graduation with LLM-based pruning; the x-axis counts processed samples. Both panels report an efficiency-adjusted return $\tilde{R}(\tau)=R(\tau)\,[1-\kappa\,C(\tau)]$, where $C(\tau)\in[0,1]$ is the budget-normalized tool-call count and $\kappa=0.2$; learning still uses the binary $R(\tau)$. Owing to Experiences withheld half the rollouts, these returns fall below deployment accuracy.
\begin{figure}[!t]
    \centering
    \includegraphics[width=\linewidth]{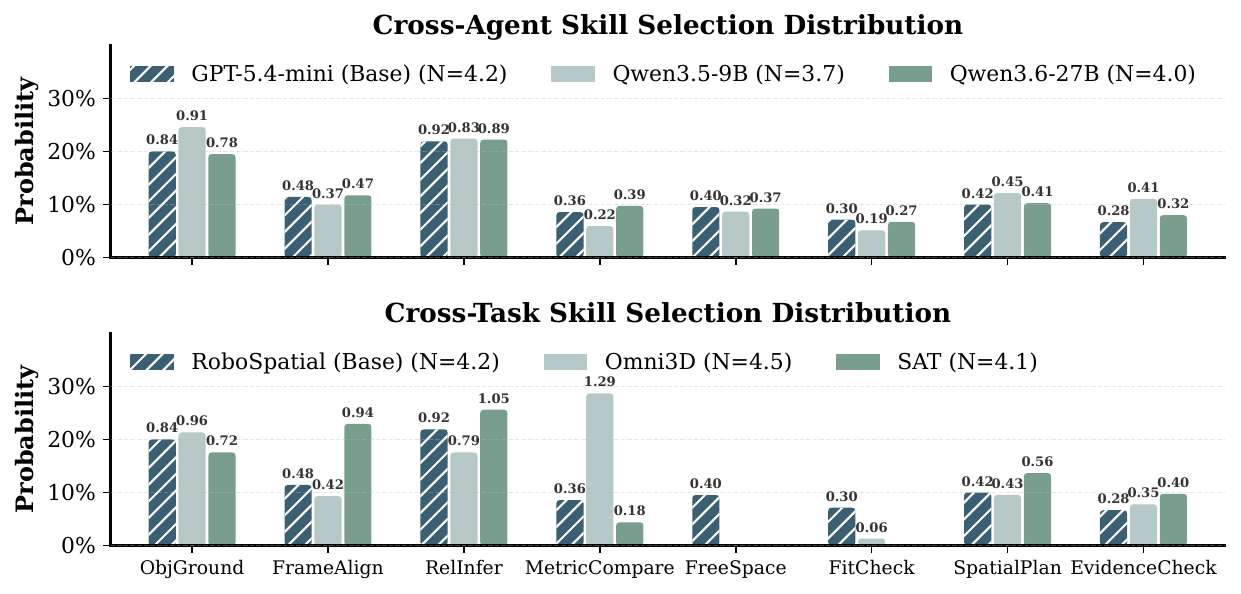}
    \vspace{-2em}
    \caption{Skill distribution across LLM agents and benchmarks.}
    \label{fig:skilldistribution}
    \vspace{-1em}
\end{figure}
\myfinding{Sustained skill improvement requires both reliable refinement and selective retention} Removing Semantic Gradients limits later gains, while removing Candidate Verification produces recurrent setbacks. FIFO eviction repeatedly discards recovered Skills, proving why candidate quality and knowledge retention require separate mechanisms. Fig.~\ref{fig:skillevolve} traces how generic seeds spawn spatial Skills that are refined with absorbed Experiences or pruned.


\myfinding{Effective experience accumulation depends on extracting useful lessons and filtering what is retained} Removing Visual Grounding or Cross-Rollout Critique limits later improvement, whereas removing CUS leads to regression after initial gains: extraction determines what is learned, while utility estimation determines what is kept.

\myfinding{Additional rollouts help most when the backbone can exploit within-task contrast}
Fig.~\ref{fig:training}(c) varies the number of rollouts $N$, where $N=0$ denotes the agent with seed Skills only, and deployment still uses a single run per task. From $N=0$ to $N=4$, Qwen3.6-27B and GPT-5.4-mini improve by 6.3 and 13.7 points with Qwen3.5-9B gains only 1.7. For GPT-5.4-mini, the largest jump occurs from $N=1$ to $N=2$, the smallest $N$ at which an Experience can be both shown and withheld within a task, so CUS starts filtering.

\begin{figure}[!t]
    \centering
    \includegraphics[width=\linewidth]{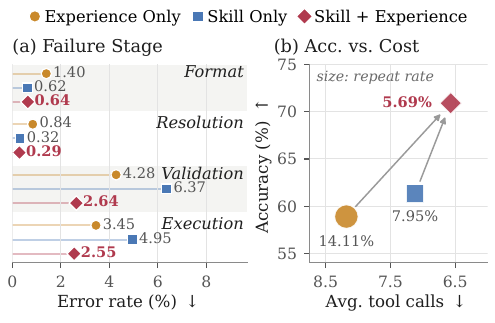}
    \vspace{-2em}
    \caption{Error analysis on RoboSpatial with GPT-5.4-mini.}
    \label{fig:erroranalysis}
    \vspace{-1em}
\end{figure}
\subsection{Does SpatialCraft Truly Enable Reusability?}
\label{sec:reuse_efficiency}

We next ask whether the accumulated knowledge remains useful once frozen and moved beyond the setting in which it was learned. We take the knowledge base accumulated by \texttt{GPT-5.4-mini} on RoboSpatial and deploy it unchanged in three settings: in-domain on RoboSpatial, cross-task on Omni3D and SAT, and cross-agent with \texttt{Qwen3.5-9B} and \texttt{Qwen3.6-27B} as executors. The reuse rate is the fraction of stored units invoked at least once during evaluation.

\myfinding{Memory transfer depends on both its representation and the executor}
Table~\ref{tab:reuse_efficiency_final} shows that the accumulated knowledge transfers beyond its source setting. Skills reach 77.8--80.6\% cross-task reuse, and the transferred knowledge improves accuracy in all four transfer settings, recovering most of the target-accumulated gain (Appendix~\ref{app:transfer}). Fig.~\ref{fig:skilldistribution} shows which Skills carry over: \texttt{ObjGround} and \texttt{RelInfer} stay prominent across agents, while \texttt{MetricCompare} dominates on Omni3D and \texttt{FrameAlign} rises on SAT. The preferred form depends on the backbones: Qwen3.5-9B relies more on Experiences, and Qwen3.6-27B more on Skills. Since neither form consistently dominates, retaining both is warranted.

\myfinding{Compact memory can support frequent adaptation without large prompt overhead}
SpatialCraft stores far fewer tokens and adds fewer prompt tokens per step than MemP and SMA, yet activates Skills more often. The key distinction is how often memory is accessed versus how much content enters each decision: selecting a concise active Skill accommodates changing states while bounding the injected context, so a high retrieval ratio here reflects frequent access to compact units rather than inefficiency.

\subsection{Error Analysis}






\begin{figure}[!t]
    \centering
    \includegraphics[width=\linewidth]{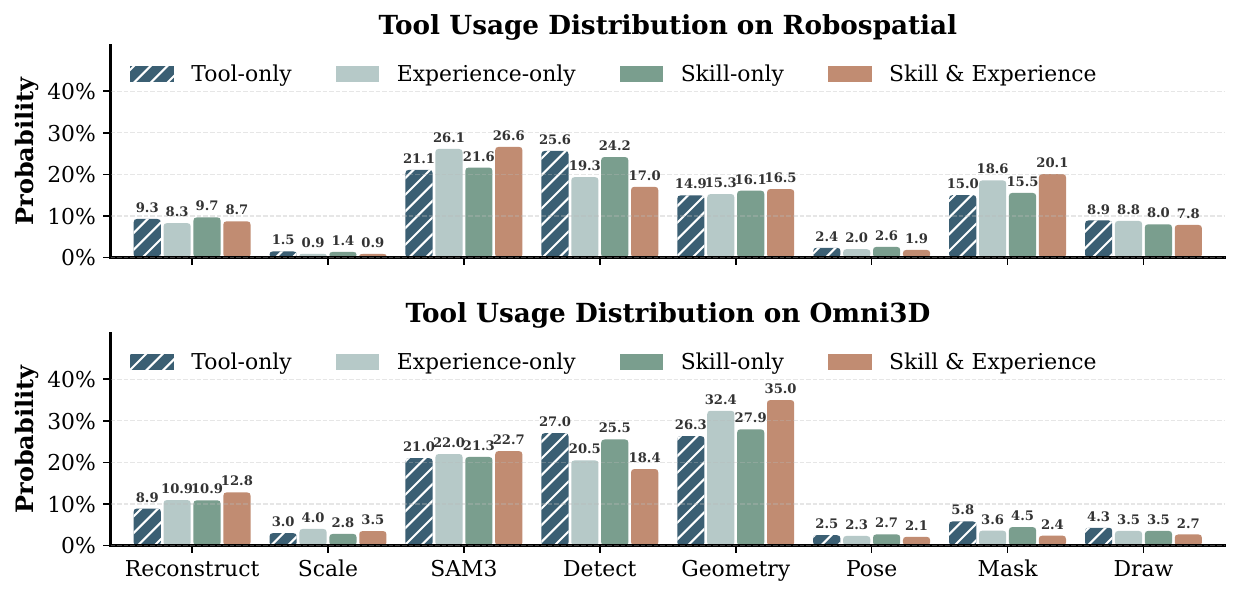}
    \vspace{-2em}
    \caption{Tool usage distribution on Robospatial and Omni3D.}
    \label{fig:tooldistribution}
    \vspace{-1em}
\end{figure}

To identify what each form of knowledge contributes, we analyze tool calls on RoboSpatial with \texttt{GPT-5.4-mini} (Fig.~\ref{fig:erroranalysis}). Each invocation attempt is assigned to the first check it fails: \textbf{Format} (malformed call), \textbf{Resolution} (unknown tool name), \textbf{Validation} (inputs rejected before execution), or \textbf{Execution} (failure after validation), with rates normalized by all attempts (details in Appendix~\ref{app:error_analysis}).


\myfinding{Skills organize invocation, while Experiences adapt it to the current input} Skill Only makes fewer format and resolution errors than Experience Only but more validation and execution failures, as Skills structure the call sequence yet leave input-specific requirements unresolved. Adding Experiences to Skills leaves format errors essentially unchanged but roughly halves validation and execution failures. Fig.~\ref{fig:erroranalysis}(b) shows the downstream effect: from Experience Only to Skill Only to Skill + Experience, accuracy rises while both tool calls and repeated calls fall.

\subsection{Experiences Enable Flexible Orchestration}
\label{sec:orchestration}
We investigate whether Experiences also change \emph{which} tools the agent calls, comparing tool-usage shares on RoboSpatial and Omni3D with \texttt{GPT-5.4-mini}. Shares count all task-execution calls with valid tool names, excluding internal knowledge operations (More in Appendix~\ref{app:addexp}).



\myfinding{Experiences steer tool selection toward what each task requires}
Adding Experiences shifts tool usage in a task-dependent direction (Fig.~\ref{fig:tooldistribution}). On RoboSpatial, the combined share of SAM3 and Mask rises from 36.1\% to 44.7\% without Skills and from 37.1\% to 46.7\% with them, while Detect declines, favoring precise object regions over bounding boxes. On Omni3D, Geometry rises from 26.3\% to 32.4\% and from 27.9\% to 35.0\%, along with Reconstruct, favoring explicit 3D computation. Since the shift appears regardless of Skills, it stems from the Experiences themselves.

\section{Related Work}
\label{sec:related}
\paragraph{Spatial reasoning with MLLMs.}
Spatial reasoning in MLLMs has been advanced by post-training on spatial data or geometric environments~\cite{chen2024spatialvlmendowingvisionlanguagemodels,cai2025spatialbotprecisespatialunderstanding,cheng2024spatialrgptgroundedspatialreasoning,ray2025satdynamicspatialaptitude,li2026spatialevoselfevolvingspatialintelligence,zhao2026ouroborosspatialclosingdatamodelloop}, and by tool-augmented agents that invoke 3D reconstruction and geometry modules to compute spatial quantities explicitly~\cite{marsili2025visualagenticai,chen2026spacetoolstoolaugmentedspatialreasoning,dai2026sagentspatialtooluseelicits,cho2026spatialclawrethinkingactioninterface,luo2026pyspatialvisualprograms,zhou2026spatialclireasonspatialtools}. Most tool-augmented agents remain stateless, discarding lessons from one task to the next.

\vspace{-1em}
\paragraph{Self-evolving agents.}
Parameter-update-free agents instead accumulate external knowledge from interaction, as reflective insights~\cite{shinn2023reflexion,zhao2024expel,ouyang2026reasoningbank,tang2025agent}, reusable workflows~\cite{wang2024agent,fang2026mempexploringagentprocedural}, or procedural skills~\cite{wang2023voyageropenendedembodiedagent,zheng2025skillweaver,mi2026skill}, and recent work combines action-level experiences with task-level skills~\cite{jiang2026xskillcontinuallearningexperience}. To decide what to retain, these systems rely on LLM judgment~\cite{jiang2026xskillcontinuallearningexperience}, token-likelihood ratios~\cite{mi2026skill}, or the rewards of episodes in which an entry is retrieved~\cite{zhang2026memrlselfevolvingagentsruntime,zhang2026spatial}; the last is confounded by task difficulty, as entries retrieved mostly for easy tasks appear useful regardless of their effect. A full discussion of related work is given in Appendix~\ref{app:fullrelatedwork}.
\section{Conclusion}
\label{sec:conclusion}

We presented \textsc{SpatialCraft}, which lets a frozen spatial agent improve tool orchestration by anchoring Experiences to the Skills they support, validating them through randomized within-task contrast, and folding proven ones into those Skills. Empirically, it achieves the best accuracy in all 20 benchmark-backbone settings. Our results suggest that, as tools grow more capable, the decisive knowledge lies less in what an agent knows than in knowing when to apply it, and in keeping only what has been shown to help.


{
    \small
    \bibliographystyle{ieeenat_fullname}
    \bibliography{main}
}
\appendix
\clearpage
\setcounter{page}{1}
\maketitlesupplementary


\begin{table}[h]
\centering
\footnotesize
\setlength{\tabcolsep}{3pt}
\caption{Main notation.}
\vspace{-1em}
\label{tab:notation}
\begin{tabular}{@{}p{0.27\linewidth}p{0.69\linewidth}@{}}
\toprule
\textbf{Symbol} & \textbf{Meaning} \\
\midrule
$\mathcal{T}=(q,\mathcal{I})$ & Task: spatial query and image inputs \\
$\mathcal{M}_{\text{exec}}$, $\mathcal{M}_{\text{kb}}$ & Frozen executor; knowledge-operation instance \\
$\mathcal{F}$ & Tool set \\
$R(\tau)=\mathbb{1}[\hat y=y^*]$ & Binary reward of trajectory $\tau$ \\
$\mathcal{D}_{\text{env}}$, $\mathcal{D}_{\text{dep}}$ & Accumulation and deployment splits \\
\midrule
$k=(\omega_k,\pi_k,\beta_k)$ & Skill: activation, procedure, termination \\
$e=(c,a,\alpha_e,\mathbf{v}_e)$ & Experience: condition, action, anchor, embedding \\
$\mathcal{E}_\varnothing$ & Global Experiences ($\alpha_e=\varnothing$) \\
$\mathcal{E}^{\text{ret}}$, $\mathcal{E}^{\text{anc}}_k$ & Retrieved global / anchored Experiences of $k$ \\
\midrule
$N$ & Rollouts per accumulation task \\
$M_i(e)$ & Exposure subset of $e$ on task $i$, $|M_i(e)|=\lfloor N/2\rfloor$ \\
$X_i^{+}(e)$, $X_i^{-}(e)$ & Eligible rollouts where $e$ is shown / withheld \\
$A_i(k)$, $\bar A_i(k)$ & Rollouts where $k$ is / is not activated \\
$\delta_i(\cdot)$ & Within-task contrast \\
$(S,n)$ & Accumulated contrast and number of contrasts \\
$U(e)$, $\mathrm{Score}(k)$ & Shrunk utility, $S/(\lambda+n)$ \\
$\lambda$, $n_{\min}$, $\tau_g$ & Shrinkage, probation length, graduation threshold \\
\midrule
$g_i$, $\bar g_k$ & Semantic gradient; aggregated gradient \\
$\mathcal{G}_k$ & Graduated Experiences anchored to $k$ \\
$\mathcal{C}(k)$, $N_c$ & Candidate revisions of $k$ and their number \\
$j_i(k')$, $J(k')$ & Per-state judge score; mean over $B$ states \\
$B$, $N_p$ & Trajectories per revision; Skills revised per round \\
\bottomrule
\end{tabular}
\vspace{-1em}
\end{table}

\section{Methodology}
\label{app:method}
\subsection{Accumulation Algorithm}
\label{app:algorithm}
Algorithm~\ref{alg:accumulation} summarizes accumulation on $\mathcal{D}_{\text{env}}$. Utility statistics are updated before the critique, so that the modification rule uses the latest $U(e)$.

\begin{algorithm}[t]
\caption{Knowledge accumulation}
\label{alg:accumulation}
\footnotesize
\begin{algorithmic}[1]
\Require $\mathcal{D}_{\text{env}}$, frozen $\mathcal{M}_{\text{exec}},\mathcal{M}_{\text{kb}}$, seeds $\mathcal{K}_0$
\State $\mathcal{K}\gets\mathcal{K}_0$,\; $\mathcal{E}\gets\varnothing$
\For{each task $\mathcal{T}_i\in\mathcal{D}_{\text{env}}$}
  \State $\mathcal{E}^{\text{ret}}\gets$ Rewrite(DecompRetrieve($\mathcal{T}_i,\mathcal{E}_\varnothing$)) \Comment{Eq.~(8)}
  \State Run $N$ rollouts: select $k_t$ at boundaries (Eq.~(9)), inject $\mathcal{E}^{\text{anc}}_{k_t}$; show eligible $e$ in rollout $j$ iff $j\in M_i(e)$
  \For{each $e$ with $X_i^{+}(e),X_i^{-}(e)\neq\varnothing$}
    \State $S_e \mathrel{+}= \delta_i(e)$;\; $n_e \mathrel{+}= 1$ \Comment{Eq.~(3)}
  \EndFor
  \For{each $k$ with $A_i(k),\bar A_i(k)\neq\varnothing$}
    \State $S_k \mathrel{+}= \delta_i(k)$;\; $n_k \mathrel{+}= 1$ \Comment{Eq.~(7)}
  \EndFor
  \State $\mathcal{S}_i\gets$ Summary; $\Delta\mathcal{E}_i\gets$ Critique \Comment{Eq.~(2)}
  \State Apply $\Delta\mathcal{E}_i$ with anchors; merge duplicates
  \State Evict by ascending $U$ if $|\mathcal{E}|>N^{\mathcal{E}}_{\max}$
  \State Remove retired $e$; collect graduated into $\mathcal{G}_k$
  \For{up to $N_p$ eligible Skills $k$}
    \State $\bar g_k\gets\mathrm{Agg}(\{g_i\}_{i=1}^{B})$ \Comment{Eq.~(4)}
    \State $\mathcal{C}(k)\gets\{k\oplus(\bar g_k,\mathcal{G}_k)\}^{N_c}$ \Comment{Eq.~(5)}
    \If{$\max_{k'\in\mathcal{C}(k)}J(k')>0$} \Comment{Eq.~(6)}
      \State Replace $k$ by the best $k'$; remove absorbed $\mathcal{G}_k$ from $\mathcal{E}$
    \EndIf
  \EndFor
  \State Prune $\mathcal{K}$; set $\alpha_e\gets\varnothing$ for $e$ anchored to removed Skills
\EndFor
\State \Return frozen $\mathcal{KB}=(\mathcal{K},\mathcal{E})$
\end{algorithmic}
\end{algorithm}

\subsection{Contrastive Utility Estimation}
\label{app:cus}

\paragraph{Lazy exposure assignment.}
An anchored Experience becomes eligible only when its Skill is activated, which cannot be known before the rollouts start. We therefore fix $M_i(e)$ in advance by seeding a pseudo-random permutation of $\{1,\dots,N\}$ with a hash of the task and Experience identifiers, and look it up only when $e$ becomes eligible in rollout $j$. A global Experience is eligible in every rollout once it is retrieved and kept by Experience Rewrite, which outputs one entry per Experience identifier so that each can be withheld individually.

\paragraph{Why the contrast is unbiased.}
Whether $e$ becomes eligible is decided before $e$ is shown, so it cannot depend on $e$'s own exposure. Conditioned on eligibility, exposure is thus independent of how the rollout would otherwise unfold, and $\mathbb{E}[\delta_i(e)]$ equals the effect of showing $e$ on task $i$. Because Experiences are randomized independently, this effect is averaged over the exposure of co-occurring Experiences: $U(e)$ estimates the marginal effect of $e$ when other Experiences are present about half of the time, rather than under the full exposure used at deployment.

\paragraph{Probation under capacity limits.}
When $\mathcal{E}$ exceeds $N^{\mathcal{E}}_{\max}$, Experiences past probation are evicted first in ascending order of $U$; probationary entries are evicted only if no others remain. This prevents a new Experience from being removed on the basis of its first few noisy contrasts.

\paragraph{Small $N$.}
With $N=1$, $\lfloor N/2\rfloor=0$ and no task yields a contrast. In the rollout-count study (Sec.~\ref{sec:evolve}), eligible Experiences are therefore always shown when $N=1$, and all remain in probation.

\begin{table*}[t]
\centering
\footnotesize
\setlength{\tabcolsep}{5pt}
\renewcommand{\arraystretch}{1.05}
\setlength{\aboverulesep}{0pt}\setlength{\belowrulesep}{0pt}
\caption{Initial skill seeds with their activation conditions $\omega_k$,
termination conditions $\beta_k$, and execution procedures $\pi_k$.}
\label{tab:skill_seeds}
\begin{tabularx}{\textwidth}{
    >{\raggedright\arraybackslash\Tpad}p{0.115\textwidth}<{\Bpad}
    >{\raggedright\arraybackslash}p{0.215\textwidth}<{\Bpad}
    >{\raggedright\arraybackslash}p{0.175\textwidth}<{\Bpad}
    >{\raggedright\arraybackslash}X<{\Bpad}}
\toprule
\textbf{Skill seed} &
\textbf{Activation} $\boldsymbol{\omega_k}$ &
\textbf{Termination} $\boldsymbol{\beta_k}$ &
\textbf{Execution procedure} $\boldsymbol{\pi_k}$ \\
\midrule
\texttt{Structured CoT}
& When a decision requires reasoning over multiple constraints or
  previously observed feedback.
& A concrete action satisfying the identified constraints has been selected.
& \begin{Steps}
  \item Identify the immediate objective of the current decision.
  \item Enumerate the constraints implied by the current state and
        interaction history.
  \item Summarize previous observations and feedback relevant to the decision.
  \item Compare candidate actions according to these constraints.
  \item Select the action that best satisfies the current objective.
  \end{Steps} \\
\rowcolor{rowalt}
\texttt{ReAct Decision}
& When new environmental feedback becomes available and should affect
  the next decision.
& The next action has been selected based on the newly incorporated
  feedback.
& \begin{Steps}
  \item Interpret the most recent environmental feedback.
  \item Update the current belief about plausible states or outcomes.
  \item Determine whether additional evidence is required.
  \item Select the next action according to the updated belief.
  \end{Steps} \\
\texttt{Hypothesis Elimination}
& When multiple competing hypotheses remain plausible under the
  available evidence.
& An action targeting the unresolved hypothesis uncertainty has been
  selected.
& \begin{Steps}
  \item Enumerate the major hypotheses consistent with existing observations.
  \item Remove hypotheses that contradict known evidence.
  \item Identify the uncertainty that distinguishes the remaining hypotheses.
  \item Select an action that is expected to reduce this uncertainty.
  \end{Steps} \\
\rowcolor{rowalt}
\texttt{Self-Consistency Check}
& When a candidate decision must satisfy known constraints or remain
  consistent with previous observations.
& A candidate passes all identified consistency checks.
& \begin{Steps}
  \item Generate a candidate action or conclusion.
  \item Check the candidate against known constraints and previous feedback.
  \item Revise the candidate if any inconsistency is detected.
  \item Repeat the verification until no clear contradiction remains.
  \end{Steps} \\
\texttt{Explore--Exploit Arbitration}
& When several plausible actions are available and the agent must
  balance information gathering against exploiting existing evidence.
& An action has been selected according to an explicit exploration or
  exploitation decision.
& \begin{Steps}
  \item Assess the current uncertainty and the information gained from
        recent actions.
  \item Favor exploration when uncertainty remains high or recent
        information gain is insufficient.
  \item Favor exploitation when accumulated evidence strongly supports a
        particular action.
  \item Select the action corresponding to the chosen
        exploration--exploitation strategy.
  \end{Steps} \\
\rowcolor{rowalt}
\texttt{Strategic Planning}
& At the beginning of a task, when little or no task-specific
  interaction feedback is available.
& An initial strategy has been instantiated and the first informative
  feedback becomes available.
& \begin{Steps}
  \item Identify the overall task objective and available action space.
  \item Establish an initial hypothesis or search space from the task
        constraints.
  \item Select an informative initial strategy that can reduce uncertainty.
  \end{Steps} \\
\bottomrule
\end{tabularx}
\vspace{-0.6em}
\end{table*}

\subsection{Skill Evolution Details}
\label{app:skill_details}

\paragraph{Trigger and batch composition.}
A Skill becomes eligible for revision once it has been active in $B$ new trajectories since its last revision. Each round revises at most $N_p$ eligible Skills, prioritizing those with the most pending trajectories. The $B$ trajectories combine the lowest- and highest-return halves, so that semantic gradients reflect both failure modes and successful practice.

\paragraph{Revision acceptance.}
An accepted refinement replaces $k$ and inherits $(S_k,n_k)$ under the Experience modification rule: the statistics are kept if $\mathrm{Score}(k)>0$ and reset otherwise, in which case the revision re-enters probation. A Skill synthesized in discovery mode covers a situation that $k$ does not, so it is added alongside $k$ with $S=n=0$. Graduated Experiences in $\mathcal{G}_k$ are removed from the bank only after the revision that absorbs them is accepted; otherwise they remain as anchored Experiences.

\subsection{Skill Seeds}
\label{app:seeds}
We adopt the six general-purpose seeds of Skill-Pro~\cite{mi2026skill} (Table~\ref{tab:skill_seeds}). They encode domain-agnostic reasoning procedures and contain no spatial or tool-specific knowledge. Spatially specialized Skills, such as \texttt{ObjGround} and \texttt{MetricCompare} in Fig.~\ref{fig:skilldistribution}, emerge from these seeds through refinement and discovery.

\section{Experimental Setup}
\label{app:setup}

\subsection{Benchmarks and Splits}
\label{app:benchmark}

\paragraph{Benchmarks.}
\textbf{RoboSpatial}~\cite{song2026robospatialteachingspatialunderstanding} poses three question types on single indoor images: \emph{context} questions ask for image points marking free space that satisfies a spatial relation, while \emph{configuration} and \emph{compatibility} questions ask yes/no whether two objects stand in a given relation or whether an object fits a target location.
\textbf{ERQA}~\cite{geminirobotics2025bringingphysicalworld} contains multiple-choice questions on robot-centric scenes, spanning spatial relations, state estimation, and action and trajectory reasoning, with one or more images per question.
\textbf{Omni3D-Bench}~\cite{marsili2025visualagenticai} builds open-ended queries on single real images that depend on 3D quantities, such as metric size, distance ratios, and 3D-conditioned counting; answers are numbers, yes/no, or short phrases.
\textbf{SAT}~\cite{ray2025satdynamicspatialaptitude} asks two-option questions over one or two frames about ego motion, object motion, action consequences, goal aiming, and perspective taking.
\textbf{ViewSpatial}~\cite{li2025viewspatialbenchevaluatingmultiperspectivespatial} asks multiple-choice direction and orientation questions that must be answered either from the camera's viewpoint or from that of a person in the scene.

\paragraph{Splits.}
Except for SAT, we split each benchmark into $\mathcal{D}_{\text{env}}$ and $\mathcal{D}_{\text{dep}}$ by shuffling every question category with seed 42 and assigning half of it to each split; when a category has an odd size, the extra item goes alternately to $\mathcal{D}_{\text{env}}$ and $\mathcal{D}_{\text{dep}}$, starting with $\mathcal{D}_{\text{env}}$. For SAT, $\mathcal{D}_{\text{dep}}$ is the official test set with circular option expansion, and $\mathcal{D}_{\text{env}}$ consists of an equal number of validation items drawn to match its question-type distribution. Table~\ref{tab:split} lists the resulting sizes.

\begin{table}[h]
\centering\footnotesize
\setlength{\tabcolsep}{4pt}
\caption{Split sizes. All results are reported on $\mathcal{D}_{\text{dep}}$.}
\label{tab:split}
\begin{tabular}{lrrr}
\toprule
\textbf{Benchmark} & \textbf{Total} & $|\mathcal{D}_{\text{env}}|$ & $|\mathcal{D}_{\text{dep}}|$ \\
\midrule
RoboSpatial  & 350   & 175   & 175   \\
ERQA         & 400   & 200   & 200   \\
Omni3D-Bench & 501   & 251   & 250   \\
SAT          & 600   & 300   & 300   \\
ViewSpatial  & 5,712 & 2,856 & 2,856 \\
\bottomrule
\end{tabular}
\vspace{-2em}
\end{table}

\paragraph{Scoring.}
Multiple-choice and yes/no answers are scored by exact match after normalizing case and option format. A RoboSpatial context answer is correct if the predicted point lies inside the ground-truth region. A numeric Omni3D-Bench answer is correct if its relative error is at most 10\%, and a short-phrase answer if it matches the reference after normalization. Answers that are missing or cannot be parsed count as incorrect.


\subsection{Implementation and Hyperparameters}
\label{app:hyper}
Table~\ref{tab:hyperparameters} lists all settings. Open-weight backbones run in non-thinking mode for every call, including those of $\mathcal{M}_{\text{kb}}$ and the Judge. If an execution call reaches its token limit without producing a valid action, one shorter recovery call is issued without consuming a step. Once a trajectory reaches the step limit, a final call produces the answer without further tool use.

\begin{table*}[t]
\centering\footnotesize
\setlength{\tabcolsep}{4pt}
\caption{Hyperparameters of \textsc{SpatialCraft}, shared across backbones and benchmarks.}
\vspace{-1em}
\label{tab:hyperparameters}
\begin{tabular}{@{}lll@{}}
\toprule
\textbf{Parameter} & \textbf{Value} & \textbf{Description} \\
\midrule
\multicolumn{3}{@{}l}{\cellcolor{gray!15}\emph{Execution ($\mathcal{M}_{\text{exec}}$)}} \\
Temperature / top-$p$ (accumulation) & 0.7 / 0.9 & Diverse rollouts for contrast \\
Temperature (deployment) & 0 & Greedy decoding \\
Output budget per call & 16,384 & Execution calls \\
Steps per trajectory & 50 & Tool and answer steps \\
Recovery call & $\le$1 per step, 1,024 & Issued after truncation; consumes no step \\
Forced final answer & 512 & No tools after the step limit \\
Image resolution & $\sim$1M pixels & Maximum per input image \\
\midrule
\multicolumn{3}{@{}l}{\cellcolor{gray!15}\emph{Experience accumulation ($\mathcal{M}_{\text{kb}}$)}} \\
Rollouts per task $N$ & 4 & Knowledge fixed within a task's rollouts \\
Rollout summary & Temp.\ 0, 8,192 & Per task \\
Cross-rollout critique & Temp.\ 0, 16,384 & Per task \\
Pairwise merge & 1,024 & Triggered at similarity $>\theta_{\text{sim}}$ \\
Bank-level merge & 8,192 & Triggered when $|\mathcal{E}|>N^{\mathcal{E}}_{\max}$ \\
Experience length & 64 words & Per Experience \\
Merge threshold $\theta_{\text{sim}}$ & 0.70 & Cosine similarity \\
Bank capacity $N^{\mathcal{E}}_{\max}$ & 100 & Experiences \\
\midrule
\multicolumn{3}{@{}l}{\cellcolor{gray!15}\emph{Contrastive utility estimation}} \\
Exposure ratio & $\lfloor N/2\rfloor/N=2/4$ & Shown fraction among eligible rollouts \\
Shrinkage $\lambda$ & 2 & Shared by $U(e)$ and $\mathrm{Score}(k)$ \\
Probation $n_{\min}$ & 3 & Contrasts before retirement, graduation, or pruning \\
Graduation threshold $\tau_g$ & 0.2 & On $U(e)$, anchored Experiences only \\
\midrule
\multicolumn{3}{@{}l}{\cellcolor{gray!15}\emph{Skill evolution ($\mathcal{M}_{\text{kb}}$)}} \\
Trajectories per revision $B$ & 6 & 3 lowest- and 3 highest-return \\
Skills per round $N_p$ & 2 & Most pending trajectories first \\
Candidates per Skill $N_c$ & 3 & Per revision \\
Gradient / generation budget & 8,192 / 16,384 & Per trajectory / per candidate \\
Stored Skill length & 1,024 tokens & Per Skill \\
Judge & Same backbone, temp.\ 0 & Unlabeled, randomly ordered action pairs \\
Acceptance & $J(k')>0$ & Best candidate only \\
Pool capacity $N^{\mathcal{K}}_{\max}$ & 20 & Skills \\
Pending trajectories & Discarded & Not reused after a revision is accepted \\
\midrule
\multicolumn{3}{@{}l}{\cellcolor{gray!15}\emph{Inference (accumulation and deployment)}} \\
Embedding model & \texttt{text-embedding-3-large} & Retrieval, merging, and Skill activation \\
Task decomposition & Once per task, 1,024 & At task start \\
Global retrieval & Top-3 per subtask, $\theta_{\min}=0.40$ & Over $\mathcal{E}_\varnothing$ \\
Experience Rewrite & Temp.\ 0, 2,048 & One entry per Experience identifier \\
Skill activation & Top-1, $\theta_{\text{act}}=0.50$ & At task start and after termination \\
Anchored Experiences & Top-3 per activation & Ranked by similarity to the query \\
Skill horizon $H_{\max}$ & 8 steps & Forced termination \\
Termination check & Temp.\ 0, 256 & Per Skill boundary \\
\bottomrule
\end{tabular}
\end{table*}

\subsection{Baselines}
\label{app:baselines}
All memory-based baselines use the same backbone, $\mathcal{D}_{\text{env}}$/$\mathcal{D}_{\text{dep}}$ splits, $N=4$ accumulation rollouts per task, and tool library as \textsc{SpatialCraft}, and their memories are frozen at deployment. They differ only in what is stored and how it is selected.
\begin{itemize}[leftmargin=*,itemsep=1pt,topsep=2pt]
\item \textbf{No memory (CoT)} answers directly with a chain-of-thought prompt, without tools or memory.
\item \textbf{Tool-Only (ReAct)}~\cite{yao2022react} interleaves reasoning and tool calls without accumulated knowledge, isolating the effect of tool access.
\item \textbf{RAG}~\cite{lewis2021retrievalaugmentedgenerationknowledgeintensivenlp} stores raw accumulation trajectories and retrieves those of the most similar tasks as in-context demonstrations, without distillation or utility estimates.
\item \textbf{MemP}~\cite{fang2026mempexploringagentprocedural} distills trajectories into procedural memories retrieved by semantic similarity, without feedback on their later usefulness.
\item \textbf{MemRL-R}~\cite{zhang2026memrlselfevolvingagentsruntime} attaches a value to each memory, updates it from runtime rewards, and ranks retrieval by similarity and value; its reflection sees only the reward.
\item \textbf{MemRL-GT} is identical except that reflection also sees the ground-truth answer, matching the supervision available to \textsc{SpatialCraft}.
\item \textbf{SMA}~\cite{zhang2026spatial} stores procedural memory cards and re-ranks similarity-filtered candidates by a Transfer Reliability Score, the shrunk mean reward of tasks on which each card was retrieved.
\end{itemize}

\subsection{Evaluation Metrics}
\label{app:metrics}

\paragraph{Accuracy.}
We report the fraction of $\mathcal{D}_{\text{dep}}$ questions answered correctly under the scoring rules of Appendix~\ref{app:benchmark}, together with its macro average over benchmarks.

\paragraph{Reuse.}
Let $\mathcal{M}$ be the stored units of a method: Skills and Experiences for \textsc{SpatialCraft} (reported separately), trajectories for RAG, and memory cards for MemP and SMA. A unit is \emph{used} in an episode if it enters the executor's context: an Experience shown after retrieval or anchoring, a Skill activated, or a baseline memory retrieved. For a target split $\mathcal{D}$ and executor $\mathcal{M}_{\text{exec}}$,
\begin{equation}
\mathrm{Reuse}(\mathcal{D},\mathcal{M}_{\text{exec}})=\frac{|\{u\in\mathcal{M}:u\text{ is used on }\mathcal{D}\text{ by }\mathcal{M}_{\text{exec}}\}|}{|\mathcal{M}|}.
\end{equation}
With $\mathcal{M}$ accumulated by \texttt{GPT-5.4-mini} on RoboSpatial, \emph{in-domain} reuse evaluates RoboSpatial $\mathcal{D}_{\text{dep}}$ with the source executor, \emph{cross-task} reuse evaluates Omni3D-Bench and SAT with the source executor, and \emph{cross-agent} reuse evaluates RoboSpatial with \texttt{Qwen3.5-9B} or \texttt{Qwen3.6-27B}.

\paragraph{Efficiency.}
\emph{Total stored tokens} and \emph{average tokens per unit} measure the size of $\mathcal{M}$, counted with the same tokenizer for all methods. \emph{$\Delta$Prompt tokens per step} is the mean difference, over decision steps, between the executor's prompt length with and without memory content. \emph{Retrieval ratio} is the fraction of decision steps at which memory is accessed, by retrieval or by Skill selection; it reflects access frequency rather than cost and should be read together with $\Delta$Prompt tokens per step.

\section{Analysis Protocols}
\label{app:protocols}
Unless stated otherwise, analyses use \texttt{GPT-5.4-mini}, the settings of Table~\ref{tab:hyperparameters}, and greedy single-run deployment on $\mathcal{D}_{\text{dep}}$. Results are averaged over three independent accumulation runs.

\subsection{Component Ablations}
\label{app:ablation}
Table~\ref{tab:ablation} evaluates RoboSpatial and Omni3D-Bench; each benchmark accumulates on its own $\mathcal{D}_{\text{env}}$. Variants fall into two groups.

\paragraph{Deployment-time variants.}
These reuse the three knowledge bases of the full model and disable one stream only at deployment.
\begin{itemize}[leftmargin=*,itemsep=1pt,topsep=2pt]
\item \textbf{w/o Experiences}: neither global nor anchored Experiences are shown; Skill activation is unchanged.
\item \textbf{w/o Skills}: no Skill is activated. Anchored Experiences join $\mathcal{E}_\varnothing$ and are retrieved and rewritten as global ones.
\end{itemize}

\paragraph{Accumulation-time variants.}
These disable one mechanism, re-run accumulation with the same task order, and deploy the resulting knowledge base.
\begin{itemize}[leftmargin=*,itemsep=1pt,topsep=2pt]
\item \textbf{w/o Anchoring}: the critique sets $\alpha_e=\varnothing$ for every Experience, so all Experiences are retrieved once per task. Because graduation requires an anchor, consolidation is also inactive; this drop therefore upper-bounds the effect of anchoring alone.
\item \textbf{w/o Consolidation}: the lifecycle is unchanged, but $\oplus$ receives $\mathcal{G}_k=\varnothing$, so graduated Experiences stay in the bank as anchored Experiences.
\item \textbf{w/o CUS}: $\mathcal{M}_{\text{kb}}$ decides retirement, graduation, and eviction without outcome contrasts.
\item \textbf{w/o Skill Evolution}: $\Phi$ is disabled and the pool keeps the six seeds. Experiences are still anchored to seeds, and graduated ones remain in the bank.
\item \textbf{w/o Decomposition Retrieval}: global Experiences are retrieved with the full query as a single retrieval query, with the number retrieved matched to the full model's average.
\item \textbf{w/o Experience Rewrite}: retrieved global Experiences are shown verbatim; eligibility for randomized exposure then follows retrieval alone.
\end{itemize}

\subsection{Accumulation Curves}
\label{app:curves}

\paragraph{Shared protocol.}
Both panels of Fig.~\ref{fig:training}(a,b) use $\mathcal{D}_{\text{env}}$ of RoboSpatial. All variants start from the six seed Skills and an empty Experience Bank, use randomized exposure, and process the same shuffled task order. Each configuration is run three times.

\paragraph{Efficiency-adjusted return.}
Let $S(\tau)\in\{0,1\}$ be correctness under the scoring rules of Appendix~\ref{app:benchmark}, with missing answers counted as incorrect. Let $n_{\text{tool}}(\tau)$ count task-execution tool-call attempts, including failed calls, invalid tool names, and retries, but excluding knowledge operations such as retrieval, rewriting, Skill selection, termination checks, and Judge calls. We define
\begin{equation}
C(\tau)=\frac{n_{\text{tool}}(\tau)}{B_{\text{tool}}},\qquad R(\tau)=S(\tau)\,[1-\kappa\,C(\tau)],
\end{equation}
with $B_{\text{tool}}=50$ and $\kappa=0.2$, so correct answers receive returns in $[0.8,1]$. A trajectory that exhausts either budget produces a final answer without tools and is scored normally. This return is used only for reporting; learning uses the binary reward.

\paragraph{Skill-evolution variants.}
Experience accumulation remains enabled under identical rules.
\begin{itemize}[leftmargin=*,itemsep=1pt,topsep=2pt]
\item \textbf{w/o Semantic Gradients}: $\mathcal{M}_{\text{kb}}$ receives neutral summaries of the same $B$ trajectories instead of semantic gradients; $\mathcal{G}_k$ and $N_c$ are unchanged.
\item \textbf{w/o Candidate Verification}: one of the $N_c$ candidates, sampled uniformly, replaces the parent without Judge comparison.
\item \textbf{w/o Online Score}: Skill scores are not used for pruning; when the pool exceeds capacity, the earliest-added Skill is evicted (FIFO). Redundancy removal is unchanged.
\end{itemize}
The x-axis counts Skill Pool update rounds. A round occurs after a task whenever at least one Skill is eligible for revision, including rounds in which no candidate is accepted. The y-axis reports the mean return of all rollouts collected since the previous round.

\paragraph{Experience-evolution variants.}
Skill evolution remains enabled under identical rules, so the resulting Skill Pools may differ as an indirect effect.
\begin{itemize}[leftmargin=*,itemsep=1pt,topsep=2pt]
\item \textbf{w/o Visual Grounding}: the rollout summary receives the task text, actions, and textual tool outputs but no images; the executor and Skill evolution keep their visual inputs.
\item \textbf{w/o Cross-Rollout Critique}: Experiences are extracted from each rollout separately, under the same per-task proposal budget and 64-word limit; anchoring and merging are unchanged.
\item \textbf{w/o CUS}: utility statistics are not computed. After each task, $\mathcal{M}_{\text{kb}}$ labels each Experience shown in it as retained, retired, or graduated from its content and the rollout summaries, and also selects the entries removed when the bank exceeds capacity. Randomized exposure is kept so that all variants see the same exposure pattern.
\end{itemize}
The x-axis counts processed tasks. The y-axis reports the mean return of the $N=4$ rollouts of each task, measured before they are used to update the knowledge base.

\paragraph{Plotting.}
Each run is smoothed by a trailing moving average over ten points, using the available prefix for the first nine. Lines and shaded areas show the mean and standard deviation across the three runs, at indices reached by all of them.

\subsection{Number of Accumulation Rollouts}
\label{app:rollouts}
Fig.~\ref{fig:training}(c) varies $N\in\{0,1,2,3,4\}$ on RoboSpatial for \texttt{GPT-5.4-mini}, \texttt{Qwen3.6-27B}, and \texttt{Qwen3.5-9B}, keeping all other settings fixed. $N=0$ deploys the six seed Skills with an empty bank and no accumulation. With $N=1$, no contrast exists, so eligible Experiences are always shown and none leave probation (Appendix~\ref{app:cus}); capacity ties are then broken by insertion order. Since $B$ counts trajectories across tasks, Skill revision remains possible for every $N\ge1$.

\subsection{Reuse and Efficiency}
\label{app:reuse_protocol}
All methods use the memory accumulated by \texttt{GPT-5.4-mini} on RoboSpatial, frozen and deployed without modification. \emph{Cross-task} runs deploy it on $\mathcal{D}_{\text{dep}}$ of Omni3D-Bench and SAT with the same executor; \emph{cross-agent} runs deploy it on RoboSpatial with \texttt{Qwen3.5-9B} or \texttt{Qwen3.6-27B}, which then also perform the deployment-time knowledge operations (decomposition, rewriting, and termination checks). Token counts use the \texttt{cl100k\_base} tokenizer for all methods, counting the full text of a Skill and the condition and action of an Experience. Efficiency metrics are measured in the in-domain setting. For \textsc{SpatialCraft}, a step counts toward the Skill retrieval ratio if a Skill is selected at it, and toward the Experience ratio if global retrieval or anchored injection occurs at it.

\subsection{Tool-Use Error Analysis}
\label{app:error_analysis}
We deploy the same frozen knowledge base on RoboSpatial. Skill Only disables Experiences; Experience Only disables Skills, so that anchored Experiences are retrieved as global ones, matching the deployment-time variants in Appendix~\ref{app:ablation}.

\paragraph{Attempts and categories.}
An attempt is any model output submitted to the tool-call interface. Attempts are recorded before parsing, final answers are excluded, and each retry counts as a new attempt. A failed attempt is assigned to the first check it fails:
\begin{itemize}[leftmargin=*,itemsep=1pt,topsep=2pt]
\item \textbf{Format}: the output cannot be parsed into a valid call, \eg, malformed JSON, missing call fields, or several calls in one step.
\item \textbf{Resolution}: the call is well formed but names an unregistered tool. Choosing an existing but unsuitable tool is not counted.
\item \textbf{Validation}: the tool resolves, but its arguments fail pre-execution checks, \eg, missing, ill-typed, or out-of-range arguments, or references to nonexistent artifacts.
\item \textbf{Execution}: the call passes all checks but the tool raises an exception, times out, or returns an explicit failure. A valid empty result, such as no detected object, is not a failure.
\end{itemize}
Each rate divides the number of failures in its category by the total number of attempts, so all four categories share one denominator. Failures outside the tools, such as API errors and job interruptions, are excluded.

\paragraph{Efficiency.}
\emph{Average tool calls} is the number of attempts per task. The \emph{exact repeat rate} is the fraction of attempts that match an earlier attempt in tool and normalized arguments on the same input state, with no relevant state change in between.

\section{Additional Experimental Results}
\label{app:addexp}

\providecommand{\second}[1]{\cellcolor{gray!15}#1}
\providecommand{\std}[1]{{\scriptsize$\pm$#1}}

\subsection{Accuracy under Knowledge Transfer}
\label{app:transfer}

Table~\ref{tab:reuse_efficiency_final} measures whether transferred units are \emph{invoked}; here we measure whether they \emph{help}. We use the same deployments as Sec.~\ref{sec:reuse_efficiency} (protocol in Appendix~\ref{app:reuse_protocol}): the knowledge base accumulated by \texttt{GPT-5.4-mini} on RoboSpatial is frozen and deployed unchanged in-domain, cross-task on Omni3D-Bench and SAT, and cross-agent on RoboSpatial with \texttt{Qwen3.5-9B} and \texttt{Qwen3.6-27B} as executors. Two reference rows bound each column: ReAct uses no memory, and the bottom row is \textsc{SpatialCraft} accumulated on the target setting itself (Table~\ref{tab:main}). The \emph{w/o Experiences} and \emph{w/o Skills} rows disable one stream at deployment, so their in-domain values coincide with Table~\ref{tab:ablation}. Over the four transfer settings, Rec.\ is the share of the target-accumulated gain that a method recovers:
\begin{equation}
\mathrm{Rec.}=\frac{\overline{\mathrm{Acc}}_{m}-\overline{\mathrm{Acc}}_{\text{ReAct}}}{\overline{\mathrm{Acc}}_{\text{target}}-\overline{\mathrm{Acc}}_{\text{ReAct}}},
\end{equation}
where $\overline{\mathrm{Acc}}$ averages accuracy over the four settings. Averaging before taking the ratio avoids the unstable per-setting denominator on \texttt{Qwen3.5-9B}, whose target-accumulated gain is only 2.4 points.

\begin{table*}[t]
    \centering
    \caption{Accuracy (\%) after transferring the knowledge base accumulated by \texttt{GPT-5.4-mini} on RoboSpatial, without modification. Cross-task columns keep \texttt{GPT-5.4-mini} as the executor; cross-agent columns evaluate RoboSpatial with the listed executor. Avg.\ and Rec.\ cover the four transfer settings; Rec.\ is the share of the target-accumulated gain over ReAct that is recovered. Gray rows are references. Mean$\pm$std over three runs. Bold: best; shaded: second best among memory-based methods.}
    \label{tab:transfer_acc}
    \resizebox{\linewidth}{!}{%
    \begin{tabular}{l c cc cc cc}
        \toprule
        & \textbf{In-domain} & \multicolumn{2}{c}{\textbf{Cross-task}} & \multicolumn{2}{c}{\textbf{Cross-agent}} & \multicolumn{2}{c}{\textbf{Transfer summary}} \\
        \cmidrule(lr){2-2}\cmidrule(lr){3-4}\cmidrule(lr){5-6}\cmidrule(lr){7-8}
        \rowcolor{gray!12}
        \textbf{Method} & \textbf{RoboSpatial} & \textbf{Omni3D} & \textbf{SAT} & \textbf{Qwen3.5-9B} & \textbf{Qwen3.6-27B} & \textbf{Avg.} & \textbf{Rec. (\%)} \\
        \midrule
        \textcolor{gray}{ReAct (no memory)}
            & \textcolor{gray}{55.2\std{1.2}} & \textcolor{gray}{29.5\std{1.0}} & \textcolor{gray}{60.2\std{0.8}}
            & \textcolor{gray}{57.0\std{2.2}} & \textcolor{gray}{62.9\std{1.5}} & \textcolor{gray}{52.4} & \textcolor{gray}{0.0} \\
        \midrule
        RAG
            & 61.5\std{2.0} & 29.0\std{1.3} & 70.4\std{1.6}
            & 54.8\std{1.9} & 60.1\std{1.7} & 53.6 $\uparrow$1.2 & 8.3 \\
        MemP
            & 63.8\std{2.2} & 30.6\std{1.4} & 74.6\std{1.3}
            & 55.9\std{1.6} & 64.2\std{1.5} & 56.3 $\uparrow$3.9 & 27.7 \\
        SMA
            & \second{67.4\std{2.1}} & \second{32.8\std{1.1}} & \second{79.2\std{1.0}}
            & \second{57.6\std{1.3}} & \second{66.0\std{1.2}} & \second{58.9 $\uparrow$6.5} & \second{45.9} \\
        \textsc{SpatialCraft}
            & \textbf{70.9}\std{0.6} & \textbf{39.1}\std{1.2} & \textbf{85.2}\std{0.8}
            & \textbf{60.3}\std{0.9} & \textbf{69.4}\std{1.0} & \textbf{63.5} $\uparrow$11.1 & \textbf{78.4} \\
        \quad w/o Experiences
            & 61.3\std{0.9} & 33.6\std{1.0} & 70.7\std{1.1}
            & 57.6\std{1.4} & 65.6\std{1.2} & 56.9 $\uparrow$4.5 & 31.6 \\
        \quad w/o Skills
            & 58.9\std{0.6} & 31.9\std{1.3} & 67.2\std{1.2}
            & 58.4\std{1.1} & 64.5\std{1.3} & 55.5 $\uparrow$3.1 & 21.9 \\
        \midrule
        \textcolor{gray}{\textsc{SpatialCraft} (target-accumulated)}
            & \textcolor{gray}{70.9\std{0.6}} & \textcolor{gray}{46.0\std{1.4}} & \textcolor{gray}{90.3\std{0.7}}
            & \textcolor{gray}{59.4\std{0.6}} & \textcolor{gray}{70.5\std{1.8}} & \textcolor{gray}{66.6} & \textcolor{gray}{100.0} \\
        \bottomrule
    \end{tabular}%
    }
\end{table*}

\begin{figure}[t]
    \centering
    \includegraphics[width=\linewidth]{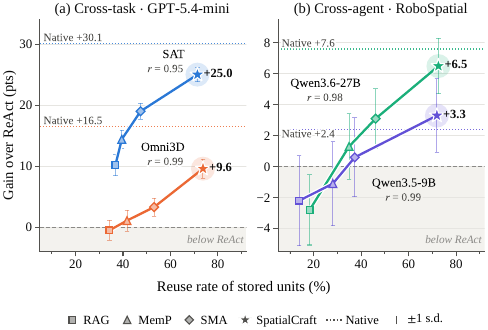}
    \caption{Accuracy gain over ReAct versus reuse rate (Table~\ref{tab:reuse_efficiency_final}) under (a) cross-task and (b) cross-agent transfer. Each line connects the four memory-based methods within one setting. The reuse rate of \textsc{SpatialCraft} pools its 12 Skills and 45 Experiences.}
    \label{fig:transfer_reuse_gain}
\end{figure}

\paragraph{\textsc{SpatialCraft} retains most of its gain after transfer.}
Across the four transfer settings, \textsc{SpatialCraft} averages 63.5\%, 11.1 points above ReAct, and recovers 78.4\% of the gain obtained by accumulating on each target directly. SMA, MemP, and RAG recover 45.9\%, 27.7\%, and 8.3\%. Retention is highest on SAT (25.0 of 30.1 points) and lowest on Omni3D-Bench (9.6 of 16.5), whose metric queries depend on scale estimation that RoboSpatial rarely requires.

\paragraph{Raw trajectories do not survive a shift in task or executor.}
RAG falls below ReAct on Omni3D-Bench and with both Qwen executors (by 0.5, 2.2, and 2.8 points), and MemP falls below it on \texttt{Qwen3.5-9B}. Retrieved trajectories carry the objects, coordinates, and tool outputs of their source tasks, which mislead once the task or the executor changes. These are also the methods with the lowest reuse rates in Table~\ref{tab:reuse_efficiency_final}.

\paragraph{Reuse translates into accuracy.}
Fig.~\ref{fig:transfer_reuse_gain} plots each method's accuracy gain against its reuse rate. Within every setting, the gain increases monotonically with reuse, and \textsc{SpatialCraft} lies at the upper right in all four. The higher reuse of \textsc{SpatialCraft} in Table~\ref{tab:reuse_efficiency_final} therefore reflects knowledge that helps once invoked, not merely knowledge that is invoked often.

\paragraph{A weaker executor benefits from knowledge distilled by a stronger one.}
On \texttt{Qwen3.5-9B}, the transferred knowledge base reaches 60.3\%, above the 59.4\% obtained when \texttt{Qwen3.5-9B} accumulates its own; on \texttt{Qwen3.6-27B}, it stays below (69.4\% vs.\ 70.5\%). This agrees with Fig.~\ref{fig:training}(c), where \texttt{Qwen3.5-9B} gains only 1.7 points from its own accumulation: it can follow knowledge that it cannot reliably distill. XSkill~\cite{jiang2026xskillcontinuallearningexperience} reports that transferred knowledge can hurt weaker executors; here the gain stays positive, plausibly because the stored units are compact and anchored Experiences enter the context only when their Skill activates.

\paragraph{The two streams remain complementary under transfer.}
Disabling either stream at deployment recovers only 31.6\% (w/o Experiences) or 21.9\% (w/o Skills) of the target-accumulated gain, against 78.4\% with both, so the combined gain exceeds the sum of the individual ones, as in-domain (Table~\ref{tab:ablation}). Which stream matters more depends on the executor and task: Skills contribute more cross-task and on \texttt{Qwen3.6-27B}, whereas Experiences contribute more on \texttt{Qwen3.5-9B} (58.4\% vs.\ 57.6\%), matching the reuse pattern in Table~\ref{tab:reuse_efficiency_final}.

\providecommand{\std}[1]{{\scriptsize$\pm$#1}}
\providecommand{\dn}[1]{{\scriptsize\textcolor{gray}{$\downarrow$#1}}}
\providecommand{\up}[1]{{\scriptsize\textcolor{gray}{$\uparrow$#1}}}

\subsection{Diagnosing the Contrastive Utility Score}
\label{app:cus_diag}

Sec.~\ref{sec:cus} argues that valuing an Experience by the rewards of the episodes in which it appears is confounded by task difficulty, and that randomized exposure removes this confound. We test this claim directly on RoboSpatial with \texttt{GPT-5.4-mini}, using three diagnostics: (i) whether each score tracks task difficulty, (ii) how the Experience lifecycle evolves during accumulation, and (iii) whether the score identifies the Experiences that carry the accuracy gain at deployment. Diagnostics (i) and (ii) are computed from the accumulation logs and require no additional rollouts; (iii) changes only the deployed knowledge base.

\paragraph{Setup.}
For (i), we consider every Experience that completed probation ($n_e \ge n_{\min}$). Each receives two scores computed from the same accumulation rollouts. The \emph{naive score} is the mean reward of the eligible rollouts in which $e$ was shown, which corresponds to the reward-based valuation of SMA~\cite{zhang2026spatial} and MemRL~\cite{zhang2026memrlselfevolvingagentsruntime} without retrieval-time re-ranking. For the Table~\ref{tab:ablation} ablation, we replace $U(e)$ with the naive score minus the global mean reward during accumulation, keeping randomized exposure and all lifecycle thresholds unchanged. The second is $U(e)$, taken when the state of $e$ was last decided. As a difficulty proxy that shares no rollouts with either score, we run the memory-free ReAct agent three times on every task of $\mathcal{D}_{\text{env}}$ and average its success over the tasks on which $e$ was eligible. Using the withheld rollouts instead would correlate spuriously with $U(e)$, since $\delta_i(e)$ subtracts their outcomes. For (iii), we rank the Experiences of the deployed bank that have completed probation by $U(e)$ or by the naive score, and remove or add 20\% of the bank size at deployment, leaving Skills unchanged. The added Experiences are drawn uniformly from the retired ones and keep their original anchors, or become global if their anchor Skill no longer exists. Random removal is averaged over three draws.

\begin{figure*}[t]
    \centering
    \includegraphics[width=\textwidth]{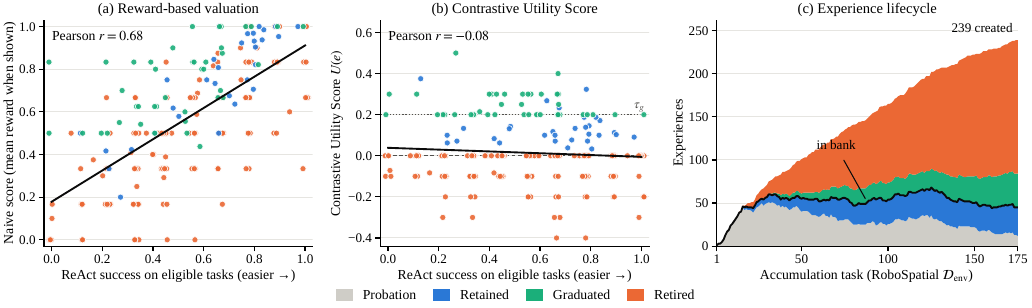}
    \caption{Diagnostics of the Contrastive Utility Score on RoboSpatial with \texttt{GPT-5.4-mini}. (a, b) Each point is an Experience that completed probation, plotted against the success rate of the memory-free ReAct agent on the tasks where it was eligible, and colored by its final state; black lines are least-squares fits. The naive score follows task difficulty, whereas $U(e)$ does not. Retained points above $\tau_g$ in (b) are global Experiences, which cannot graduate. (c) Cumulative fate of all created Experiences over accumulation; the black line is the bank size (probation and retained, including graduated Experiences awaiting absorption).}
    \label{fig:cus_diagnostics}
\end{figure*}

\begin{table}[t]
    \centering
    \caption{Deployment-time removal and addition of Experiences with \texttt{GPT-5.4-mini}. Each variant removes or adds 20\% of the deployed bank size; Skills are unchanged. Arrows give the change from the full knowledge base. Mean$\pm$std over three runs.}
    \label{tab:cus_removal}
    \resizebox{\linewidth}{!}{%
    \begin{tabular}{l ll}
        \toprule
        \rowcolor{gray!12}
        \textbf{Deployed Experience bank} & \textbf{RoboSpatial} & \textbf{Omni3D} \\
        \midrule
        Full knowledge base                 & 70.9\std{0.6}            & 46.0\std{1.4} \\
        \midrule
        $-$ Top 20\% by $U(e)$              & 66.3\std{0.9} \dn{4.6}   & 43.1\std{1.2} \dn{2.9} \\
        $-$ Top 20\% by naive score         & 68.6\std{1.0} \dn{2.3}   & 44.7\std{1.3} \dn{1.3} \\
        $-$ Random 20\%                     & 68.9\std{0.8} \dn{2.0}   & 44.9\std{1.1} \dn{1.1} \\
        $-$ Bottom 20\% by $U(e)$           & 71.2\std{0.7} \up{0.3}   & 46.3\std{1.2} \up{0.3} \\
        $+$ 20\% retired Experiences        & 69.3\std{0.9} \dn{1.6}   & 45.2\std{1.3} \dn{0.8} \\
        \midrule
        \textcolor{gray}{w/o Experiences (Table~\ref{tab:ablation})} & \textcolor{gray}{61.3\std{0.9}} \dn{9.6} & \textcolor{gray}{40.5\std{0.6}} \dn{5.5} \\
        \bottomrule
    \end{tabular}%
    }
\end{table}
\vspace{-1em}
\paragraph{Reward-based valuation tracks task difficulty; the Contrastive Utility Score does not.}
Fig.~\ref{fig:cus_diagnostics}(a) shows that the naive score rises with the ReAct success rate on the tasks where an Experience was eligible: an Experience retrieved mostly for easy tasks scores high regardless of its own effect. $U(e)$ is nearly uncorrelated with the same proxy (Fig.~\ref{fig:cus_diagnostics}b), because each contrast compares rollouts of the same task. The two scores rank Experiences differently (Spearman $\rho=0.46$). Of the 155 Experiences retired by CUS, 59 (38\%) score above the median naive value, so a reward-based valuation would have kept them.

\paragraph{Most distilled Experiences are retired, and the bank stays compact.}
Of the 239 Experiences created during accumulation, 155 (65\%) are retired, 41 graduate (39 of them already absorbed into their anchor Skills), 31 are retained, and 12 remain in probation (Fig.~\ref{fig:cus_diagnostics}c). The bank peaks at 68 entries and ends at 45 while creation continues, so retirement and consolidation, rather than the capacity limit, keep it compact. The high retirement rate is consistent with our motivation of CUS: many lessons that read as plausible do not improve outcomes once their exposure is randomized.

\paragraph{The score identifies the Experiences that carry the gain.}
Table~\ref{tab:cus_removal} shows that removing the top 20\% of Experiences by $U(e)$ costs 4.6 points on RoboSpatial, nearly half of the 9.6 points lost by removing all of them, and more than twice the cost of removing a random 20\%. Removing the top 20\% by naive score costs about as much as random removal, so the naive ranking carries little information about which Experiences matter. Removing the bottom 20\% by $U(e)$ leaves accuracy unchanged, and restoring retired Experiences lowers it, confirming that retirement discards entries that are useless or harmful. Omni3D-Bench shows the same ordering.

\begin{figure*}[t]
    \centering
    \includegraphics[width=\linewidth]{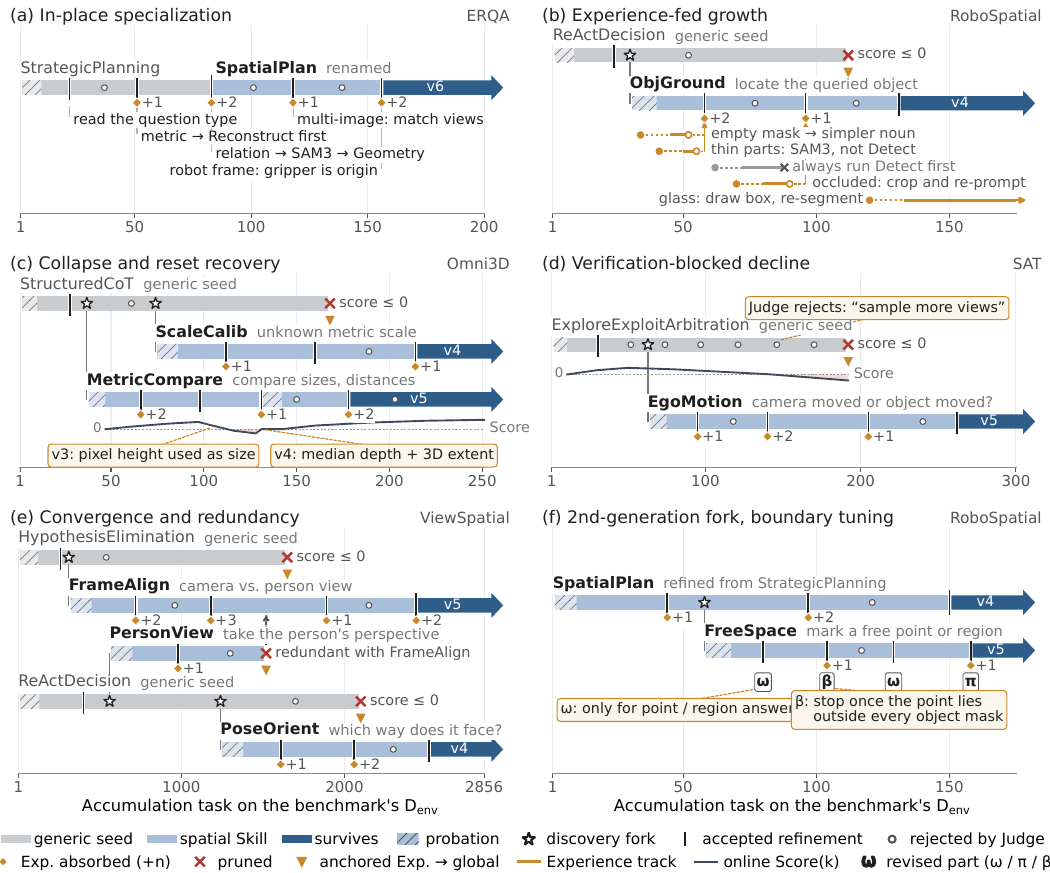}
    \vspace{-1em}
    \caption{Six recurring modes of Skill evolution, each taken from accumulation with \texttt{GPT-5.4-mini} on a different benchmark (top right of each panel); the x-axis counts accumulation tasks on that benchmark's $\mathcal{D}_{\text{env}}$. (a) In-place specialization: \texttt{StrategicPlanning} acquires spatial steps through successive refinements and is renamed \texttt{SpatialPlan}; labels give what each refinement adds. (b) Experience-fed growth: \texttt{ObjGround}, forked from \texttt{ReActDecision}, absorbs graduated Experiences from its anchored track, while the retired ``always run Detect first'' is never absorbed. (c) Collapse and reset recovery: the online score of \texttt{MetricCompare} falls below zero after v3 uses pixel height as size; v4 replaces it with median depth and 3D extent, restarts probation, and recovers. (d) Verification-blocked decline: the Judge rejects most revisions of \texttt{ExploreExploitArbitration}, whose score declines until it is pruned, while its fork \texttt{EgoMotion} survives. (e) Convergence and redundancy: \texttt{PersonView}, forked from \texttt{ReActDecision}, duplicates \texttt{FrameAlign} and is pruned as redundant; \texttt{PoseOrient} survives. (f) Second-generation fork with boundary tuning: \texttt{FreeSpace} forks from \texttt{SpatialPlan}, itself refined from a seed, and three of its four refinements revise the activation ($\omega$) or termination ($\beta$) condition rather than the procedure ($\pi$).}
    \label{fig:lineage_app}
    \vspace{-1em}
\end{figure*}


\subsection{Skill Evolution Diagnostics}
\label{app:skill_diag}

Sec.~\ref{sec:evolve} tracks returns under ablations of Skill evolution, and Fig.~\ref{fig:skillevolve} traces the lineage of the RoboSpatial Skill Pool. Here we examine the evolution process itself: how often revisions are accepted and whether accepted revisions improve full trajectories (Table~\ref{tab:skill_stats}), how each evolution mechanism changes these statistics (Table~\ref{tab:skill_variants}), and which lineage patterns recur across benchmarks (Fig.~\ref{fig:lineage_app}). All statistics come from the accumulation logs of \texttt{GPT-5.4-mini}.

\paragraph{Definitions.}
A \emph{revision attempt} revises one Skill in one round with $N_c$ candidates, and is \emph{accepted} if its best candidate has $J(k')>0$. A \emph{Judge tie} is a replayed state scored $0$, including states at which the parent and candidate actions are identical. Because $J$ compares single actions, it cannot show whether a revision improves full trajectories. We therefore also report the \emph{post-acceptance contrast}: the fraction of accepted revisions whose within-task contrast $\sum\delta_i(k)$, summed over the first $n_{\min}$ contrasts after acceptance, is positive. Unlike $\mathrm{Score}(k)$, this quantity excludes the statistics inherited from the parent.

\begin{table}[t]
    \centering
    \caption{Skill evolution statistics with \texttt{GPT-5.4-mini}. Accepted revisions are split into refinements (Ref.) and discoveries (Disc.). Post-acc.\ $\delta>0$: fraction of accepted revisions whose contrast over the next $n_{\min}$ tasks is positive. Created Skills include the six seeds.}
    \label{tab:skill_stats}
    \resizebox{\linewidth}{!}{%
    \begin{tabular}{l rrrrr}
        \toprule
        \rowcolor{gray!12}
        & \textbf{Robo.} & \textbf{ERQA} & \textbf{Omni.} & \textbf{SAT} & \textbf{View.} \\
        \midrule
        $|\mathcal{D}_{\text{env}}|$         & 175   & 200   & 251   & 300   & 2,856 \\
        Revision attempts                    & 188   & 221   & 274   & 318   & 3,042 \\
        Acceptance rate (\%)                 & 48.9  & 46.6  & 51.5  & 41.2  & 44.8 \\
        Accepted (Ref.\,/\,Disc.)            & 78\,/\,14 & 89\,/\,14 & 124\,/\,17 & 117\,/\,14 & 1,321\,/\,42 \\
        Judge ties (\%)                      & 34.7  & 38.2  & 29.6  & 41.5  & 36.9 \\
        Post-acc.\ $\delta>0$ (\%)           & 81.4  & 77.9  & 79.1  & 72.3  & 80.6 \\
        Experiences absorbed                 & 39    & 41    & 63    & 48    & 312 \\
        \midrule
        Skills created                       & 20    & 20    & 23    & 20    & 48 \\
        Pruned (seed\,/\,spatial)            & 4\,/\,4 & 4\,/\,5 & 4\,/\,6 & 5\,/\,5 & 5\,/\,27 \\
        Final pool                           & 12    & 11    & 13    & 10    & 16 \\
        \bottomrule
    \end{tabular}%
    }
\end{table}

\begin{table}[t]
    \centering
    \caption{Internal statistics of the Skill-evolution variants of Fig.~\ref{fig:training}(a) on RoboSpatial with \texttt{GPT-5.4-mini}, at the end of accumulation. Final Score: mean $\mathrm{Score}(k)$ over the final pool. Spatial: discovered Skills in the final pool of 12.}
    \label{tab:skill_variants}
    \resizebox{\linewidth}{!}{%
    \begin{tabular}{l cccc}
        \toprule
        \rowcolor{gray!12}
        \textbf{Variant} & \textbf{Accept. (\%)} & \textbf{Post-acc. $\delta>0$ (\%)} & \textbf{Final Score} & \textbf{Spatial} \\
        \midrule
        \textsc{SpatialCraft} (full)    & 48.9  & \textbf{81.4} & \textbf{0.12} & \textbf{10} \\
        w/o Semantic Gradients          & 36.2  & 68.5 & 0.06 & 7 \\
        w/o Candidate Verification      & 100.0 & 51.7 & 0.02 & 9 \\
        w/o Online Score (FIFO)         & 47.3  & 79.8 & $-$0.01 & 6 \\
        \bottomrule
    \end{tabular}%
    }
\end{table}

\paragraph{The Judge is selective, and its local verdicts carry over to full trajectories.}
Across benchmarks, 41--52\% of revision attempts are accepted (Table~\ref{tab:skill_stats}), and 72--81\% of accepted revisions have a positive contrast over the following tasks. When a randomly chosen candidate is adopted instead, this fraction drops to 51.7\% (Table~\ref{tab:skill_variants}), close to what an unhelpful revision would produce by chance. Wins at replayed decision states therefore predict improvement on full trajectories, which the Judge cannot observe directly. SAT has the lowest acceptance and the most ties: its two-frame questions leave few alternative actions at a single replayed state, so parent and candidate often choose the same action. Fig.~\ref{fig:lineage_app}(d) shows the extreme case.

\paragraph{Refinement dominates, but discovery builds the spatial pool.}
Discoveries account for only 3--15\% of accepted revisions, yet discovered Skills make up most of each final pool, for example 10 of 12 on RoboSpatial (Table~\ref{tab:skill_variants}), while four or five of the six seeds are pruned on every benchmark. Refinements also carry consolidation: on RoboSpatial, they fold 39 graduated Experiences into their anchor Skills, the same Experiences counted as absorbed in Appendix~\ref{app:cus_diag}.

\paragraph{Each mechanism protects a different property of the pool.}
Table~\ref{tab:skill_variants} explains the curves of Fig.~\ref{fig:training}(a). Without semantic gradients, candidates written from neutral summaries are accepted less often (36.2\%) and help less once accepted, and fewer spatial Skills emerge. Without candidate verification, every revision is adopted, half of them fail to help, and the mean final Score falls to 0.02. FIFO eviction leaves acceptance and post-acceptance contrast nearly unchanged but ends with the lowest final Score and only six spatial Skills, because it removes Skills by age, including mature ones that have recovered from earlier setbacks. Revision quality and retention thus fail independently and require separate mechanisms.

\begin{table*}[t]
    \centering
    \caption{Composition of the final knowledge bases. Experiences in the final bank are split into anchored (anc.) and global (glob.). Created counts Experiences committed after merging; every created Experience is absorbed into a Skill, retired, or still in the bank, so Created $=$ Absorbed $+$ Retired $+$ Experiences. Grad.\ rate: Absorbed / Created. Stored tokens list the total with the per-unit average in parentheses. The RoboSpatial / \texttt{GPT-5.4-mini} row appears in both blocks and matches Table~\ref{tab:reuse_efficiency_final}.}
    \label{tab:kb_stats}
    \resizebox{\linewidth}{!}{%
    \begin{tabular}{l cc ccc c ccc}
        \toprule
        & \multicolumn{2}{c}{\textbf{Final knowledge base}} & \multicolumn{3}{c}{\textbf{Experience lifecycle}} & & \multicolumn{3}{c}{\textbf{Stored tokens}} \\
        \cmidrule(lr){2-3}\cmidrule(lr){4-6}\cmidrule(lr){8-10}
        \rowcolor{gray!12}
        \textbf{Setting} & \textbf{Skills} & \textbf{Experiences (anc./glob.)} & \textbf{Created} & \textbf{Absorbed} & \textbf{Retired} & \textbf{Grad. rate (\%)} & \textbf{Skills} & \textbf{Experiences} & \textbf{Total} \\
        \midrule
        \multicolumn{10}{l}{\textit{\texttt{GPT-5.4-mini} across benchmarks}} \\
        RoboSpatial    & 12 & 45 (24\,/\,21) & 239   & 39  & 155   & 16.3 & 2,580 (215) & 3,510 (78) & 6,090 \\
        ERQA           & 11 & 52 (27\,/\,25) & 268   & 41  & 175   & 15.3 & 2,178 (198) & 3,848 (74) & 6,026 \\
        Omni3D-Bench   & 13 & 58 (33\,/\,25) & 331   & 63  & 210   & 19.0 & 3,146 (242) & 4,814 (83) & 7,960 \\
        SAT            & 10 & 49 (24\,/\,25) & 287   & 48  & 190   & 16.7 & 1,870 (187) & 3,479 (71) & 5,349 \\
        ViewSpatial    & 16 & 96 (51\,/\,45) & 2,214 & 312 & 1,806 & 14.1 & 3,696 (231) & 7,680 (80) & 11,376 \\
        \midrule
        \multicolumn{10}{l}{\textit{RoboSpatial across backbones}} \\
        \texttt{GPT-5.4}      & 13 & 41 (20\,/\,21) & 221 & 44 & 136 & 19.9 & 3,068 (236) & 3,444 (84) & 6,512 \\
        \texttt{GPT-5.4-mini} & 12 & 45 (24\,/\,21) & 239 & 39 & 155 & 16.3 & 2,580 (215) & 3,510 (78) & 6,090 \\
        \texttt{Qwen3.6-27B}  & 11 & 52 (25\,/\,27) & 263 & 31 & 180 & 11.8 & 2,244 (204) & 3,952 (76) & 6,196 \\
        \texttt{Qwen3.5-9B}   &  9 & 61 (26\,/\,35) & 302 & 18 & 223 &  6.0 & 1,584 (176) & 4,209 (69) & 5,793 \\
        \bottomrule
    \end{tabular}%
    }
\end{table*}

\begin{table}[t]
    \centering
    \caption{Accumulation cost per $\mathcal{D}_{\text{env}}$ task on RoboSpatial with \texttt{GPT-5.4-mini}: LLM calls, input and output tokens (thousands), and share of total tokens. Baselines use the same $N{=}4$ rollouts; Rel.\ is the cost relative to each baseline.}
    \label{tab:accum_cost}
    \resizebox{\linewidth}{!}{%
    \begin{tabular}{l rrrr}
        \toprule
        \rowcolor{gray!12}
        \textbf{Component} & \textbf{Calls} & \textbf{Input (k)} & \textbf{Output (k)} & \textbf{Share (\%)} \\
        \midrule
        \multicolumn{5}{l}{\textit{Rollouts}} \\
        \quad Execution ($N{=}4$)            & 31.2 & 268.3 & 11.9 & 46.4 \\
        \quad Decomposition \& rewrite       &  2.0 &   3.7 &  0.6 &  0.7 \\
        \quad Termination checks             &  8.4 &  23.5 &  0.4 &  4.0 \\
        \multicolumn{5}{l}{\textit{Experience accumulation}} \\
        \quad Rollout summary                &  1.0 &  46.0 &  2.6 &  8.1 \\
        \quad Cross-rollout critique         &  1.0 &  11.0 &  1.4 &  2.1 \\
        \quad Experience merge               &  0.7 &   1.3 &  0.2 &  0.2 \\
        \quad Contrastive Utility Score      &  0.0 &   0.0 &  0.0 &  0.0 \\
        \multicolumn{5}{l}{\textit{Skill evolution}} \\
        \quad Semantic gradients             &  6.4 &  39.7 &  4.5 &  7.3 \\
        \quad Aggregation \& candidates      &  4.3 &  24.1 &  6.0 &  5.0 \\
        \quad Key-state replay               & 19.3 & 108.1 &  5.8 & 18.9 \\
        \quad Judge                          & 12.6 &  42.8 &  1.9 &  7.4 \\
        \midrule
        \textbf{\textsc{SpatialCraft} total} & \textbf{86.9} & \textbf{568.5} & \textbf{35.2} & \textbf{100.0} \\
        \midrule
        \rowcolor{gray!12}
        \textbf{Baseline} & \textbf{Calls} & \textbf{Input (k)} & \textbf{Output (k)} & \textbf{Rel.} \\
        \midrule
        RAG       & 31.2 & 257.6 & 11.9 & 2.24$\times$ \\
        MemP      & 35.2 & 293.6 & 15.1 & 1.96$\times$ \\
        MemRL-GT  & 35.2 & 295.6 & 14.7 & 1.95$\times$ \\
        SMA       & 35.2 & 299.6 & 15.5 & 1.92$\times$ \\
        \bottomrule
    \end{tabular}%
    }
\end{table}

\paragraph{Six recurring lineage modes.}
Fig.~\ref{fig:lineage_app} shows six patterns that recur across benchmarks, each illustrating a distinct mechanism.
\begin{itemize}[leftmargin=*,itemsep=1pt,topsep=2pt]
\item \textbf{Specialization without discovery} (a). Refinement alone can turn a generic procedure into a spatial one. Each accepted revision of \texttt{StrategicPlanning} on ERQA adds one spatial step, such as routing metric questions to \texttt{Reconstruct} first or treating the gripper as the origin of the robot frame, until the Skill is renamed to reflect its new role.
\item \textbf{Consolidation filtered by utility} (b). \texttt{ObjGround} absorbs three of the Experiences anchored to it, written into the steps at which their triggers arise. The plausible-sounding ``always run Detect first'' is retired by CUS before it can graduate, so it never enters the procedure.
\item \textbf{Recovery through reset} (c). An accepted revision can still be harmful: v3 of \texttt{MetricCompare} wins at the replayed states yet lowers full-trajectory outcomes, and its online score turns negative. Because the next accepted revision inherits no statistics from a negative-scoring parent (Appendix~\ref{app:skill_details}), v4 restarts probation instead of being pruned, and its score recovers. Verification and online scoring thus correct each other's errors.
\item \textbf{Decline under verification} (d). On SAT, candidates for \texttt{ExploreExploitArbitration}, such as ``sample more views,'' do not change the action at replayed states, since the task provides at most two frames. The Judge rejects them, the seed cannot adapt, and its score drifts below zero until it is pruned. Its fork \texttt{EgoMotion}, which addresses the camera-versus-object motion question directly, is refined and survives.
\item \textbf{Convergent Skills} (e). Skills forked from different seeds can converge to the same function: \texttt{PersonView} and \texttt{FrameAlign} both resolve whose viewpoint a question refers to. Redundancy pruning removes the younger one and keeps the pool compact, while \texttt{PoseOrient}, which addresses object orientation, survives alongside \texttt{FrameAlign}.
\item \textbf{Boundary tuning of mature Skills} (f). Once a procedure stabilizes, revisions shift to its boundaries. \texttt{FreeSpace} is restricted to point and region answers ($\omega$) and told to stop once the point lies outside every object mask ($\beta$). This addresses premature or misplaced activation rather than the procedure itself.
\end{itemize}
In four of the six panels, the generic seed is pruned after spawning spatial Skills, and its anchored Experiences become global. Seeds therefore act as scaffolding: they supply the initial procedures from which spatial Skills are forked, and are removed once their descendants handle the states they covered.


\subsection{Knowledge Statistics and Accumulation Cost}
\label{app:kb_cost}

Table~\ref{tab:reuse_efficiency_final} reports the size and deployment cost of the knowledge base accumulated by \texttt{GPT-5.4-mini} on RoboSpatial. Here we report the composition of every knowledge base used in Table~\ref{tab:main} (Table~\ref{tab:kb_stats}) and the cost of building it (Table~\ref{tab:accum_cost}). Stored tokens are counted with the \texttt{cl100k\_base} tokenizer as in Appendix~\ref{app:reuse_protocol}. Accumulation tokens are those reported by the API, including image tokens, and cover every call made while processing $\mathcal{D}_{\text{env}}$.

\paragraph{The knowledge base stays small as accumulation grows.}
Every final knowledge base holds 9--16 Skills and 41--96 Experiences, totaling 5.3k--11.4k stored tokens (Table~\ref{tab:kb_stats}). Even the largest, accumulated over the 2,856 tasks of ViewSpatial, is smaller than the 22.5k and 39.2k tokens that MemP and SMA store for RoboSpatial alone (Table~\ref{tab:reuse_efficiency_final}). ViewSpatial has 16 times as many accumulation tasks as RoboSpatial but only a 1.9 times larger knowledge base, because retirement and consolidation remove Experiences about as fast as new ones are created once the common question types are covered.

\paragraph{Most Experiences leave the bank, mostly through retirement.}
Across benchmarks, 63--82\% of created Experiences are retired and 14--19\% are absorbed into Skills. The rest remain in the bank, about half anchored and half global. ViewSpatial retires the largest share (82\%): its long accumulation repeatedly produces near-duplicate lessons about the same viewpoint questions, which add no contrast once an equivalent entry exists. Omni3D-Bench has the highest graduation rate (19.0\%) and the longest Skills (242 tokens), consistent with its multi-step metric procedures, which offer more steps into which a tactic can be folded.

\paragraph{Weaker backbones keep more Experiences and graduate fewer.}
On RoboSpatial, the graduation rate falls from 19.9\% with \texttt{GPT-5.4} to 6.0\% with \texttt{Qwen3.5-9B}, while the bank grows from 41 to 61 Experiences and the pool shrinks from 13 to 9 Skills. A weaker executor follows a shown Experience less consistently, which shrinks its contrast, so fewer Experiences reach $\tau_g$ and more remain as separate prompts instead of being folded into Skills. This agrees with Fig.~\ref{fig:training}(c), where \texttt{Qwen3.5-9B} gains least from accumulation, and with Appendix~\ref{app:transfer}, where it benefits from a knowledge base distilled by a stronger backbone.

\paragraph{Half of the accumulation cost is rollouts that every baseline also pays.}
\textsc{SpatialCraft} spends 604k tokens per accumulation task, about 106M tokens over RoboSpatial's 175 tasks (Table~\ref{tab:accum_cost}). Half of this (51.1\%) comes from the $N{=}4$ rollouts, which all memory-based baselines also run. The Contrastive Utility Score adds no calls, since it reuses the outcomes of these rollouts. Experience accumulation adds two calls per task (10.3\% of tokens). Skill evolution accounts for the remaining 38.6\%, most of it from key-state replay and the Judge (26.3\% together). This is the price of verifying revisions without token likelihoods, so that the same procedure applies to proprietary backbones. Removing it (w/o Candidate Verification) would save about a quarter of the tokens but lets half of the accepted revisions fail to help (Table~\ref{tab:skill_variants}).

\paragraph{The extra cost is paid once, before deployment.}
Overall, accumulation costs 1.9--2.2$\times$ that of the baselines per task. The cost is incurred once per benchmark and backbone, and knowledge is frozen afterwards. At deployment, Skills and Experiences together add 555.9 prompt tokens per step (209.3 and 346.6), fewer than MemP (601.8) and SMA (632.6; Table~\ref{tab:reuse_efficiency_final}), so the per-step cost does not grow despite the richer knowledge. The same knowledge base can also be reused across tasks and executors without further accumulation (Appendix~\ref{app:transfer}).
\section{Case Study}
\label{app:case study}

\subsection{Knowledge Base Examples}
\label{app:case_kb}
\begin{figure*}[t]
    \centering
    \includegraphics[width=\textwidth]{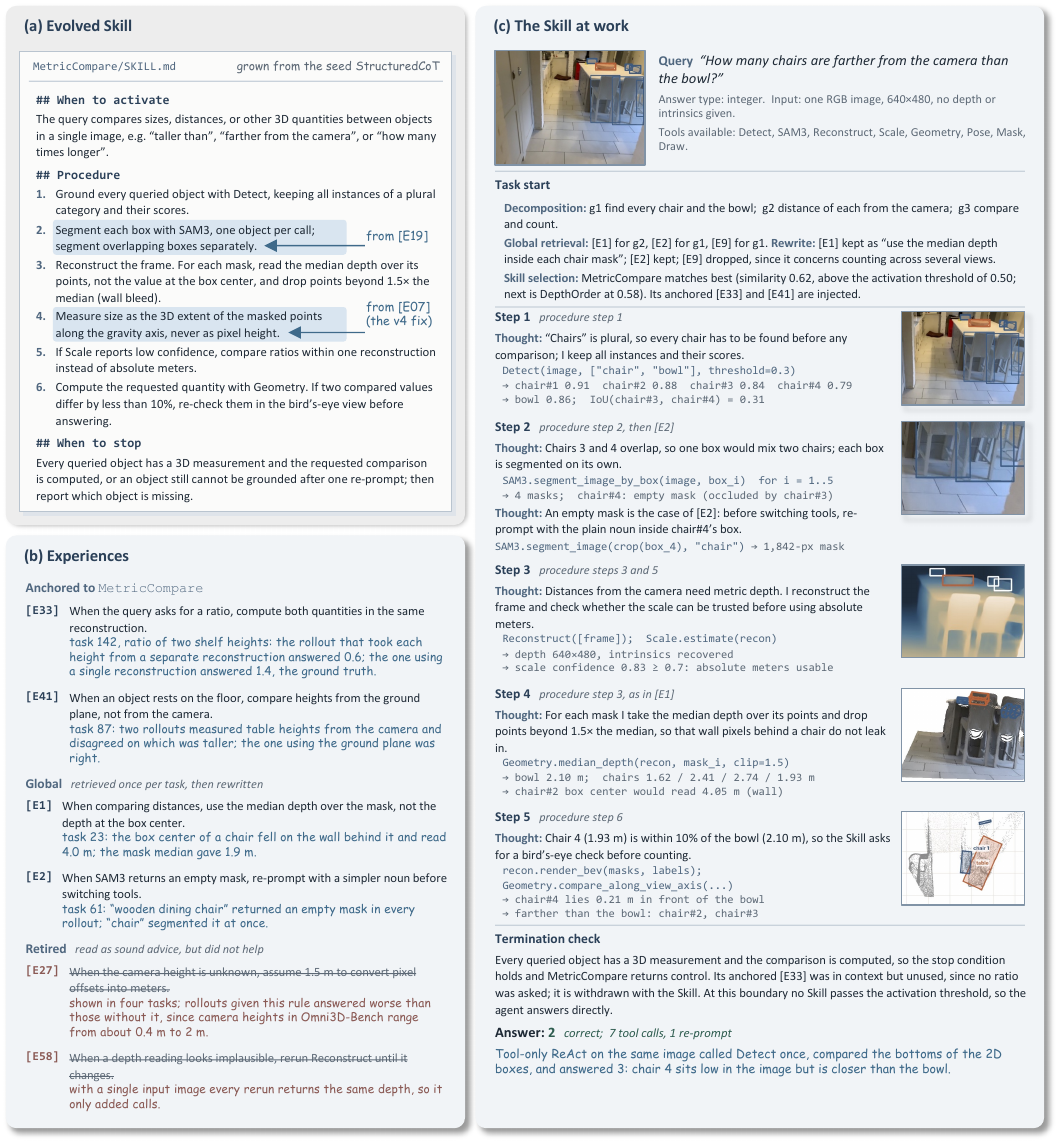}
    \caption{Knowledge base representative case example}
    \label{fig:case_kb}
\end{figure*}

Fig.~\ref{fig:case_kb} shows part of the knowledge base accumulated by \texttt{GPT-5.4-mini} on Omni3D-Bench (statistics in Table~\ref{tab:kb_stats}) and traces how it guides the agent on one deployment query.

\subsection{From Rollouts to Experiences}
\label{app:case_rollout}
\begin{figure*}[t]
    \centering
    \includegraphics[width=\textwidth]{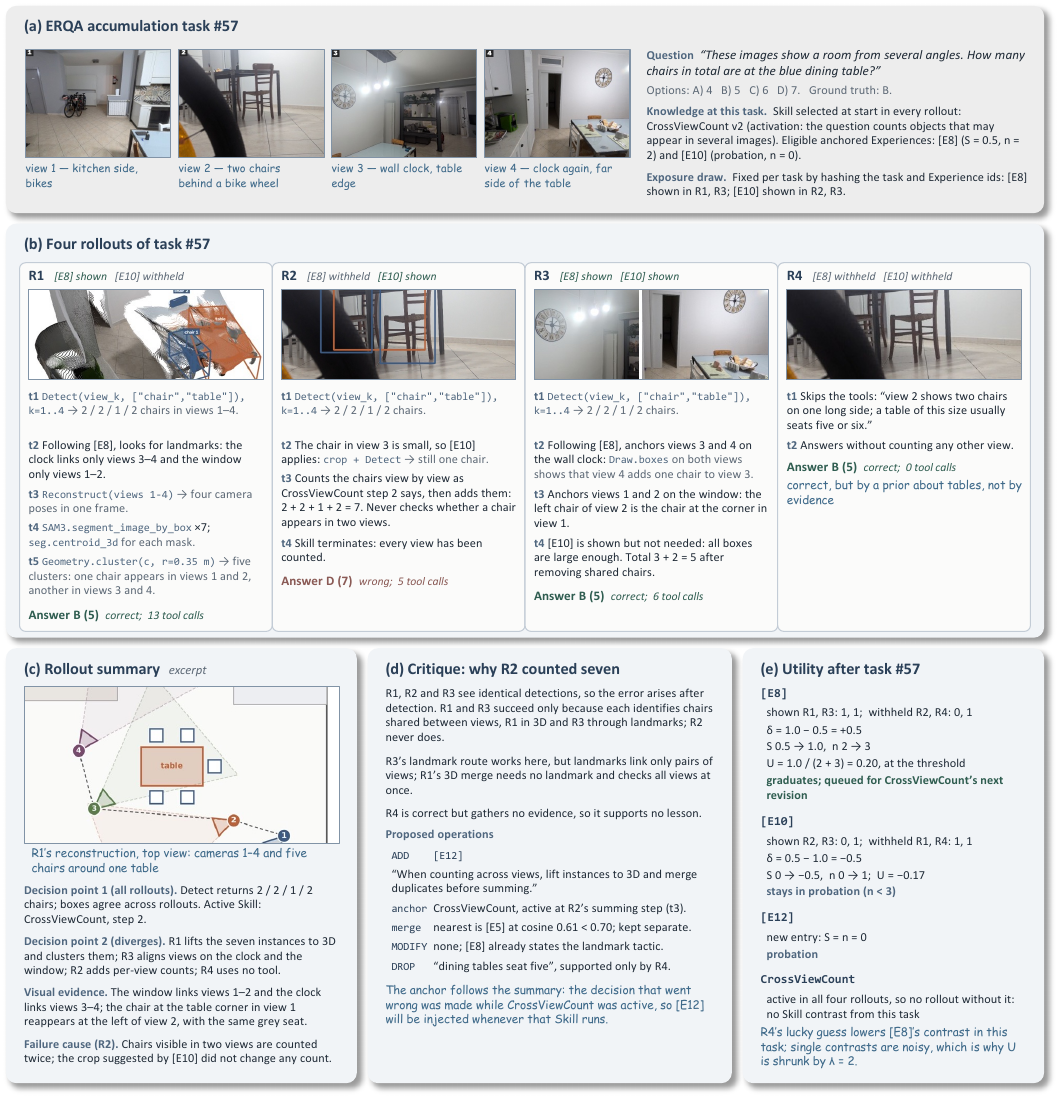}
    \caption{From rollouts of one accumulation task to a new Experience.}
    \label{fig:case_rollout}
\end{figure*}

As shown in Figure~\ref{fig:case_rollout}, the rollouts share the images and the Skill and differ only in which anchored Experiences they are shown. The summary locates where they diverge, and the critique turns the failure of R2 into \texttt{[E12]}, anchored to the Skill that was active when the wrong decision was made. The same rollouts also update the utility of the Experiences under test: \texttt{[E8]} reaches the graduation threshold, \texttt{[E10]} receives a negative contrast, and R4, correct without evidence, contributes to the contrasts but to no lesson.

\subsection{Semantic Gradients}
\label{app:case_gradient}
 
\begin{figure*}[t]
    \centering
    \includegraphics[width=\textwidth]{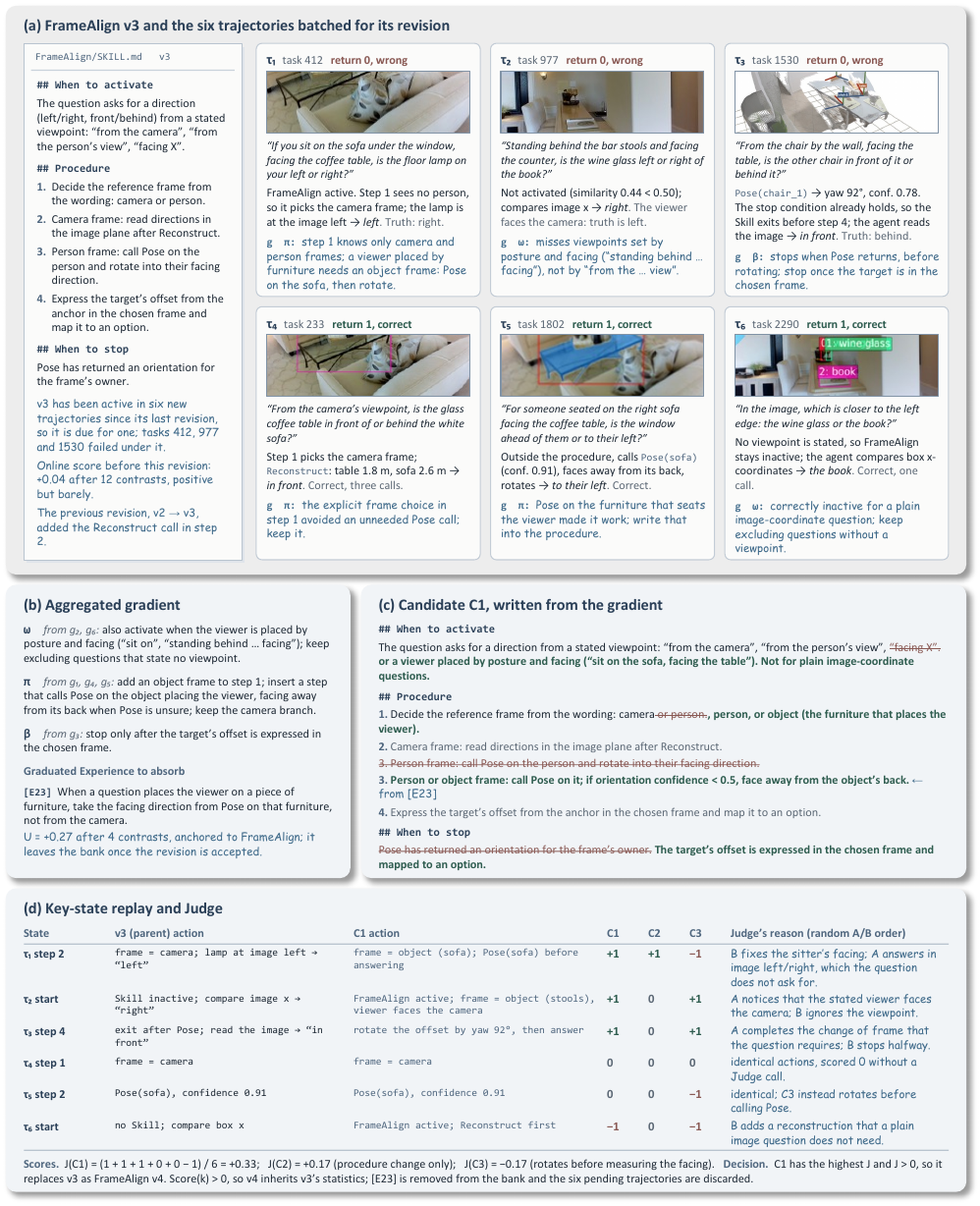}
    \caption{Semantic gradients example.}
    \label{fig:case_gradient}
\end{figure*}

Figure~\ref{fig:case_gradient} illustrates the case of semantic gradients generation. Each of the six batched trajectories yields a semantic gradient that names the faulty component: the failures expose a missing object frame ($\pi$), a too narrow activation ($\omega$), and a premature stop ($\beta$), while the successes mark what to keep. The aggregated gradient and the graduated \texttt{[E23]} are written into candidate C1, which wins the key-state replay against two alternatives and becomes v4. The Judge also penalizes the one state at which C1's wider activation adds an unnecessary call, so the revision is accepted on its net effect rather than on uniform agreement.

\subsection{Failure Cases}
\label{app:case_failure}
 
\begin{figure*}[t]
    \centering
    \includegraphics[width=\textwidth]{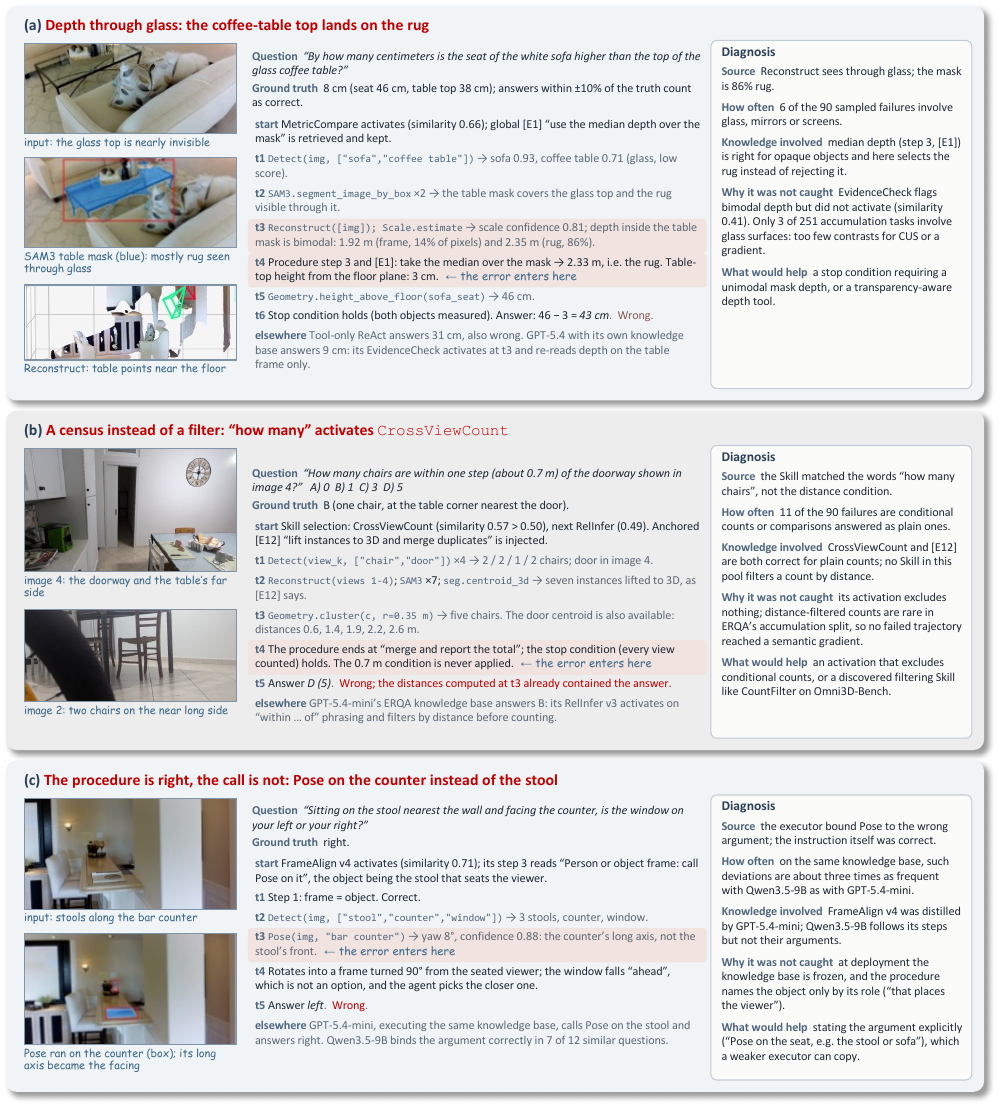}
    \caption{Representative failure cases.}
    \label{fig:case_failure}
\end{figure*}

Fig.~\ref{fig:case_failure} shows three representative deployment failures, one per source. In (a), a tool error on a transparent surface passes through a procedure step that is correct for opaque objects; in (b), a Skill activates on the wording of a question whose condition its procedure does not cover; in (c), a weaker executor follows a transferred procedure but binds its tool call to the wrong object. Each failure type is rare in its accumulation split, so it produces too few contrasts or failed trajectories for the Contrastive Utility Score and semantic gradients to correct it before the knowledge base is frozen.

\section{Full Related Work}
\label{app:fullrelatedwork}

\paragraph{Spatial reasoning benchmarks.}
A growing set of benchmarks probes the spatial abilities of MLLMs. BLINK~\cite{fu2024blinkmultimodallargelanguage} shows that models that describe images fluently still fail at perceptual tasks such as relative depth and correspondence, and 3DSRBench~\cite{ma20253dsrbenchcomprehensive3dspatial}, OmniSpatial~\cite{jia2026omnispatialcomprehensivespatialreasoning}, SITE~\cite{wang2025sitespatialintelligencethorough}, and VSI-Bench~\cite{yang2025thinkingspacemultimodallarge} extend evaluation to 3D, multi-view, and video settings. Other benchmarks isolate specific capabilities: perspective taking in ViewSpatial~\cite{li2025viewspatialbenchevaluatingmultiperspectivespatial}, robot-centric relations in RoboSpatial~\cite{song2026robospatialteachingspatialunderstanding} and EmbSpatial-Bench~\cite{du-etal-2024-embspatial}, embodied reasoning in ERQA~\cite{geminirobotics2025bringingphysicalworld}, dynamic and egocentric reasoning in SAT~\cite{ray2025satdynamicspatialaptitude}, and metric 3D queries in Omni3D-Bench~\cite{marsili2025visualagenticai}. Earlier work on 3D and embodied question answering~\cite{das2017embodiedquestionanswering,azuma2022scanqa3dquestionanswering,ma2023sqa3dsituatedquestionanswering,DBLP:conf/cvpr/MajumdarA0PYHSM24} and embodied agent benchmarks~\cite{yang2025embodiedbenchcomprehensivebenchmarkingmultimodal} assesses spatial understanding inside reconstructed or simulated scenes. We evaluate on RoboSpatial, ERQA, Omni3D-Bench, SAT, and ViewSpatial, which together cover robot-centric, embodied, metric, dynamic, and perspective reasoning.

\paragraph{Training MLLMs for spatial reasoning.}
One line of work equips models with 3D inputs, encoding point clouds or scene features directly~\cite{hong20233dllminjecting3dworld,xu2024pointllmempoweringlargelanguage,wang2023chat3ddataefficientlytuninglarge,zhu20233dvistapretrainedtransformer3d,huang2024embodiedgeneralistagent3d}. Another post-trains 2D MLLMs on synthesized spatial question-answer data~\cite{chen2024spatialvlmendowingvisionlanguagemodels,cheng2024spatialrgptgroundedspatialreasoning,cai2025spatialbotprecisespatialunderstanding,daxberger2025mmspatialexploring3dspatial,ray2025satdynamicspatialaptitude}. Recent methods close the loop between data and model, deriving training signals from deterministic geometric environments~\cite{li2026spatialevoselfevolvingspatialintelligence}, from the consistency of geometric logic~\cite{liu2026selfevolvingspatialreasoningvision}, or from an iterated data-model pipeline~\cite{zhao2026ouroborosspatialclosingdatamodelloop}, and whether explicit object localization is needed for spatial reasoning at all is itself under study~\cite{liu2026groundingisntknowingvlms}. These approaches improve the backbone, so they must be repeated whenever the backbone or toolset changes. \textsc{SpatialCraft} keeps the backbone frozen and instead improves how it uses existing tools.

\paragraph{Tool-augmented spatial agents.}
Vision foundation models now provide reliable geometric primitives: monocular and multi-view geometry~\cite{yang2024depthv2,bochkovskii2025depthprosharpmonocular,hu2025metric3dv2versatilemonoculargeometric,wang2026moge,lin2025depth}, open-vocabulary detection and segmentation~\cite{liu2024grounding,carion2026sam}, and object orientation~\cite{wang2024orient}. Following tool-using language agents~\cite{yao2022react,schick2023toolformerlanguagemodelsteach}, spatial agents compose these modules into explicit computations. VADAR~\cite{marsili2025visualagenticai} lets the agent synthesize new API functions for 3D queries, pySpatial~\cite{luo2026pyspatialvisualprograms} generates 3D visual programs zero-shot, and GCA~\cite{chen2026geometrically} first formalizes a query as geometric constraints and then solves it under them. SpaceTools~\cite{chen2026spacetoolstoolaugmentedspatialreasoning} and S-Agent~\cite{dai2026sagentspatialtooluseelicits} train the policy to call spatial tools, SpatialClaw~\cite{cho2026spatialclawrethinkingactioninterface} redesigns the action interface through which the agent operates, and SpatialCLI~\cite{zhou2026spatialclireasonspatialtools} learns to reason with spatial tools and then without them. General multimodal agents follow the same trend~\cite{hong2025deepeyesv2,guo2025octopus}, and dedicated benchmarks evaluate tool-enabled visual reasoning~\cite{ashraf2025agent,guo2025beyond}. These agents either require training or approach each task anew. \textsc{SpatialCraft} adds a knowledge base that a frozen agent accumulates across tasks, recording when and how to orchestrate the same tools.

\paragraph{Memory and self-evolving agents.}
LLM agents extend their context with external memory, through hierarchical memory management~\cite{packer2024memgptllmsoperatingsystems,yu2026agenticmemorylearningunified}, long-term memory stores~\cite{zhong2023memorybankenhancinglargelanguage,chhikara2025mem0buildingproductionreadyai,xu2025amemagenticmemoryllm}, retrieval-augmented generation~\cite{lewis2021retrievalaugmentedgenerationknowledgeintensivenlp,borgeaud2022improvinglanguagemodelsretrieving}, and the memory streams of generative agents~\cite{park2023generativeagentsinteractivesimulacra}; surveys summarize this design space~\cite{zhang2024surveymemorymechanismlarge,hu2025memory,hatalis2025review}. Self-evolving agents use such memory to improve without parameter updates. They store reflective insights~\cite{shinn2023reflexion,zhao2024expel,ouyang2026reasoningbank,tang2025agent} or evolving context playbooks~\cite{zhang2026agentic}, induce reusable workflows~\cite{wang2024agent,fang2026mempexploringagentprocedural}, and build libraries of procedural skills, written as executable code~\cite{wang2023voyageropenendedembodiedagent,wang2025inducing,zheng2025skillweaver} or as natural-language procedures~\cite{anthropic2026skills,mi2026skill}. XSkill~\cite{jiang2026xskillcontinuallearningexperience} combines action-level experiences with task-level skills in multimodal agents, and multimodal memory has been developed for visual skills~\cite{wang2026atlasvaselfevolvingvisualskill} and semantic knowledge~\cite{bo2025agenticlearnergrowrefinemultimodal}. Closest to our setting, SMA~\cite{zhang2026spatial} accumulates experience-grounded procedural memory for spatial reasoning without tools. \textsc{SpatialCraft} differs in how the two forms of knowledge interact: an Experience is anchored to the Skill it serves, is injected only while that Skill is active, and, once proven useful, is consolidated into its procedure. This last step parallels the proceduralization of declarative knowledge in human skill acquisition~\cite{anderson1982acquisition}.

\clearpage
\newpage

\clearpage
\newpage

\clearpage
\newpage

\clearpage
\newpage
 
\clearpage
\newpage

\section{Tool APIs}
\label{app:tool_definitions}

\subsection{\texttt{tools.Reconstruct}}
\label{app:tool_reconstruct}
\apicode{tools.Reconstruct} is a client wrapper for the perception serving role. Its single entry method calls Depth Anything 3~\cite{lin2025depth} on a frame batch and returns a \texttt{Reconstruction} object that exposes per-frame depth, camera intrinsics, camera-to-world extrinsics, a dense per-frame point map in world coordinates, and a top-down rendering helper function. The output coordinate system is gravity-aligned with respect to the first camera: $+Y$ points up, the ground plane sits at $Y \approx 0$, and the first camera looks toward $-Z$. The output spatial resolution matches the input frame resolution, so masks from \apicode{tools.SAM3} can be combined directly with the reconstructed point maps without resizing. Table~\ref{tab:reconstruct_api} summarizes the interface.

\begin{table*}[ht]
\centering
\caption{Interface of \texttt{tools.Reconstruct} and its returned
\texttt{Reconstruction} object. Throughout, \texttt{fi} denotes an absolute
video frame index, rather than a position in the input list.}
\label{tab:reconstruct_api}
\begingroup
\small
\setlength{\tabcolsep}{5pt}
\renewcommand{\arraystretch}{1.15}
\begin{tabularx}{\linewidth}{@{}
  >{\raggedright\arraybackslash}p{0.20\linewidth}
  >{\raggedright\arraybackslash}p{0.23\linewidth}
  >{\raggedright\arraybackslash}X@{}}
\toprule
\textbf{Field} & \textbf{Type / shape} & \textbf{Description} \\
\midrule
\multicolumn{3}{@{}l}{\textbf{A. Reconstruction call}} \\
\rowcolor{gray!6}\multicolumn{3}{@{}l}{\texttt{Reconstruct(frames, frame\_indices=None)}} \\
\addlinespace[3pt]
\texttt{frames}
& List of \texttt{FrameImage} or PIL images
& Input batch, with at most \texttt{reconstruct\_max\_frames} entries
  (64 in our experiments). PIL denotes Python Imaging Library. \\
\texttt{frame\_indices}
& Optional frame indices
& For \texttt{FrameImage} inputs, absolute indices are extracted from
  each image's \texttt{frame\_index} attribute, so this argument may be omitted. \\
Return value
& \texttt{Reconstruction}
& Per-frame geometry, camera parameters, and a top-down rendering helper. \\
\midrule
\multicolumn{3}{@{}l}{\textbf{B. Reconstruction attributes}} \\
\addlinespace[3pt]
\texttt{frame\_indices}
& \texttt{list[int]}
& Absolute frame indices in the same order as \texttt{frames}. \\
\texttt{num\_frames}
& \texttt{int}
& Number of reconstructed frames. \\
\texttt{metric\_scale}
& \texttt{float}
& Metric scale value. \\
\texttt{depth[fi]}
& $(H,W)$; \texttt{float32}
& Depth map at absolute frame \texttt{fi}. \\
\texttt{extrinsics[fi]}
& $(4,4)$; \texttt{float64}
& Camera-to-world $\mathrm{SE}(3)$ matrix in OpenCV convention. \\
\texttt{intrinsics[fi]}
& Dictionary
& Camera intrinsics with keys \texttt{fx}, \texttt{fy}, \texttt{cx}, and \texttt{cy}. \\
\texttt{points[fi]}
& $(H,W,3)$; \texttt{float32}
& Dense per-frame point map in world coordinates. \\
\midrule
\rowcolor{gray!6}\multicolumn{3}{@{}l}{\textbf{C. Bird's-eye-view rendering}} \\
\multicolumn{3}{@{}l}{\texttt{recon.render\_bev(masks=None, labels=None, ref\_frame=0, ego\_trajectory=True)}} \\
\addlinespace[3pt]
\texttt{masks}
& \texttt{PerFrameMask} or boolean array
& Per-frame object masks; raw arrays have shape $(N,N_{\mathrm{obj}},H,W)$. \\
\texttt{labels}
& Object names
& Per-object names when masks are supplied as a raw array. \\
\texttt{ref\_frame}
& Absolute frame index
& Reference camera frame; defaults to \texttt{0}. \\
\texttt{ego\_trajectory}
& \texttt{bool}
& Toggles drawing of the camera path; defaults to \texttt{True}. \\
Return value
& \texttt{VisualFeedback}
& Inspected by the agent with \texttt{show()}. Stationary objects use oriented
  bounding boxes; moving objects use colour-graded trajectory lines. \\
\bottomrule
\end{tabularx}
\endgroup
\end{table*}

\subsection{\texttt{tools.SAM3}}
\label{app:tool_sam3}
\texttt{tools.SAM3} is a client wrapper for the perception serving role
and exposes single-frame and video-mode segmentation against
SAM3~\cite{carion2026sam}, together with an object-existence check.
All segmentation methods return a \texttt{PerFrameMask} whose absolute
video frame indices align with those of any \texttt{Reconstruction}
the agent has built. Table~\ref{tab:sam3_api} summarizes the image-mode
methods, existence check, and mask interface. Each image-mode method
operates on a single PIL image and returns masks for one frame.

\begin{table*}[ht]
\centering
\caption{Interface of \texttt{tools.SAM3} and its returned
\texttt{PerFrameMask} object. Throughout, \texttt{fi} is an absolute
video frame index, and \texttt{k} is an object index or string label.}
\label{tab:sam3_api}
\begingroup
\small
\setlength{\tabcolsep}{5pt}
\renewcommand{\arraystretch}{1.15}
\begin{tabularx}{\linewidth}{@{}
  >{\raggedright\arraybackslash}p{0.20\linewidth}
  >{\raggedright\arraybackslash}p{0.23\linewidth}
  >{\raggedright\arraybackslash}X@{}}
\toprule
\textbf{Field} & \textbf{Type / shape} & \textbf{Description} \\
\midrule
\multicolumn{3}{@{}l}{\textbf{A. Image-mode segmentation}} \\
\addlinespace[3pt]
\rowcolor{gray!6}\multicolumn{3}{@{}l}{\texttt{segment\_image\_by\_text(image, prompt, label=None)}} \\
Inputs & PIL image; text
& \texttt{image}: input image; \texttt{prompt}: free-form text description. \\
Return value & \texttt{PerFrameMask}
& One channel per detected instance. \\
\addlinespace[4pt]
\rowcolor{gray!6}\multicolumn{3}{@{}l}{\texttt{segment\_image\_by\_points(image, points, point\_labels, label="object")}} \\
Inputs & PIL image; point and label lists
& \texttt{points}: \texttt{[x, y]} pixel coordinates;
  \texttt{point\_labels}: parallel list with $1$ for foreground and $0$ for background. \\
Return value & \texttt{PerFrameMask}
& Mask for the prompted object. \\
\addlinespace[4pt]
\rowcolor{gray!6}\multicolumn{3}{@{}l}{\texttt{segment\_image\_by\_box(image, box, label="object")}} \\
Inputs & PIL image; bounding box
& \texttt{box}: \texttt{[x1, y1, x2, y2]} in pixels, using the xyxy convention. \\
Return value & \texttt{PerFrameMask}
& Mask for the boxed object. \\
\midrule
\multicolumn{3}{@{}l}{\textbf{B. Object-existence check}} \\
\rowcolor{gray!6}\multicolumn{3}{@{}l}{\texttt{is\_object\_exist(images, object\_name)}} \\
\addlinespace[3pt]
Inputs & List of PIL images; text
& \texttt{images}: input images; \texttt{object\_name}: text description. \\
Return value & Dictionary
& \texttt{exists}: list of booleans;
  \texttt{counts}: list of instance counts per image;
  \texttt{summary}: short text summary. \\
\midrule
\multicolumn{3}{@{}l}{\textbf{C. PerFrameMask attributes}} \\
\addlinespace[3pt]
\texttt{frame\_indices} & Frame indices
& Absolute video frame indices aligned with \texttt{Reconstruction}. \\
\texttt{labels} & Object labels
& Labels associated with the segmented objects. \\
\texttt{num\_frames} & Frame count
& Number of frames in the mask object. \\
\texttt{num\_objects} & Object count
& Number of objects in the mask object. \\
\midrule
\multicolumn{3}{@{}l}{\textbf{D. PerFrameMask access patterns}} \\
\addlinespace[3pt]
\rowcolor{gray!6}\multicolumn{3}{@{}l}{\texttt{seg.get\_mask(frame=fi, object=k)}} \\
Return value & $(H,W)$; boolean
& 2D mask of object \texttt{k} at frame \texttt{fi}. \\
\addlinespace[4pt]
\rowcolor{gray!6}\multicolumn{3}{@{}l}{\texttt{seg[fi]}} \\
Return value & $(N_{\mathrm{obj}},H,W)$; boolean
& Stack of object masks at frame \texttt{fi}. \\
\addlinespace[4pt]
\rowcolor{gray!6}\multicolumn{3}{@{}l}{\texttt{seg.get\_centroid\_3d(recon, frame=fi, object=k)}} \\
Return value & 3D position or \texttt{None}
& Median world position of the masked points, or \texttt{None}. \\
\addlinespace[4pt]
\rowcolor{gray!6}\multicolumn{3}{@{}l}{\texttt{seg.get\_masked\_points(recon, frame=fi, object=k)}} \\
Return value & $(K,3)$ array
& Masked points in world coordinates. \\
\addlinespace[4pt]
\rowcolor{gray!6}\multicolumn{3}{@{}l}{\texttt{seg.visualize(fi)}} \\
Return value & \texttt{VisualFeedback}
& Mask overlay for verification with \texttt{show()}. \\
\bottomrule
\end{tabularx}
\endgroup
\end{table*}

\subsection{\texttt{tools.Geometry}}
\label{app:tool_geometry}
\texttt{tools.Geometry} is a static class that provides the numerical
primitives most commonly needed for spatial reasoning over reconstructed
scenes. All methods operate on \texttt{numpy} arrays.
Table~\ref{tab:geometry_api} summarizes their inputs and outputs.

\begin{table*}[ht]
\centering
\caption{Numerical primitives provided by \texttt{tools.Geometry}
for spatial reasoning over reconstructed scenes.}
\label{tab:geometry_api}
\begingroup
\small
\setlength{\tabcolsep}{5pt}
\renewcommand{\arraystretch}{1.15}
\begin{tabularx}{\linewidth}{@{}
  >{\raggedright\arraybackslash}p{0.20\linewidth}
  >{\raggedright\arraybackslash}p{0.23\linewidth}
  >{\raggedright\arraybackslash}X@{}}
\toprule
\textbf{Field} & \textbf{Type / shape} & \textbf{Description} \\
\midrule
\multicolumn{3}{@{}l}{\textbf{A. Distances and angles}} \\
\addlinespace[3pt]
\rowcolor{gray!6}\multicolumn{3}{@{}>{\raggedright\arraybackslash}p{\linewidth}@{}}{%
  \texttt{euclidean\_distance(p1, p2)}} \\
Inputs & Two $(3,)$ arrays
& Two 1D arrays of length 3. \\
Return value & \texttt{float}
& Euclidean distance between \texttt{p1} and \texttt{p2}. \\
\addlinespace[4pt]
\rowcolor{gray!6}\multicolumn{3}{@{}>{\raggedright\arraybackslash}p{\linewidth}@{}}{%
  \texttt{angle\_between\_vectors(v1, v2)}} \\
Inputs & Two 3D vectors
& Vectors \texttt{v1} and \texttt{v2}. \\
Return value & \texttt{float}
& Angle between the vectors, in degrees. \\
\midrule
\multicolumn{3}{@{}l}{\textbf{B. Projection and transformations}} \\
\addlinespace[3pt]
\rowcolor{gray!6}\multicolumn{3}{@{}>{\raggedright\arraybackslash}p{\linewidth}@{}}{%
  \texttt{project\_point\_to\_camera(point\_3d, c2w, fx, fy, cx, cy)}}\\
Inputs & 3D point; $(4,4)$ matrix; four scalars
& World point, camera-to-world $\mathrm{SE}(3)$ matrix,
  and intrinsics \texttt{fx}, \texttt{fy}, \texttt{cx}, \texttt{cy}. \\
Return value & \texttt{(u, v)} or \texttt{None}
& Pixel coordinates; \texttt{None} when the point is behind the camera. \\
\addlinespace[4pt]
\rowcolor{gray!6}\multicolumn{3}{@{}>{\raggedright\arraybackslash}p{\linewidth}@{}}{%
  \texttt{rotation\_matrix\_from\_vectors(v\_from, v\_to)}} \\
Inputs & Two 3D vectors
& Source and target vectors; unit length is not required. \\
Return value & $(3,3)$ matrix
& Rotation matrix aligning \texttt{v\_from} with \texttt{v\_to}. \\
\addlinespace[4pt]
\rowcolor{gray!6}\multicolumn{3}{@{}>{\raggedright\arraybackslash}p{\linewidth}@{}}{%
  \texttt{transform\_points(points, matrix)}} \\
Inputs & $(\ldots,3)$ array; $(4,4)$ matrix
& Point array and an $\mathrm{SE}(3)$ transformation matrix. \\
Return value & Same shape as input
& Transformed point array. \\
\midrule
\multicolumn{3}{@{}l}{\textbf{C. Ground-plane estimation}} \\
\addlinespace[3pt]
\rowcolor{gray!6}\multicolumn{3}{@{}>{\raggedright\arraybackslash}p{\linewidth}@{}}{%
  \texttt{fit\_ground\_plane\_ransac(points, confidence, conf\_threshold=0.3, n\_iterations=1000, inlier\_threshold=0.05)}}\\
Inputs & $(H,W,3)$ or $(N,3)$ point cloud
& Point cloud and matching confidence array.
  Thresholds and iteration count are optional, with defaults shown above. \\
Return value & Normal and inlier mask
& \texttt{(plane\_normal, inlier\_mask)} via
  RANSAC, or \texttt{(None, None)} if fitting fails.
  The normal's sign is unspecified and disambiguated by the caller. \\
\midrule
\multicolumn{3}{@{}l}{\textbf{D. Coordinate conversion}} \\
\addlinespace[3pt]
\rowcolor{gray!6}\multicolumn{3}{@{}>{\raggedright\arraybackslash}p{\linewidth}@{}}{%
  \texttt{normalized\_to\_pixel(coords, width, height)}} \\
Inputs & Coordinate sequence; image dimensions
& Coordinates on the 0--1000 normalized scale used by
  \texttt{vlm.locate}; image width and height in pixels. \\
Return value & \texttt{list[float]}
& Pixel coordinates. \\
\bottomrule
\end{tabularx}
\endgroup
\end{table*}

\subsection{\texttt{tools.Mask}}
\label{app:tool_mask}
\texttt{tools.Mask} is a static class that provides standard mask
statistics over the boolean mask arrays returned by \texttt{tools.SAM3}.
All inputs are 2D \texttt{numpy} arrays with shape $(H,W)$ unless
otherwise noted. Table~\ref{tab:mask_api} summarizes the interface.

\begin{table*}[ht]
\centering
\caption{Mask statistics and operations provided by \texttt{tools.Mask}.
Coordinates are expressed in pixels.}
\label{tab:mask_api}
\begingroup
\small
\setlength{\tabcolsep}{5pt}
\renewcommand{\arraystretch}{1.15}
\begin{tabularx}{\linewidth}{@{}
  >{\raggedright\arraybackslash}p{0.20\linewidth}
  >{\raggedright\arraybackslash}p{0.23\linewidth}
  >{\raggedright\arraybackslash}X@{}}
\toprule
\textbf{Field} & \textbf{Type / shape} & \textbf{Description} \\
\midrule
\multicolumn{3}{@{}l}{\textbf{A. Centroids and area}} \\
\addlinespace[3pt]
\rowcolor{gray!6}\multicolumn{3}{@{}>{\raggedright\arraybackslash}p{\linewidth}@{}}{%
  \texttt{centroid(mask)}} \\
Input & $(H,W)$; boolean
& A 2D boolean mask. \\
Return value & \texttt{(cx, cy)}
& Median pixel coordinates of the mask;
  \texttt{(nan, nan)} when the mask is empty. \\
\addlinespace[4pt]
\rowcolor{gray!6}\multicolumn{3}{@{}>{\raggedright\arraybackslash}p{\linewidth}@{}}{%
  \texttt{centroids(masks)}} \\
Input & $(N,H,W)$; boolean
& A batch of $N$ boolean masks. \\
Return value & List of \texttt{(cx, cy)} pairs
& One centroid per element of the batch. \\
\addlinespace[4pt]
\rowcolor{gray!6}\multicolumn{3}{@{}>{\raggedright\arraybackslash}p{\linewidth}@{}}{%
  \texttt{area(mask)}} \\
Input & $(H,W)$; boolean
& A 2D boolean mask. \\
Return value & \texttt{int}
& Number of \texttt{True} pixels. \\
\midrule
\multicolumn{3}{@{}l}{\textbf{B. Bounding boxes}} \\
\addlinespace[3pt]
\rowcolor{gray!6}\multicolumn{3}{@{}>{\raggedright\arraybackslash}p{\linewidth}@{}}{%
  \texttt{bounding\_box(mask)}} \\
Input & $(H,W)$; boolean
& A 2D boolean mask. \\
Return value & Coordinate tuple or \texttt{None}
& \texttt{(x1, y1, x2, y2)} in pixels, or \texttt{None} for an empty mask.
  When the mask contains more than 100 \texttt{True} pixels, the box is
  computed from the 1st and 99th percentiles to discard outliers. \\
\addlinespace[4pt]
\rowcolor{gray!6}\multicolumn{3}{@{}>{\raggedright\arraybackslash}p{\linewidth}@{}}{%
  \texttt{mask\_to\_bbox(mask)}} \\
Input & $(H,W)$; boolean
& A 2D boolean mask. \\
Return value & $(4,)$ \texttt{numpy} array or \texttt{None}
& \texttt{[x1, y1, x2, y2]} in pixels, or \texttt{None} for an empty mask. \\
\midrule
\multicolumn{3}{@{}l}{\textbf{C. Overlap and intersection}} \\
\addlinespace[3pt]
\rowcolor{gray!6}\multicolumn{3}{@{}>{\raggedright\arraybackslash}p{\linewidth}@{}}{%
  \texttt{iou(mask\_a, mask\_b)}} \\
Inputs & Two $(H,W)$ boolean masks
& Masks of identical shape. \\
Return value & \texttt{float}
& Intersection over union. \\
\addlinespace[4pt]
\rowcolor{gray!6}\multicolumn{3}{@{}>{\raggedright\arraybackslash}p{\linewidth}@{}}{%
  \texttt{intersection(mask\_a, mask\_b)}} \\
Inputs & Two $(H,W)$ boolean masks
& Masks of identical shape. \\
Return value & $(H,W)$; boolean
& Element-wise AND of the input masks, with the same shape as the inputs. \\
\bottomrule
\end{tabularx}
\endgroup
\end{table*}

\subsection{\texttt{tools.Draw}}
\label{app:tool_draw}
\texttt{tools.Draw} is a static class that produces PIL-based annotation
 overlays. All methods accept a \texttt{numpy} $(H,W,3)$ \texttt{uint8}
array, a PIL image, or an \texttt{InputImages[i]} entry, and return a PIL
image of the same size; the input image is never modified. Colours may
be passed as an \texttt{(R, G, B)} \texttt{uint8} tuple or as any
\texttt{matplotlib} or CSS4 colour name. Each method accepts either a
single primitive or a list of primitives, with a matching list of colours
when per-primitive colouring is desired.
Table~\ref{tab:draw_api} summarizes the drawing methods.

\begin{table*}[ht]
\centering
\caption{Annotation overlays provided by \texttt{tools.Draw}.
All primitive coordinates are expressed in pixels.}
\label{tab:draw_api}
\begingroup
\small
\setlength{\tabcolsep}{5pt}
\renewcommand{\arraystretch}{1.15}
\begin{tabularx}{\linewidth}{@{}
  >{\raggedright\arraybackslash}p{0.20\linewidth}
  >{\raggedright\arraybackslash}p{0.23\linewidth}
  >{\raggedright\arraybackslash}X@{}}
\toprule
\textbf{Field} & \textbf{Type / shape} & \textbf{Description} \\
\midrule
\multicolumn{3}{@{}l}{\textbf{A. Bounding-box overlays}} \\
\addlinespace[3pt]
\rowcolor{gray!6}\multicolumn{3}{@{}>{\raggedright\arraybackslash}p{\linewidth}@{}}{%
  \texttt{draw\_bbox(image, bboxes, colors=None, thickness=None)}} \\
\texttt{bboxes} & Single box or list of boxes
& Each box is \texttt{(x1, y1, x2, y2)} in pixels. \\
Return value & PIL image
& Axis-aligned rectangle outlines drawn on a new image of the same size. \\
\midrule
\multicolumn{3}{@{}l}{\textbf{B. Line overlays}} \\
\addlinespace[3pt]
\rowcolor{gray!6}\multicolumn{3}{@{}>{\raggedright\arraybackslash}p{\linewidth}@{}}{%
  \texttt{draw\_line(image, lines, colors=None, thickness=None)}} \\
\texttt{lines} & Single segment or list of segments
& Each segment is \texttt{(x1, y1, x2, y2)} in pixels. \\
Return value & PIL image
& Line segments drawn on a new image of the same size. \\
\midrule
\multicolumn{3}{@{}l}{\textbf{C. Point overlays}} \\
\addlinespace[3pt]
\rowcolor{gray!6}\multicolumn{3}{@{}>{\raggedright\arraybackslash}p{\linewidth}@{}}{%
  \texttt{draw\_point(image, points, colors=None, radius=None)}} \\
\texttt{points} & Single point or list of points
& Each point is \texttt{(x, y)} in pixels. \\
Return value & PIL image
& Filled circles drawn on a new image of the same size. \\
\bottomrule
\end{tabularx}
\endgroup
\end{table*}

\subsection{\texttt{tools.Detect}}
\label{app:tool_detect}

\texttt{tools.Detect} is a proposed client wrapper for an open-vocabulary
 detector such as GroundingDINO~\cite{liu2024grounding}. It complements
\texttt{vlm.locate} and text-prompted SAM3 segmentation by exposing
 detector-native boxes, labels, and confidence scores without requiring
 a separate VLM session. The output boxes use pixel-space \texttt{xyxy}
 coordinates and can be passed directly to
\texttt{tools.SAM3.}\allowbreak\texttt{segment\_image\_by\_box}.
Table~\ref{tab:detect_api} summarizes the proposed interface and its
 composition with SAM3.

\begin{table*}[ht]
\centering
\caption{Proposed interface of \texttt{tools.Detect}, returned detection fields, and composition with SAM3. Here, $N$ is the number of retained detections, and \texttt{fi} is an absolute frame index.}
\label{tab:detect_api}
\begingroup
\small
\setlength{\tabcolsep}{5pt}
\renewcommand{\arraystretch}{1.15}
\begin{tabularx}{\linewidth}{@{}
  >{\raggedright\arraybackslash}p{0.20\linewidth}
  >{\raggedright\arraybackslash}p{0.23\linewidth}
  >{\raggedright\arraybackslash}X@{}}
\toprule
\textbf{Field} & \textbf{Type / shape} & \textbf{Description} \\
\midrule
\multicolumn{3}{@{}l}{\textbf{A. Detection call}} \\
\addlinespace[3pt]
\rowcolor{gray!6}\multicolumn{3}{@{}>{\raggedright\arraybackslash}p{\linewidth}@{}}{%
  \texttt{detect(image, prompt, box\_threshold=0.35, text\_threshold=0.25)}} \\
\texttt{image} & PIL image, \texttt{FrameImage}, or \texttt{InputImages[i]} entry
& Input image. \\
\texttt{prompt} & String or list of text descriptions
& Open-vocabulary detection prompt. \\
\texttt{box\_threshold} & Confidence threshold
& Detector confidence filtering; defaults to \texttt{0.35}. \\
\texttt{text\_threshold} & Confidence threshold
& Detector confidence filtering; defaults to \texttt{0.25}. \\
Return value & \texttt{DetectionResult}
& Empty result when no candidate survives thresholding. \\
\midrule
\multicolumn{3}{@{}l}{\textbf{B. DetectionResult attributes and visualization}} \\
\addlinespace[3pt]
\texttt{boxes} & $(N,4)$; \texttt{float32}
& Array of \texttt{[x1, y1, x2, y2]} boxes in pixels. \\
\texttt{labels} & \texttt{list[str]}; length $N$
& Labels of the retained detections. \\
\texttt{scores} & $(N,)$; \texttt{float32}
& Detector confidence scores. \\
\texttt{num\_objects} & \texttt{int}
& Number of detections, equal to $N$. \\
\addlinespace[4pt]
\rowcolor{gray!6}\multicolumn{3}{@{}>{\raggedright\arraybackslash}p{\linewidth}@{}}{%
  \texttt{result.visualize(image)}} \\
Return value & \texttt{VisualFeedback}
& Overlay with indexed boxes, labels, and scores for inspection with \texttt{show()}. \\
\bottomrule
\end{tabularx}
\endgroup
\end{table*}

\subsection{\texttt{tools.Pose}}
\label{app:tool_pose}

\texttt{tools.Pose} is a proposed object-centric pose tool that combines
 a reconstructed object point set with a single-image semantic orientation
 estimator such as Orient Anything~\cite{wang2024orient}. It estimates
 a semantic object frame rather than only an unoriented 3D bounding box.
Following GCA~\cite{chen2026geometrically}, the local origin is placed
 at the robust centroid of the masked object points,
$+Z_{\mathrm{obj}}$ points toward the object's semantic front,
$+Y_{\mathrm{obj}}$ points toward its semantic bottom, and
$+X_{\mathrm{obj}}$ follows the right-hand rule. This object frame enables
 perspective-taking and reference-frame reasoning such as ``left of the
 chair from the chair's viewpoint.''
Tables~\ref{tab:pose_api} and~\ref{tab:object_pose_attributes} summarize
 the proposed interface and returned object.

\begin{table*}[ht]
\centering
\caption{Proposed pose estimation and visualization interface of \texttt{tools.Pose}.}
\label{tab:pose_api}
\begingroup
\footnotesize
\setlength{\tabcolsep}{5pt}
\renewcommand{\arraystretch}{1.15}
\begin{tabularx}{\linewidth}{@{}
  >{\raggedright\arraybackslash}p{0.20\linewidth}
  >{\raggedright\arraybackslash}p{0.23\linewidth}
  >{\raggedright\arraybackslash}X@{}}
\toprule
\textbf{Field} & \textbf{Type / shape} & \textbf{Description} \\
\midrule
\multicolumn{3}{@{}l}{\textbf{A. Object-pose estimation}} \\
\addlinespace[3pt]
\rowcolor{gray!6}\multicolumn{3}{@{}>{\raggedright\arraybackslash}p{\linewidth}@{}}{%
  \texttt{predict\_obj\_pose(image, recon, seg, frame, object=0, label="object")}} \\
\texttt{image} & PIL image or \texttt{FrameImage}
& Image used for orientation estimation. \\
\texttt{recon} & \texttt{Reconstruction}
& Corresponding reconstructed scene. \\
\texttt{seg} & Single-frame \texttt{PerFrameMask}
& Aligned with \texttt{recon}. Whenever possible, the selected mask
  should contain one complete, non-occluded object. \\
\texttt{frame} & Absolute frame index
& Frame used for orientation estimation. \\
\texttt{object} & Channel index or string label
& Selects the object mask; defaults to \texttt{0}. \\
\texttt{label} & Display string
& Human-readable name for visualization; defaults to \texttt{"object"}. \\
Return value & \texttt{ObjectPose} or \texttt{None}
& Returns \texttt{None} if too few valid reconstructed points remain,
  or if the orientation model determines that the object has no
  meaningful semantic front. \\
\midrule
\multicolumn{3}{@{}l}{\textbf{B. Pose visualization}} \\
\addlinespace[3pt]
\rowcolor{gray!6}\multicolumn{3}{@{}>{\raggedright\arraybackslash}p{\linewidth}@{}}{%
  \texttt{pose.visualize(recon, frame=None)}} \\
\texttt{recon} & \texttt{Reconstruction}
& Source reconstruction. \\
\texttt{frame} & Optional camera frame
& Selects the camera used for projection; defaults to
  \texttt{pose.frame\_index} when omitted. \\
Return value & \texttt{VisualFeedback}
& Overlay of projected object axes, centroid, and object-aligned bounding
  box. Intended for verification with \texttt{show()} before the pose is
  used in downstream computation. \\
\bottomrule
\end{tabularx}
\endgroup
\end{table*}

\begin{table*}[ht]
\centering
\caption{Attributes of the proposed \texttt{ObjectPose} object. The object-aligned bounding box is expressed in world coordinates.}
\label{tab:object_pose_attributes}
\begingroup
\small
\setlength{\tabcolsep}{5pt}
\renewcommand{\arraystretch}{1.15}
\begin{tabularx}{\linewidth}{@{}
  >{\raggedright\arraybackslash}p{0.20\linewidth}
  >{\raggedright\arraybackslash}p{0.23\linewidth}
  >{\raggedright\arraybackslash}X@{}}
\toprule
\textbf{Field} & \textbf{Type / shape} & \textbf{Description} \\
\midrule
\multicolumn{3}{@{}l}{\textbf{A. Frame and identity}} \\
\addlinespace[3pt]
\texttt{frame\_index} & \texttt{int}
& Absolute frame used for orientation estimation. \\
\texttt{label} & \texttt{str}
& Object label. \\
\midrule
\multicolumn{3}{@{}l}{\textbf{B. Semantic object frame}} \\
\addlinespace[3pt]
\texttt{centroid} & $(3,)$; \texttt{float32}
& Robust object centroid in world coordinates. \\
\texttt{rotation} & $(3,3)$; \texttt{float64}
& Object-to-world rotation matrix. Columns are the local $+X$, $+Y$,
  and $+Z$ axes expressed in world coordinates. \\
\texttt{object\_to\_world} & $(4,4)$; \texttt{float64}
& $\mathrm{SE}(3)$ homogeneous transformation matrix with
  \texttt{rotation} and \texttt{centroid} as its rotational and
  translational components. \\
\texttt{front} & $(3,)$; \texttt{float64}
& Normalized semantic-front vector $+Z_{\mathrm{obj}}$ in world coordinates. \\
\texttt{orientation\_}\allowbreak\texttt{confidence} & \texttt{float} in $[0,1]$
& Available when provided by the orientation backend. \\
\midrule
\multicolumn{3}{@{}l}{\textbf{C. Object-aligned bounding box}} \\
\addlinespace[3pt]
\texttt{obb\_center} & $(3,)$; \texttt{float32}
& Object-aligned bounding-box center. \\
\texttt{obb\_extent} & $(3,)$; \texttt{float32}
& Object-aligned bounding-box extent. \\
\texttt{obb\_corners} & $(8,3)$; \texttt{float32}
& Eight corners of the object-aligned bounding box. \\
\bottomrule
\end{tabularx}
\endgroup
\end{table*}

\subsection{\texttt{tools.Scale}}
\label{app:tool_scale}

\texttt{tools.Scale} is an optional client wrapper for metric-scale
verification with a monocular metric-geometry model such as
MoGe-2~\cite{wang2026moge}. Its purpose is to estimate a multiplicative
correction factor and model disagreement, rather than to assume that
the original reconstruction is scale-free.
Table~\ref{tab:scale_api} summarizes the interface and explicit correction.

\begin{table*}[ht]
\centering
\caption{Metric-scale verification interface of \texttt{tools.Scale}
and its returned \texttt{ScaleEstimate}. The method does not mutate
\texttt{recon}.}
\label{tab:scale_api}
\begingroup
\small
\setlength{\tabcolsep}{5pt}
\renewcommand{\arraystretch}{1.15}
\begin{tabularx}{\linewidth}{@{}
  >{\raggedright\arraybackslash}p{0.20\linewidth}
  >{\raggedright\arraybackslash}p{0.23\linewidth}
  >{\raggedright\arraybackslash}X@{}}
\toprule
\textbf{Field} & \textbf{Type / shape} & \textbf{Description} \\
\midrule
\multicolumn{3}{@{}l}{\textbf{A. Scale estimation}} \\
\addlinespace[3pt]
\rowcolor{gray!6}\multicolumn{3}{@{}>{\raggedright\arraybackslash}p{\linewidth}@{}}{%
  \texttt{estimate\_scale(recon, frames, frame\_indices=None, max\_frames=3)}} \\
\texttt{recon} & \texttt{Reconstruction}
& Reconstruction to be checked. \\
\texttt{frames} & Input images
& Images corresponding to the reconstruction. \\
\texttt{frame\_indices} & Optional absolute frame indices
& Selects absolute frame indices; defaults to \texttt{None}. \\
\texttt{max\_frames} & View-count limit
& Limits the number of high-confidence views used for estimation;
  defaults to \texttt{3}. \\
Return value & \texttt{ScaleEstimate} or \texttt{None}
& Returns \texttt{None} when no frame contains enough mutually
  valid depth pixels. \\
\midrule
\multicolumn{3}{@{}l}{\textbf{B. ScaleEstimate attributes}} \\
\addlinespace[3pt]
\texttt{correction\_}\allowbreak\texttt{factor} & \texttt{float}
& Robust multiplicative factor $s$ such that
  $d_{\mathrm{corrected}}=s\,d_{\mathrm{recon}}$. \\
\texttt{per\_frame\_}\allowbreak\texttt{factors} & \texttt{dict[int, float]}
& Maps absolute frame indices to valid scale estimates. \\
\texttt{selected\_frames} & \texttt{list[int]}
& Absolute frame indices used for the final estimate. \\
\texttt{relative\_}\allowbreak\texttt{disagreement} & \texttt{float}
& Robust dispersion statistic over valid pixel-wise and per-frame depth
  ratios. Smaller values indicate stronger agreement between the
  reconstruction and metric-geometry backends. \\
\texttt{num\_valid\_pixels} & \texttt{int}
& Total number of pixels contributing to the estimate. \\
\midrule
\multicolumn{3}{@{}l}{\textbf{C. Applying the correction}} \\
\addlinespace[3pt]
\multicolumn{3}{@{}>{\raggedright\arraybackslash}p{\linewidth}@{}}{%
  Metric distances and reconstructed points can be corrected explicitly:} \\
\addlinespace[3pt]
\multicolumn{3}{@{}>{\raggedright\arraybackslash}p{\linewidth}@{}}{%
  \texttt{distance\_m = raw\_distance * scale.correction\_factor}} \\
\addlinespace[3pt]
\multicolumn{3}{@{}>{\raggedright\arraybackslash}p{\linewidth}@{}}{%
  \texttt{points\_m = recon.points[fi] * scale.correction\_factor}} \\
\bottomrule
\end{tabularx}
\endgroup
\end{table*}

\paragraph{When to apply the correction.}
The correction should be applied only when the cross-model disagreement
is acceptable and the task requires absolute metric quantities.
Relative directions, angles, rankings, and normalized trajectories
do not require this tool.

\newpage
\newlist{skillsteps}{enumerate}{1}
\setlist[skillsteps]{
    label=\arabic*.,
    leftmargin=*,
    topsep=0pt,
    partopsep=0pt,
    itemsep=2pt,
    parsep=0pt
}

\newpage
\onecolumn
\section{Prompt Design}
\label{app:prompt_design}
Detailed prompt design for each phase of the framework provided here.
\subsection{System Prompts}
\label{app:system_prompts}
\definecolor{boxblue}{HTML}{6FA8DC}
\definecolor{boxblueback}{HTML}{EEF6FC}

\begin{tcolorbox}[
    colframe=boxblue,
    colback=boxblueback,
    coltitle=white,
    title=\textbf{Prompt: Direct CoT}
]
\small
You are a spatial reasoning agent. Your goal is to answer spatial questions grounded in the provided visual observations.\\

\# INSTRUCTIONS:\\
1. \textbf{Understand the task}: Identify the queried objects, spatial relations, reference frame, and relevant visual evidence.\\
2. \textbf{Reason spatially}: Analyze geometric cues such as relative position, depth, distance, orientation, occlusion, scale, and viewpoint when relevant. Avoid relying on semantic or commonsense priors when they conflict with visual evidence.\\
3. \textbf{Verify}: Check that the predicted relation is consistent with the available observations and the specified reference frame.\\
4. \textbf{Answer}: Once confident, provide the final answer inside \verb|<answer>...</answer>| tags.\\

If the visual evidence is ambiguous, reason conservatively and use the most directly supported spatial interpretation.
\end{tcolorbox}

\begin{tcolorbox}[
colframe=boxblue,
colback=boxblueback,
coltitle=white,
title=\textbf{Prompt: Experience- and Skill-Augmented Spatial Decision Making},
breakable]
\small

\textbf{System Instruction:}\\
You are a tool-augmented spatial reasoning agent. Your goal is to solve spatial reasoning tasks by grounding each decision in the current visual state, reliable tool-derived evidence, reusable procedural Skills, and context-specific Experiences.

\textbf{Inputs:}
\begin{itemize}
    \item \textbf{Current State}: \verb|{state}|, including the task query, current visual observations, previous tool calls, and feedback history.
    \item \textbf{Active Skill}: \verb|{skill_text}|, the temporally extended procedure currently guiding the agent, or \texttt{None} if no Skill is active.
    \item \textbf{Global Experiences}: \verb|{global_experiences}|, condition--action guidance retrieved for this task and already rewritten for its query and images.
    \item \textbf{Anchored Experiences}: \verb|{anchored_experiences}|, condition--action guidance attached to the Active Skill, which applies only while that Skill is active.
    \item \textbf{Available Tools}: the spatial reasoning tools listed below.
\end{itemize}

\textbf{Available Spatial Tools:}\\
\texttt{Reconstruct}, \texttt{Scale}, \texttt{SAM3}, \texttt{Detect},
\texttt{Geometry}, \texttt{Pose}, \texttt{Mask}, \texttt{Draw}

\textbf{Instructions for Knowledge Usage:}
\begin{enumerate}
    \item \textbf{Follow Skill}: If a Skill is active, use its procedure as the primary multi-step guidance and identify which step the Current State has reached. If no Skill is active, reason from the Experiences and the evidence alone.
    \item \textbf{Apply Experiences}: Examine both sets of Experiences and apply only those whose triggering conditions match the Current State. Use them as local tactical guidance to refine tool selection, evidence gathering, or error recovery.
    \item \textbf{Ground in Evidence}: Treat Skills and Experiences as guidance rather than facts. Current visual observations and verified tool outputs take precedence whenever conflicts arise.
    \item \textbf{Act}: Select the single next action that best advances the active procedure while resolving the most relevant spatial uncertainty.
\end{enumerate}

\textbf{Spatial Reasoning Principles:}
\begin{itemize}
    \item Explicitly identify the relevant objects and reference frame before reasoning about spatial relations.
    \item Prefer direct geometric or tool-derived evidence over semantic or commonsense assumptions.
    \item Use dedicated tools for depth, pose, or metric quantities when those properties cannot be reliably inferred from raw appearance.
    \item If spatial cues conflict, gather or verify additional evidence before committing to a conclusion.
    \item Avoid unnecessary tool calls when the available evidence is already sufficient.
\end{itemize}

\textbf{Response Requirements:}\\
At each interaction step, output either one tool action or the final answer.

For a tool action:
\begin{verbatim}
<think>
skill_step: the current procedure step, or None
experience: applicable experience(s), or None
evidence_needed: the spatial uncertainty to resolve
</think>
<action>[Specific Tool Action]</action>
\end{verbatim}

When sufficient evidence has been obtained:
\begin{verbatim}
<think>
verify: concise consistency check using the available spatial evidence
</think>
<answer>[Final Answer]</answer>
\end{verbatim}

\textbf{Important:}
\begin{itemize}
    \item Output only one tool action per turn.
    \item Do not blindly follow an Experience that is irrelevant to the Current State.
    \item Keep following the Active Skill until the system replaces it; whether its termination condition holds is checked separately after each step.
    \item The final answer MUST be enclosed in complete \verb|<answer>...</answer>| tags.
\end{itemize}

\end{tcolorbox}

\begin{tcolorbox}[
colframe=boxblue,
colback=boxblueback,
coltitle=white,
title=\textbf{Prompt: Active Skill Termination},
breakable]
\small
You are a controller for a spatial reasoning agent. After each action taken while a Skill is active, your task is to determine whether the Skill's termination condition is satisfied, so that control returns to the agent.

\textbf{Inputs:}
\begin{itemize}
    \item \textbf{Current Spatial State}: \verb|{state}|, including the task query, visual observations, tool outputs, and interaction history.
    \item \textbf{Active Skill}:
    \begin{itemize}
        \item Name: \verb|{skill.name}|
        \item Execution Procedure: \verb|{skill.procedure}|
        \item Termination Condition: \verb|{skill.termination}|
    \end{itemize}
\end{itemize}

\textbf{Decision Criteria:}
\begin{enumerate}
    \item Return \texttt{TERMINATE} if the Termination Condition is satisfied by the current spatial state or by newly obtained tool evidence.
    \item Otherwise, return \texttt{CONTINUE}. Do not terminate because intermediate uncertainty remains, or because a procedure step has not yet produced the expected evidence.
\end{enumerate}
Judge only from the visual observations, tool outputs, and interaction history in the current state; do not assume actions or results that are not recorded there.

\textbf{Response Requirements:}\\
Return exactly one status tag with no additional explanation:
\begin{verbatim}
<skill_status>CONTINUE</skill_status>
\end{verbatim}
or
\begin{verbatim}
<skill_status>TERMINATE</skill_status>
\end{verbatim}
\end{tcolorbox}

\begin{tcolorbox}[
colframe=boxblue,
colback=boxblueback,
coltitle=white,
title=\textbf{Prompt: Semantic Gradient Generation},
breakable]
\small

You are a procedural knowledge evaluator for a spatial reasoning agent.
Your task is to analyze the execution of an active Skill and generate structured semantic gradients that specify how the Skill should be refined based on the observed trajectory outcome.

\textbf{Inputs:}
\begin{itemize}
    \item \textbf{Skill Definition}: \verb|{skill_info}|, including the Activation Condition, Execution Procedure, and Termination Condition.
    \item \textbf{Skill-Conditioned Trajectory}: \verb|{trajectory}|, containing the states, visual observations, tool calls, tool outputs, and actions observed while the Skill was active.
    \item \textbf{Task Outcome}: \verb|{reward}|, the binary reward of the trajectory (1 if the final answer is correct, 0 otherwise).
\end{itemize}

\textbf{Analysis Instructions:}
\begin{enumerate}
    \item \textbf{Outcome Attribution}: Determine how the active Skill contributed to the observed outcome. Base the diagnosis on concrete evidence from the visual observations, tool feedback, spatial state transitions, and executed actions.

    \item \textbf{Component Diagnosis}: Attribute systematic weaknesses to the appropriate Skill component:
    \begin{itemize}
        \item \textbf{Activation Condition}: Refine when the Skill was selected for an unsuitable spatial situation, or when its applicability was defined too broadly or narrowly.
        \item \textbf{Execution Procedure}: Refine when the procedure selected inappropriate tools, ignored relevant spatial evidence, used an incorrect reference frame, or failed to resolve geometric uncertainty.
        \item \textbf{Termination Condition}: Refine when the Skill terminated before sufficient spatial evidence was obtained, or remained active after its intended reasoning objective had already been achieved.
    \end{itemize}
    For a successful trajectory, state which parts of the Skill contributed to the success and should be preserved, rather than proposing changes the evidence does not support.

    \item \textbf{Semantic Update}: For each affected component, provide a concise and actionable refinement direction that can generalize beyond the current task instance.
\end{enumerate}

\textbf{Constraints:}
\begin{itemize}
    \item Leave a component empty if the trajectory provides no reliable evidence that it should be changed.
    \item Set \texttt{is\_related} to \texttt{false} only when the Skill had no meaningful influence on the observed outcome.
    \item Do not encode task-specific answers, object identities, or instance-specific coordinates into the update.
    \item Prefer reusable spatial reasoning principles over corrections that merely fit the current trajectory.
\end{itemize}

\textbf{Response Requirements (JSON Format):}
\begin{verbatim}
{
    "diagnosis": "Concise evidence-grounded explanation of the outcome.",
    "is_related": true/false,
    "semantic_gradient": {
        "activation": "...",
        "procedure": "...",
        "termination": "..."
    }
}
\end{verbatim}

\end{tcolorbox}

\begin{tcolorbox}[
colframe=boxblue,
colback=boxblueback,
coltitle=white,
title=\textbf{Prompt: Skill Evolution from Aggregated Semantic Gradients},
breakable]
\small

You are a procedural skill optimizer for a spatial reasoning agent.
Your task is to update a Skill using aggregated semantic feedback derived from its execution trajectories and the graduated Experiences anchored to it, while preserving reusable reasoning structure and avoiding task-specific overfitting.

\textbf{Inputs:}
\begin{itemize}
    \item \textbf{Current Skill}: \verb|{old_skill_definition}|, including its Activation Condition, Execution Procedure, and Termination Condition.
    \item \textbf{Semantic Gradients}: \verb|{semantic_gradients}|, a collection of component-wise refinement signals obtained from multiple Skill-conditioned trajectories.
    \item \textbf{Graduated Experiences}: \verb|{graduated_experiences}|, condition--action Experiences anchored to this Skill whose contrastive utility has proven consistently positive.
    \item \textbf{Skill Pool}: \verb|{skill_pool_summary}|, the names and activation conditions of the other Skills in the pool.
\end{itemize}

\textbf{Optimization Instructions:}
\begin{enumerate}
    \item \textbf{Aggregate Feedback}: Identify recurring weaknesses and consistent improvement signals across the semantic gradients, and summarize them as one aggregated update direction $\bar{g}_k$ per component. Ignore contradictory, low-confidence, or trajectory-specific suggestions.

    \item \textbf{Refine Skill Structure}: Use $\bar{g}_k$ and the Graduated Experiences $\mathcal{G}_k$ to produce a candidate Skill
    \[
        k' = k \oplus (\bar{g}_k, \mathcal{G}_k).
    \]
    Update the Skill components as follows:
    \begin{itemize}
        \item \textbf{Activation Condition}: Refine when and under what spatial situations the Skill should be invoked.
        \item \textbf{Execution Procedure}: Revise the multi-step reasoning or tool-use procedure to address recurring spatial failure modes while preserving the Skill's core purpose.
        \item \textbf{Termination Condition}: Refine the stopping criterion so that the Skill terminates only when its intended spatial reasoning objective has been sufficiently achieved.
    \end{itemize}

    \item \textbf{Consolidate Experiences}: In REFINE mode, write each Graduated Experience into the procedure step, or the termination condition, at which its triggering condition arises, phrased as part of the procedure rather than appended as a separate tip.

    \item \textbf{Preserve Generality}: Remove instance-specific object names, coordinates, answers, or environment details unless they represent reusable procedural structure. Keep the Skill concise, within about 1{,}000 tokens.
\end{enumerate}

\textbf{Evolution Mode:}
\begin{itemize}
    \item \textbf{REFINE}: Improve the current Skill while preserving its procedural identity and intended role.
    \item \textbf{DISCOVER}: Create a new candidate Skill only when the aggregated evidence consistently reveals a reusable procedural pattern that is not adequately represented by the current Skill or any Skill in the Skill Pool. The new Skill is added alongside the current one, so do not absorb the Graduated Experiences in this mode.
\end{itemize}

\textbf{Response Requirements (JSON Format):}
\begin{verbatim}
{
    "mode": "REFINE" or "DISCOVER",
    "aggregated_gradient": {
        "activation": "...",
        "procedure": "...",
        "termination": "..."
    },
    "skill_name": "ConciseSkillName",
    "activation": "When ...",
    "procedure": [
        "Step 1: ...",
        "Step 2: ...",
        "Step 3: ..."
    ],
    "termination": "Terminate when ...",
    "absorbed_experiences": ["<experience id>: <step or termination>", ...]
}
\end{verbatim}

\end{tcolorbox}

\begin{tcolorbox}[
    breakable,
    colframe=boxblue,
colback=boxblueback,
coltitle=white,
    title={\textbf{Prompt: Experience Rollout Summary}}
]
\small

You are an expert trajectory analyst for a tool-augmented spatial
reasoning agent. Given several completed rollouts of the same task,
summarize how the agent used visual observations, spatial tools,
Experiences, and procedural Skills in each rollout to solve the task.

\textbf{Input:}

\begingroup
\ttfamily
\noindent
\detokenize{<task>{task}</task>}\par
\detokenize{<images>{task_images}</images>}\par
\detokenize{<rollouts>{trajectories}</rollouts>}\par
\detokenize{<verified_answer>{ground_truth}</verified_answer>}\par
\endgroup

\textbf{Summarization Guidelines:}

\begin{enumerate}[leftmargin=*, itemsep=1pt, topsep=3pt]

    \item \textbf{For each rollout}, state whether its final answer
    matches the verified answer, then, \textbf{for each interaction step:}

    \begin{itemize}[leftmargin=1.5em, itemsep=0pt, topsep=2pt]
        \item Describe the current spatial state, the action or tool
        call taken, and the key parameters or target entities involved.

        \item Explain what spatial uncertainty or reasoning need
        motivated the action and what new evidence was obtained from
        the resulting observation or tool output.

        \item Record the Active Skill (or \texttt{None}) and the
        identifiers of the Experiences shown to the agent at this step.

        \item Record whether any shown Experience was applied, and
        explain how it influenced local decision making.

        \item If an Active Skill was used, identify the relevant
        procedural step and explain how it guided the current action.
    \end{itemize}

    \item \textbf{Track spatial evidence and state transitions:}

    \begin{itemize}[leftmargin=1.5em, itemsep=0pt, topsep=2pt]
        \item Summarize the spatial evidence accumulated across the
        rollout, such as object locations, depth, distance, orientation,
        viewpoint, motion, scale, or reference-frame information.

        \item Explain how each new observation or tool result changed
        the agent's current belief or spatial state and influenced
        subsequent decisions.
    \end{itemize}

    \item \textbf{Analyze success and failure patterns:}

    Given the rollouts, the task images, and the verified answer, identify inefficient
    tool calls, unsupported spatial assumptions, incorrect reference
    frames, misleading visual cues, premature conclusions, unnecessary
    exploration, or other failure patterns. Explain their causes and
    their impact on the final outcome. Also note how Experiences,
    Skills, or later tool evidence corrected or mitigated these errors.

    \item \textbf{Preserve decision-critical information:}

    Retain the important outcomes of every step, including failed or
    redundant actions when they reveal useful evidence about the
    agent's reasoning process.

    \item \textbf{Analyze intermediate visual and geometric evidence:}

    \begin{itemize}[leftmargin=1.5em, itemsep=0pt, topsep=2pt]
        \item For intermediate visualizations or spatial representations
        produced by tools, such as segmentation masks, depth maps,
        3D reconstructions, bird's-eye views, pose overlays, motion
        fields, or geometric annotations, summarize the evidence they
        provide, checking them against the task images.

        \item Explain whether the evidence was interpreted correctly
        and whether additional spatial verification could have
        improved the decision.
    \end{itemize}

\end{enumerate}

\medskip

Provide a concise and structured summary, organized by rollout, that emphasizes
decision-critical spatial evidence, tool-use patterns, Experience
usage, Skill execution, and the causes of success or failure.

\end{tcolorbox}

\begin{tcolorbox}[
breakable,
    colframe=boxblue,
colback=boxblueback,
coltitle=white,
 title={\textbf{Prompt: Cross Rollout Critique}}
]
\small

You are an expert analyst for tool-augmented spatial reasoning.
Given multiple rollouts of the same spatial task, compare their outcomes and extract concise, reusable Experiences that can improve future local decisions.

\textbf{Input:}

\verb|<task>{question}</task>|

\verb|<rollout_summary>{summary}</rollout_summary>|

\verb|<experiences_used>{experiences}</experiences_used>|

\verb|<verified_answer>{groundtruth}</verified_answer>|

\textbf{Analysis Framework:}

\textit{1. Cross-Rollout Analysis:}
\begin{itemize}[leftmargin=1.5em, itemsep=0pt, topsep=3pt]
    \item Identify which decisions, spatial cues, or tool operations consistently contributed to successful reasoning.
    \item Identify failure patterns such as incorrect reference-frame assumptions, insufficient geometric evidence, inappropriate tool selection, misleading visual cues, or premature conclusions.
    \item Compare successful and failed rollouts to determine which differences were causally relevant to the outcome.
    \item Examine how the shown Experiences influenced decisions: determine which were followed, which were ignored, and whether any states its condition or action inaccurately.
    \item Analyze effective combinations of spatial tools and how their outputs were interpreted or verified.
\end{itemize}

\textit{2. Experience Extraction:}

Extract local condition--action guidance in two complementary categories:
\begin{itemize}[leftmargin=1.5em, itemsep=0pt, topsep=3pt]
    \item \textbf{Tool-Use Guidance}: Practical rules about when to invoke a spatial tool, what evidence to obtain, and how to interpret or verify its output.
    \item \textbf{Spatial Decision Guidance}: Context-sensitive rules for resolving ambiguity, selecting reference frames, gathering additional evidence, or checking spatial consistency.
\end{itemize}

For each proposed update, choose one of the following operations:
\begin{itemize}[leftmargin=1.5em, itemsep=0pt, topsep=3pt]
    \item \textbf{add}: Introduce a new Experience whose lesson is not covered by the Experiences used in this task; near-duplicates of other stored Experiences are merged automatically.
    \item \textbf{modify}: Refine an existing Experience when the observed rollouts reveal that its condition or recommended action should be made more accurate, specific, or actionable.
\end{itemize}

The \verb|<experiences_used>| field lists the Experiences shown during the rollouts, including global Experiences retrieved for this task and anchored Experiences that entered with an active Skill.
Evaluate them against the verified rollout outcomes only to decide whether they should be modified; whether an Experience is useful is estimated separately and is not your decision.

\textbf{Anchoring:} For each added Experience, set \texttt{anchor} to the Skill that the summary records as active at the step whose decision the Experience addresses. Set it to \texttt{none} if no Skill was active at that step or if the lesson spans several Skills. Use only Skill names recorded in the summary. A modified Experience keeps the anchor of the Experience it refines.

You may produce at most \texttt{\{max\_ops\}} updates.

\textit{3. Experience Quality Requirements:}
\begin{itemize}[leftmargin=1.5em, itemsep=0pt, topsep=3pt]
    \item Keep each Experience under 64 words.
    \item Express each Experience as a clear triggering condition followed by an actionable recommendation.
    \item Generalize across similar spatial situations rather than memorizing task-specific objects, coordinates, or answers.
    \item Prefer guidance grounded in observable visual or tool-derived evidence.
    \item Avoid high-level multi-step procedures that should instead be represented as Skills.
\end{itemize}

Conclude with the Experience updates in the following format:
\begin{verbatim}
[
  { "option": "add",
    "experience": {
      "condition": "When ...",
      "action": "..."
    },
    "anchor": "SkillName" or "none"
  },
  { "option": "modify",
    "experience": {
      "condition": "When ...",
      "action": "..."
    },
    "modified_from": "E17"
  }
]
\end{verbatim}

\end{tcolorbox}

\begin{tcolorbox}[
breakable,
colframe=boxblue,
colback=boxblueback,
coltitle=white,
 title={\textbf{Prompt: Experience Merge}}]
\small

You are responsible for consolidating a spatial reasoning Experience Bank.
A newly added Experience overlaps semantically with existing entries. Merge the following Experiences into a single compact and reusable Experience.

\textbf{Input} (each Experience is listed with its anchor Skill, or \texttt{none} if global):

\texttt{\{experiences\_text\}}

\textbf{Requirements:}
\begin{enumerate}[leftmargin=*, itemsep=1pt, topsep=3pt]
    \item Preserve all decision-critical spatial insights that are consistent across the input Experiences.
    \item Express the merged result as a concise condition--action rule: specify \textit{when} the guidance applies and \textit{what} the agent should do.
    \item Remove redundant, conflicting, or instance-specific details such as object identities, exact coordinates, or task-specific answers.
    \item Keep the result actionable, spatially grounded, and generalizable to similar situations.
    \item If the input Experiences have different anchors, the merged Experience becomes global; state its condition in terms of the observable situation rather than a step of a particular Skill.
    \item Keep the merged Experience within 64 words.
\end{enumerate}

Output ONLY:
\begin{verbatim}
{
  "experience": {
    "condition": "When ...",
    "action": "..."
  }
}
\end{verbatim}

\end{tcolorbox}

\begin{tcolorbox}[
colframe=boxblue,
colback=boxblueback,
coltitle=white,
 title={\textbf{Prompt: Experience Manage}},
 breakable
]
\small

You are responsible for maintaining the Experience Bank of a spatial reasoning agent.
The bank has exceeded its capacity and may contain Experiences that express substantially the same guidance. Perform a global merging pass that removes such redundancy while preserving useful and diverse spatial knowledge.

\textbf{Input} (each Experience is listed with its identifier and its anchor Skill, or \texttt{none} if global):

\verb|<experiences>{experiences}</experiences>|

\textbf{Current Bank Size:} \texttt{\{exp\_count\}} Experiences

\textbf{Capacity:} \texttt{\{capacity\}} Experiences

\textbf{Refinement Objectives:}
\begin{enumerate}[leftmargin=1.5em, itemsep=0pt, topsep=3pt]
    \item \textbf{Merge Semantic Redundancy}: Consolidate Experiences that express substantially the same triggering condition and recommended action while preserving complementary spatial insights.

    \item \textbf{Generalize While Merging}: When merging, abstract task-specific object identities, coordinates, viewpoints, or isolated observations into reusable spatial patterns supported by the merged Experiences.

    \item \textbf{Preserve Knowledge Diversity}: Retain Experiences covering distinct spatial situations, tool-use decisions, ambiguity-resolution strategies, and failure-recovery patterns.
\end{enumerate}

\textbf{Allowed Operation:}
\begin{itemize}[leftmargin=1.5em, itemsep=0pt, topsep=3pt]
    \item \textbf{merge}: Combine two or more semantically redundant Experiences into one generalized Experience and provide their IDs in \texttt{merged\_from}.
\end{itemize}

Do not delete Experiences. If the bank still exceeds capacity after merging, the system evicts entries in ascending order of their estimated utility.

\textbf{Experience Quality Criteria:}
\begin{itemize}[leftmargin=1.5em, itemsep=0pt, topsep=3pt]
    \item Each Experience should remain under 64 words and express concise condition--action guidance.
    \item The triggering condition should be grounded in observable visual states, tool-derived evidence, or spatial uncertainty.
    \item The recommended action should provide concrete guidance for spatial reasoning, tool selection, evidence verification, or error recovery.
    \item Avoid task-specific answers, unnecessary object identities, and exact coordinates unless structurally essential.
    \item If the merged Experiences have different anchors, the result becomes global; state its condition in terms of the observable situation rather than a step of a particular Skill.
    \item Do not merge local Experiences into a multi-step procedural workflow; such knowledge belongs to the Skill Pool.
\end{itemize}

\textbf{Important:}
\begin{itemize}[leftmargin=1.5em, itemsep=0pt, topsep=3pt]
    \item Merge only when redundancy is sufficiently clear.
    \item Preserve complementary Experiences even when they concern similar spatial situations but recommend different actions under different conditions.
    \item If no clear redundancy remains, return an empty list.
\end{itemize}

Return the merge operations in the following format:
\begin{verbatim}
[
  {
    "option": "merge",
    "experience": {
      "condition": "When ...",
      "action": "..."
    },
    "merged_from": ["E12", "E23"]
  }
]
\end{verbatim}
\end{tcolorbox}

\begin{tcolorbox}[
colframe=boxblue,
colback=boxblueback,
coltitle=white,
title={\textbf{Prompt: Task Decomposition}},
breakable]
\small

You are a spatial reasoning strategist. Your task is to decompose the current spatial reasoning problem into 2--3 abstract subtasks that can be used to retrieve relevant global Experiences from the Experience Bank.

\textbf{Task:} \texttt{\{task\_description\}}

\textbf{Visual Context:} \texttt{\{visual\_context\}}

Global Experiences are concise condition--action guidance learned from previous spatial reasoning trajectories and apply across procedures; Experiences tied to a specific Skill are provided separately when that Skill is active. The decomposition should therefore identify local reasoning needs rather than construct a complete multi-step solution plan.

\textbf{For each subtask, generate a JSON object with:}

\begin{itemize}[leftmargin=1.5em, itemsep=0pt, topsep=3pt]

    \item \textbf{Type}: A concise category describing the spatial reasoning need.

    \item \textbf{Query}: A short retrieval query describing the underlying technical situation and the kind of tactical guidance that would be useful.

\end{itemize}

\textbf{The decomposition may focus on subtasks such as:}

\begin{enumerate}[leftmargin=1.5em, itemsep=0pt, topsep=3pt]

    \item \textbf{Spatial Evidence Acquisition}: How to obtain missing evidence about depth, distance, orientation, motion, object extent, or reference frames.

    \item \textbf{Tool-Use Decisions}: When and how to apply spatial tools such as detection, segmentation, reconstruction, geometry, pose, or metric-scale verification.

    \item \textbf{Ambiguity Resolution}: How to handle conflicting cues, uncertain viewpoints, occlusion, weak geometric evidence, or insufficient observations.

    \item \textbf{Consistency Verification}: How to check whether an inferred spatial relation is supported by the available visual and tool-derived evidence.

\end{enumerate}

\textbf{Important:}

\begin{itemize}[leftmargin=1.5em, itemsep=0pt, topsep=3pt]

    \item Abstract away from task-specific object identities, exact coordinates, and final answers.

    \item Focus on the underlying spatial reasoning challenge rather than directly prescribing an entire procedural workflow.

    \item Each subtask should address a distinct reasoning need, so that different queries retrieve different Experiences.

    \item Each query should be specific enough to retrieve actionable Experiences but general enough to match related spatial situations.

\end{itemize}

\textbf{Required Output Format:}
\scriptsize
\begin{verbatim}
[
  {
    "type": "depth_ambiguity",
    "query": "Guidance for resolving uncertain depth relations using geometric evidence"
  },
  {
    "type": "reference_frame",
    "query": "Strategies for determining the correct viewpoint before judging directional relations"
  }
]
\end{verbatim}

\normalsize
Output ONLY the complete and valid JSON array.

\end{tcolorbox}

\begin{tcolorbox}[
colframe=boxblue,
colback=boxblueback,
coltitle=white,
title={\textbf{Prompt: Experience Rewrite}},
breakable]
\small

You are a spatial reasoning knowledge adaptor. Your task is to adapt retrieved global Experiences to the current task and visual context while preserving their original reusable insight.

\textbf{Task:} \texttt{\{task\_description\}}

\textbf{Visual Context:} \texttt{\{visual\_context\}}

\textbf{Retrieved Global Experiences:} \texttt{\{experiences\_text\}}

Each Experience represents local condition--action guidance. Rewrite only the Experiences that are relevant to the current spatial situation so that they can provide actionable tactical support to the execution agent. The rewritten text is used only for the current task; the stored Experiences are not modified.

\textbf{Rewriting Guidelines:}

\begin{enumerate}[leftmargin=1.5em, itemsep=0pt, topsep=3pt]

    \item \textbf{Preserve the Core Insight}: Retain the original condition--action relationship and do not introduce a new reasoning principle unsupported by the retrieved Experience.

    \item \textbf{Ground the Triggering Condition}: Adapt the condition to the current visual state, spatial uncertainty, available evidence, or tool feedback so that it is clear when the Experience should be applied.

    \item \textbf{Operationalize the Action}: Translate generic advice into concrete local guidance for evidence acquisition, spatial tool selection, geometric verification, reference-frame checking, or error recovery when appropriate.

    \item \textbf{Respect Evidence}: Do not convert retrieved Experiences into factual claims about the current scene. Visual observations and verified tool outputs remain the primary evidence.

    \item \textbf{Avoid Procedural Overreach}: Keep the rewritten Experience as concise tactical guidance rather than expanding it into a complete multi-step Skill or task plan.

    \item \textbf{Filter Irrelevant Guidance}: Omit any Experience whose triggering condition is not supported by the current task or visual context.

    \item \textbf{Keep One Entry per Identifier}: Rewrite each kept Experience separately under its original identifier. Do not combine, split, or add Experiences.

\end{enumerate}

Use concise and constructive language.

Avoid task-specific answers and unnecessary memorization of exact object identities or coordinates.

\textbf{Required Output Format:}
\scriptsize
\begin{verbatim}
{
  "E3": {
    "condition": "When the left-right relation is ambiguous because the viewpoint is unclear, ...",
    "action": "First establish the relevant camera or object reference frame before comparing positions."
  },
  "E8": {
    "condition": "When apparent object size is being used as a proxy for depth, ...",
    "action": "Verify the relation using reconstruction or explicit depth evidence before concluding."
  }
}
\end{verbatim}
\normalsize
Output ONLY a valid JSON object. If no retrieved Experience is applicable, output:
\begin{verbatim}
{}
\end{verbatim}
\end{tcolorbox}

\end{document}